\documentclass[10pt,twocolumn,letterpaper]{article}

\usepackage{iccv}
\usepackage{times}
\usepackage{epsfig}
\usepackage{graphicx}
\usepackage{amsmath}
\usepackage{amssymb}

\usepackage{array}
\usepackage{multirow}
\usepackage[dvipsnames,table]{xcolor}
\usepackage{colortbl}
\usepackage{enumitem} 
\usepackage{arydshln}
\usepackage{tabularx}
\usepackage{booktabs} 
\usepackage{soul}
\usepackage{fontawesome}  

\usepackage{xspace}

\definecolor{mydarkblue}{rgb}{0,0.08,0.45}
\definecolor{custom_red}{rgb}{0.8500,0.3250,0.0980}%
\definecolor{custom_green}{rgb}{0.4660,0.6740,0.1880}%
\definecolor{custom_blue}{rgb}{0,0.4470,0.7410}%

\colorlet{colorFst}{Green!25}       
\colorlet{colorSnd}{SpringGreen!45} 
\colorlet{colorTrd}{Yellow!30}      
\colorlet{colorLow}{darkgray!30}    
\newcommand{\fs}{\cellcolor{colorFst}\bf}   
\newcommand{\nd}{\cellcolor{colorSnd}}      
\newcommand{\rd}{\cellcolor{colorTrd}}      

\newcommand{\ra}[1]{\textbf{\textcolor{custom_red}{R1}}\xspace}
\newcommand{\re}[1]{\textbf{\textcolor{custom_green}{R2}}\xspace}

\newcommand{\boldparagraph}[1]{\vspace{0.1em}\noindent{\bf #1} }

\definecolor{darkgreen}{rgb}{0,0.5,0}
\definecolor{semidarkgreen}{rgb}{0,0.8,0}

\newcommand{\Ours}{CLikER}

\newcommand{\setR}{\mathbb{R}}
\newcommand{\loss}{\mathcal{L}}
\newcommand{\expect}{\mathbb{E}}
\newcommand{\graph}{\mathcal{G}}
\newcommand{\nodeSet}{\mathcal{N}}
\newcommand{\edgeSet}{\mathcal{E}}

\newcommand{\lock}{\text{\faLock}}

\newcommand{\PreserveBackslash}[1]{\let\temp=\\#1\let\\=\temp}
\newcolumntype{C}[1]{>{\PreserveBackslash\centering}p{#1}}
\newcolumntype{R}[1]{>{\PreserveBackslash\raggedleft}p{#1}}
\newcolumntype{L}[1]{>{\PreserveBackslash\raggedright}p{#1}}

\makeatletter
\def\adl@drawiv#1#2#3{%
        \hskip.5\tabcolsep
        \xleaders#3{#2.5\@tempdimb #1{1}#2.5\@tempdimb}%
                #2\z@ plus1fil minus1fil\relax
        \hskip.5\tabcolsep}
\newcommand{\cdashlinelr}[1]{%
  \noalign{\vskip\aboverulesep
           \global\let\@dashdrawstore\adl@draw
           \global\let\adl@draw\adl@drawiv}
  \cdashline{#1}
  \noalign{\global\let\adl@draw\@dashdrawstore
           \vskip\belowrulesep}}
\makeatother

\usepackage[pagebackref=true,breaklinks=true,colorlinks,bookmarks=false]{hyperref}

\usepackage[capitalize]{cleveref}
\crefname{section}{Sec.}{Secs.}
\Crefname{section}{Section}{Sections}
\Crefname{table}{Table}{Tables}
\crefname{table}{Tab.}{Tabs.}

\iccvfinalcopy 

\def\iccvPaperID{10932} 

\begin{document}

\title{Detecting Objects with Context-Likelihood Graphs and Graph Refinement}

\author{Aritra Bhowmik$^1$ \hspace{5pt} Yu Wang$^2$ \hspace{5pt} Nora Baka$^2$ \hspace{5pt} Martin R. Oswald$^1$ \hspace{5pt} Cees G. M. Snoek$^1$\\
Atlas Lab - $^1$University of Amsterdam \hspace{20pt} $^2$TomTom\\
{\tt\small \{a.bhowmik, m.r.oswald, cgmsnoek\}@uva.nl} \hspace{6pt} {\tt\small \{nora.baka, yu.wang\}@tomtom.com}
}
\maketitle

\begin{abstract}
The goal of this paper is to detect objects by exploiting their interrelationships. Contrary to existing methods, which learn objects and relations separately, our key idea is to learn the object-relation distribution jointly. We first propose a novel way of creating a graphical representation of an image from inter-object relation priors and initial class predictions, we call a context-likelihood graph. We then learn the joint distribution with an energy-based modeling technique which allows to sample and refine the context-likelihood graph iteratively for a given image. Our formulation of jointly learning the distribution enables us to generate a more accurate graph representation of an image which leads to a better object detection performance. We demonstrate the benefits of our context-likelihood graph formulation and the energy-based graph refinement via experiments on the Visual Genome and MS-COCO datasets where we achieve a consistent improvement over object detectors like DETR and Faster-RCNN, as well as alternative methods modeling object interrelationships separately. Our method is detector agnostic, end-to-end trainable, and especially beneficial for rare object classes.
%
\end{abstract}


\section{Introduction}  \label{sec:intro}


Object detection is a classical task in computer vision~\cite{arkin2021survey, liuIJCV20202}, where methods in the last decade have evolved from deep convolutional architectures~\cite{ren2015faster_rcnn, redmon2016you, li2017light, cai2018cascade} to self-attention based transformer networks~\cite{nicolas2020detr, zhu2020deformable, nguyen2022boxer, wu2020visual}. Although these two approaches vary widely in architecture, they both focus exclusively on the image space for feature representation, rather than incorporating contextual semantics about object categories. Once fed with sufficient data, these architectures might implicitly learn some object relations, but prior works~\cite{jiang2018hybrid, xu2019spatial, xu2019reasoning, xu2020universal} have shown that explicitly modeling object relations enhances object detection performance by a substantial margin.

\begin{figure}[t!]
  \centering
  \footnotesize
  \setlength{\tabcolsep}{1.5pt}
  \newcommand{\sz}{0.48}
  \begin{tabular}{cc}
    \textbf{Faster-RCNN~\cite{lin2017feature}} & \textbf{Ground truth detections}\\
    \includegraphics[width=\sz\linewidth]{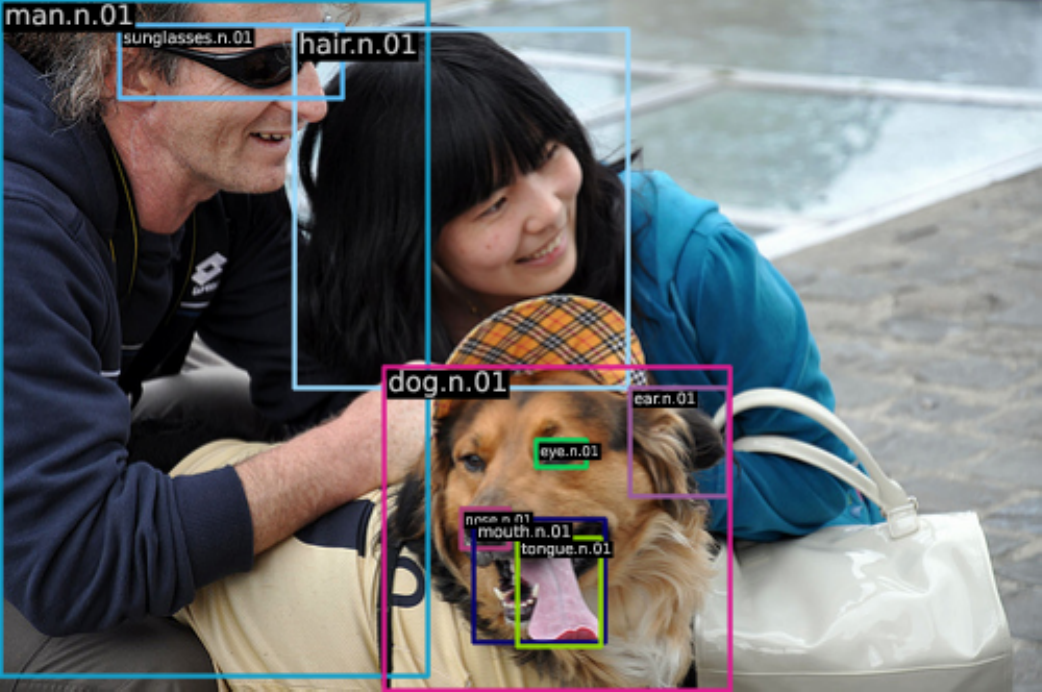} &
    \includegraphics[width=\sz\linewidth]{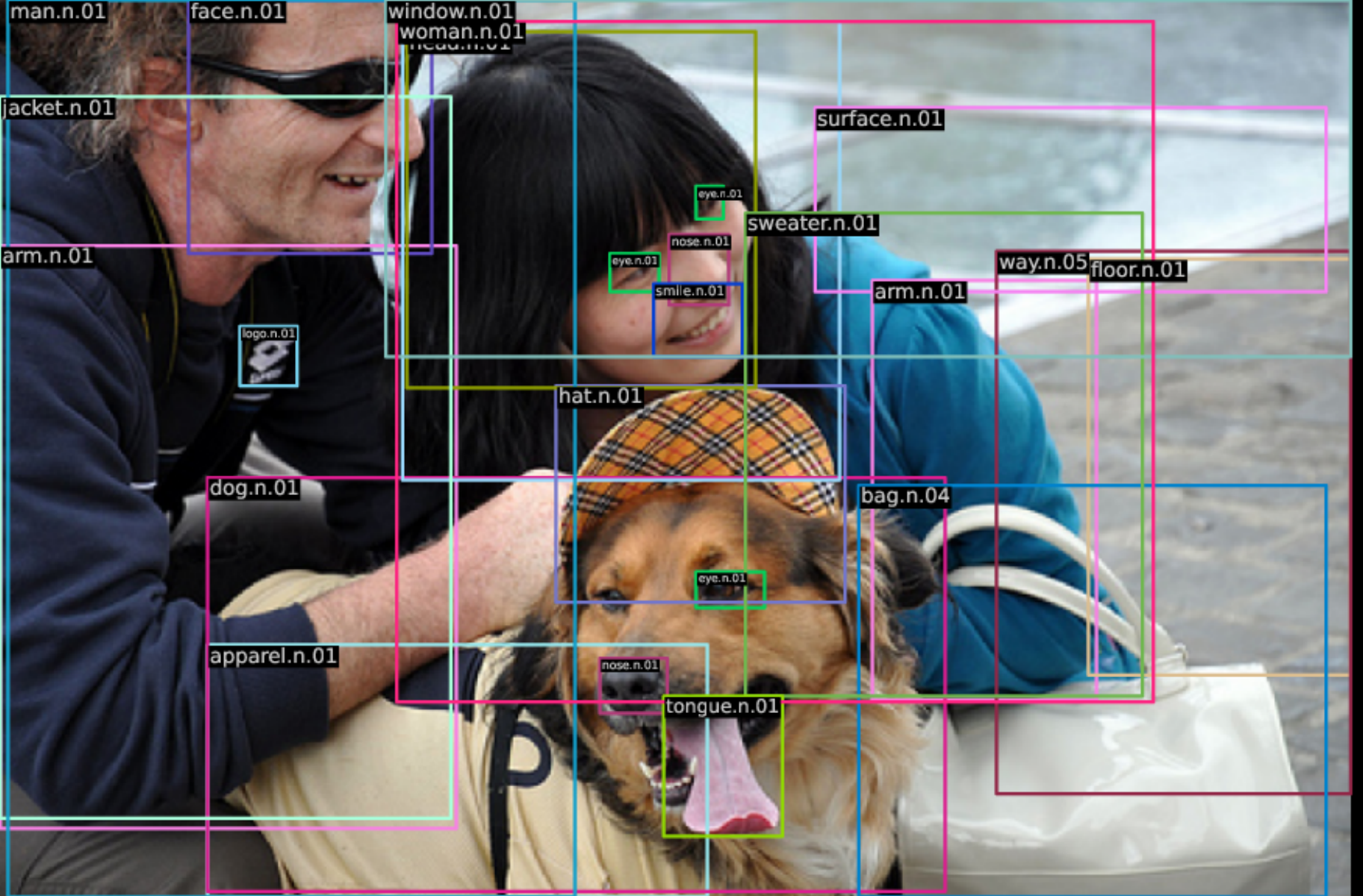} \\
    \includegraphics[width=\sz\linewidth]{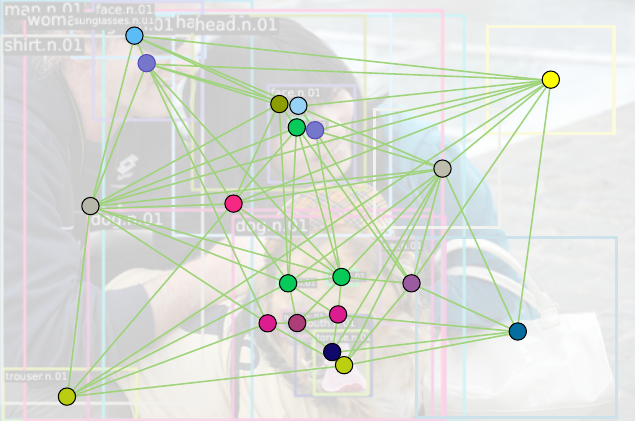} &
    \includegraphics[width=\sz\linewidth]{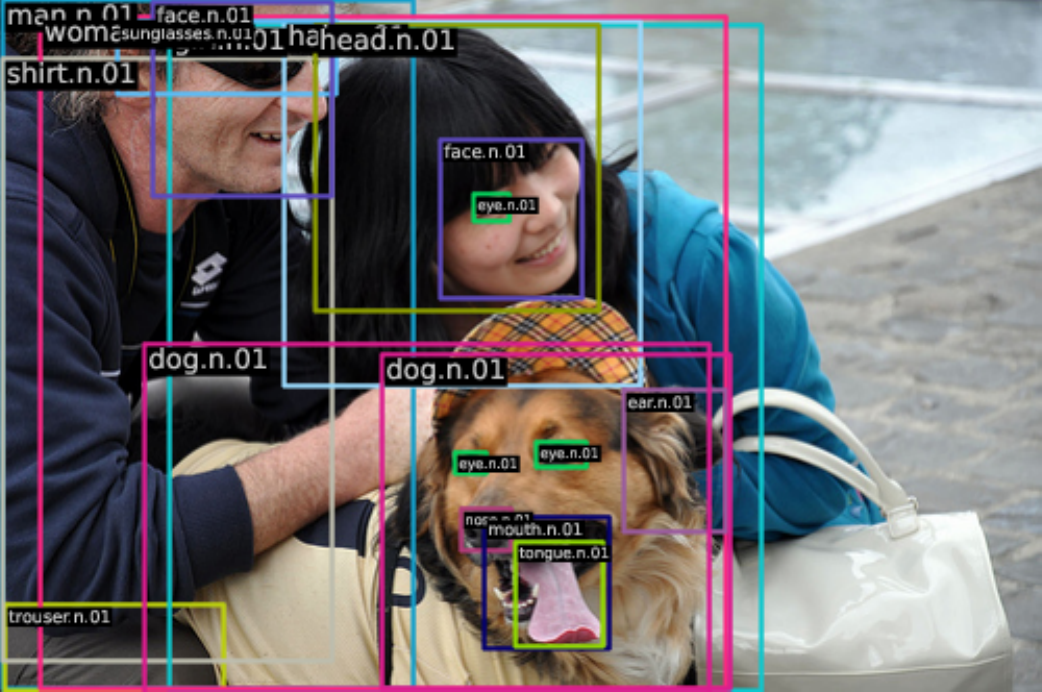} \\
    \textbf{Initial context-likelihood graph}  & \textbf{Results after energy-refinement}\\[4pt]
  \end{tabular}
  \caption{We generate a context-likelihood graph from the initial predictions of a base detector, here Faster-RCNN, by incorporating inter-object relation priors. Having learnt the joint object-relation distribution, we iteratively sample a refined graph by energy optimization. This enables us to predict objects undetected by Faster-RCNN, due to their relations or co-occurrence patterns with already detected objects. For instance, the detection of a face makes the occurrence of eyes or a nose much more likely and vice versa. Likewise for the trousers in the presence of a man, which are not even present in the ground truth.}
  \label{fig:teaser}
  
\end{figure}

Amongst existing work on modeling object relations, Jiang~\etal~\cite{jiang2018hybrid} hypothesize that in case a dataset has little information about certain classes, incorporating prior knowledge in the training of a detector should improve detection. This prior knowledge could be from another source, or it could even be distilled from the original dataset. The authors establish a novel way of explicitly incorporating semantic prior knowledge into a base object detection framework by predicting object-relations from features, creating a graphical representation of the image. Xu~\etal~\cite{xu2019spatial} propose to predict object relations solely from object features and their spatial orientation, without any extra prior information, by creating a sparse graphical representation. Both these methods are dependent on the correctness of the predicted relations since they refine the object features for classification. Instead of predicting an explicit graph representation, Xu~\etal~\cite{xu2019reasoning} enhance object detection by evolving high-level semantic representations globally, rather than relying solely on visual features. This is accomplished through a network that generates a global semantic pool and adapts each object's features by attending to different semantic contexts in the pool, automatically discovering the most relevant categories for feature evolution. Although all these works utilize prior knowledge about objects in their problem formulation, they do so consecutively making them critically dependant on the object features. Instead, we propose to model the objects and their relations jointly from the start, in order to have a better understanding of the scene and infer better class predictions. 

Our main contributions are thereby as follows:
\begin{enumerate}[itemsep=0.1pt,topsep=3pt,leftmargin=*,label=\textbf{(\arabic*)}]
  \item We propose a novel way of leveraging inter-object relation priors for object detectors during training by creating a context-likelihood graph with relation edges based on initial class predictions. (\cref{sec:graph_priors}). 
  \item We demonstrate the potential of our context-likelihood graph with an empirical evaluation showing that incorporating inter-object relations from the start leads to substantially better object detection rates when using a graph built with ground truth class predictions (\cref{sec:potential}).
  \item We introduce an energy-based method for learning object-relation joint distribution, which allows to iteratively refine our context-likelihood graph to further benefit us in our end task (\cref{sec:graph_refinement}).
\end{enumerate}
%
Experiments on the Visual Genome~\cite{krishna2017visual} and MS-COCO~\cite{lin2014microsoft} datasets demonstrate our method is detector agnostic, end-to-end trainable, and especially beneficial for rare object classes. What is more, we establish a consistent improvement over object detectors like DETR~\cite{nicolas2020detr} and Faster-RCNN~\cite{ren2015faster_rcnn}, as well as alternative methods modeling object interrelationships.
Before introducing our method, we first provide more background on related works.

%


\section{Related Work}  \label{sec:formatting}

\boldparagraph{Object detection.}
Modern object detection models can be broadly classified into two groups: those based on convolution architectures and those based on transformers. In the convolution family,  methods like Faster-RCNN~\cite{ren2015faster_rcnn} and R-FCN~\cite{dai2016r} are well established. They consists of a region proposal network to generate regions of interest and a classification and bounding box regression module to predict a set of objects. YOLO~\cite{redmon2016you} and SSD~\cite{liu2016ssd} predict objects directly from convolutional feature maps. ObjectBox \cite{zand2022objectbox} treats objects as center-points in a shape and size agnostic fashion and allows learning at all scales.  All these methods typically utilize a pretrained backbone like ResNet~\cite{he2016resnet} for feature extraction. Recently, there has been a rise in methods which use an attention mechanism in a transformer network. For instance, ViT~\cite{wu2020visual}, SwinTransFormer \cite{liu2021swin} and MViTv2 \cite{li2022mvitv2}. Carion~\etal~\cite{nicolas2020detr} introduced a detection transformer (DETR) which uses an attention mechanism for object detection in an end-to-end manner. To speed up convergence and reduce the computational cost for the self-attention on image features, Zhu~\etal~\cite{zhu2020deformable} introduced multi-head deformable attention replacing the dot-product in the attention computation with two linear projections for sampling points and computing their attention weights. Nguyen~\etal~\cite{nguyen2022boxer} introduced BoxeR which samples boxes instead of points for attention computation, and demonstrates utility for object detection, instance segmentation and 3D detection. Although these object detectors use spatial context information in the image space, they do not explicitly model the relation between (detected) objects. Using Faster-RCNN and DETR as representative examples of the convolution and transformer approach to object detection, we demonstrate that creating a graphical representation of an image from inter-object relation priors and initial class predictions improves object detection.

\begin{figure*}[t!]
  \centering
  \small
  \includegraphics[width=\linewidth]{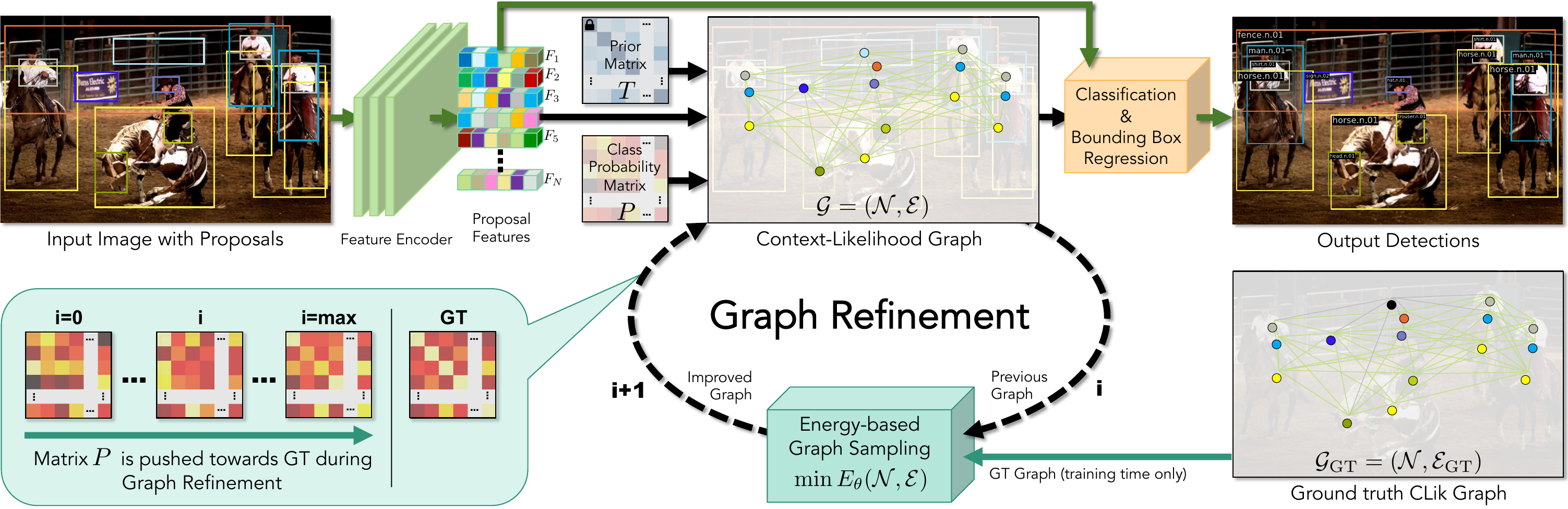}
  \caption{\textbf{Method overview.} Our end-to-end trainable architecture extends object detectors by inserting a context-likelihood (CL) graph between the first stage (object proposal generation) and the second stage (proposal classification). The default detector pipeline is visualized by the \textcolor{OliveGreen}{green} arrows. 
  With the use of our context-likelihood graph before the classification stage, the proposals can be better assessed since the presence or absence of other objects can be explicitly taken into account which for example can enforce proposals with low visual evidence.
  The context-likelihood graph is used in two ways: 
  (1) During training, to learn the energy landscape and train the classifier.
  (2) At test time, to obtain the refined graph via energy-based optimization.}
  \label{fig:architecture_overview}
\end{figure*}

\boldparagraph{Object graph representation.}
Aside from classic object detection, many works have aimed to represent the interactions between objects and scenes. Early works \cite{felzenszwalb2010object, galleguillos2008object, mottaghi2014role} use object relations as a post-processing step on simpler datasets like PASCAL VOC~\cite{Everingham15}. Recent works model objects and relations together in a graph structure, \eg, \cite{chen2018iterative, dai2017detecting, kipf2016semi, marino2016more} where Xu~\etal~\cite{xu2021joint} use a unified conditional random field to model the joint distribution of all the objects and their relations in a scene graph. However, this requires ground truth labels for both the objects and their relations for each image during training. To avoid explicit relation annotations, others, \eg, \cite{jiang2018hybrid, xu2019spatial, xu2019reasoning, xu2020universal} exploit priors like co-occurrence statistics or attributes of object classes to obtain a graph representation of an image. Jiang~\etal~\cite{jiang2018hybrid} predict the edge values representing the relations between object pairs, whereas Xu~\etal~\cite{xu2019spatial} propose a form of self-attention to obtain these edges. In a follow-up work, Xu~\etal~\cite{xu2019reasoning} further introduce a global semantic pool, created using the weights of individual object classifiers, as a relationship prior instead of generating an explicit graph representation. To aggregate information from different domains, they also introduce graph transfer learning \cite{xu2020universal}. All these works enhance the object feature representation by propagating the relation information via a graph structure. These works utilize the relation information as is and do not consider the joint modeling of the relation with the objects in the scene. In our work, we learn the joint distribution of objects and their relations by sampling meaningful relations for object pairs before the graph propagation. 

\boldparagraph{Energy-based models.} Energy-based models (EBMs) bring forth simplicity and generality in likelihood modeling \cite{ackley1985learning, hinton2006unsupervised, lecun2006tutorial, hinton2002training}. Various tasks like image generation \cite{du2020compositional, nie2021controllable}, out-of-distribution detection \cite{liu2020energy}, memory modeling \cite{bartunov2019meta} and anomaly detection \cite{dehaene2020iterative} have been addressed with the help of EBMs. Recently, there has been an increasing interest in applying EBMs to the task of image generation.
Both Du \etal \cite{du2019implicit} and Grathwohl~\etal~\cite{grathwohl2019your} learn the underlying data distribution and use the learnt distribution to generate new data. Moving forward from only modeling the data distribution, \cite{grathwohl2019your, xiao2020vaebm} model the joint distribution of data and labels to benefit classification tasks by a joint energy-based model. Recently, Suhail \etal \cite{suhail2021energy} propose an energy-based learning framework for scene graph generation to learn the joint distribution of objects and their relations together. To the best of our knowledge, our work is the first to utilize energy-based modeling to learn a graph representation of an image from just object annotations.

\section{Method}

Given an input image and object proposals, we aim to detect all objects in the image in the form of bounding boxes and their corresponding class labels. An overview of our method is provided in \cref{fig:architecture_overview}. The green arrows in the figure show the regular object detection pipeline. We add two key components to it which are described in the following subsections:
First, we present context-likelihood graphs, a new way of building a graph representations, based on initial class predictions in \cref{sec:graph_priors}. We demonstrate the potential of context-likelihood graphs with a simple oracle experiment in \cref{sec:potential}. Second, we introduce an iterative method for obtaining refined graph via energy-based optimization in \cref{sec:graph_refinement}.

\subsection{Context-Likelihood Graphs}  \label{sec:graph_priors}

\boldparagraph{Components.}
For an input image $I$, visual features $\{F_i\}_1^N$ of dimension $D$ are extracted via a base object detector for $N$ object proposals. These feature vectors along with contextual prior knowledge are used in pairs \cite{jiang2018hybrid, xu2019spatial} to obtain pair-wise edge relationships, thus creating an edge connectivity matrix $\edgeSet$. A popular source of prior knowledge about objects is the large scale Visual Genome dataset \cite{krishna2017visual} which consists of pairwise relationships between object classes such as ``subject-verb-object'' relationships (\eg, walking, swimming) and spatial relationships (\eg, at, far, in). By considering the most frequent relationship between subject and object boxes in the dataset and counting the co-occurrence among class pairs, a $C {\times} C$ frequent statistics or prior matrix $T$ is generated. Each entry in the matrix is a floating point value between 0 and 1, indicating the strength of the relation between the object pairs in the dataset.

\boldparagraph{Definition.}
We obtain a graphical representation of the input image based on the feature vectors and the edge values defined as
\begin{equation} \label{eq:graph}
  \graph {=} (\nodeSet,\edgeSet). 
\end{equation}
where $\nodeSet$ are the set of nodes and $\edgeSet$ is the edge connectivity matrix. We call this our context-likelihood graph. This graph is propagated via some message passing algorithm to calculate refined features, from which we obtain the final classification and bounding box regression results. 
To create this graph, we model object relations as a function of the initial class predictions, instead of directly predicting them from feature vector pairs \cite{jiang2018hybrid}. Corresponding to the feature vectors $\{F_i\}_1^N$, we obtain their classification probabilities $P$ from the base detector where $P \in \setR^{N \times C}$, $C$ being the number of classes in the dataset. Given the class prediction matrix $P$ and the prior matrix $T$, we calculate the edge connectivity matrix as: 
%
%
\begin{equation} \label{eq:PTP}
  \edgeSet {=} P \times T \times P^T \enspace. 
\end{equation}
Each term $e_{ij}$ in the matrix $\edgeSet$ is a result of all possible probabilistic combinations of classes for nodes $i$ and $j$ along with their corresponding relation prior values. 
 By modeling the edge matrix in this manner, we penalize the class predictions jointly with the edge values, the former guiding the correctness of the latter. Thus, optimizing for the final task of classification would ensure refined edge values, which in turn guides the end task. Our joint task loss function is given by:  
\begin{equation} \label{eq:taskloss}
  \loss_{Task} {=} \loss_{CE}\{P^{'}, P^{*}\} + \loss_{CE}\{P, P^{*}\}  \enspace,
\end{equation}
where $\loss_{CE}$ is the cross-entropy loss, $P^{'}$ is the final classification probability matrix output from our model, $P$ is the initial class prediction matrix of the base detector and $P^*$ is the ground truth object labels matrix. Optimality is obtained when initial predictions equal the ground truth labels. 

\subsection{Potential of Context-Likelihood Graphs}  \label{sec:potential}
%
To test the optimal condition of our context-likelihood graph, we perform an oracle experiment where we replace the set of edges $\edgeSet$, 
with ground truth relation set $\edgeSet_{GT}$. 
To calculate $\edgeSet_{GT}$, $P$ is replaced by $P^*$ in Eq.~\eqref{eq:PTP} where each row of $P^*$ is a one-hot encoded class label corresponding to each object proposal, obtained by IoU-matching. To perform IoU-matching, we check the overlap of the regions of interests of each proposal with each ground truth bounding box, and assign the class label of the highest overlapping ground truth bounding box to the proposal.
The ground truth context-likelihood graph $\graph_\text{GT} {=} (\nodeSet,\edgeSet_\text{GT})$, optimally captures the underlying object-pair relationships. Propagating this graph generates the refined object features which are used for the final classification results. As ground truth we rely on the 1000 ($\text{VG}_{1000}$) and 3000 ($\text{VG}_{3000}$) most frequent object classes from Visual Genome~\cite{krishna2017visual}.

\begin{figure}[t]
  \centering
  \footnotesize
  \includegraphics[trim={0 1.2cm 0 0},clip,width=0.8\linewidth]{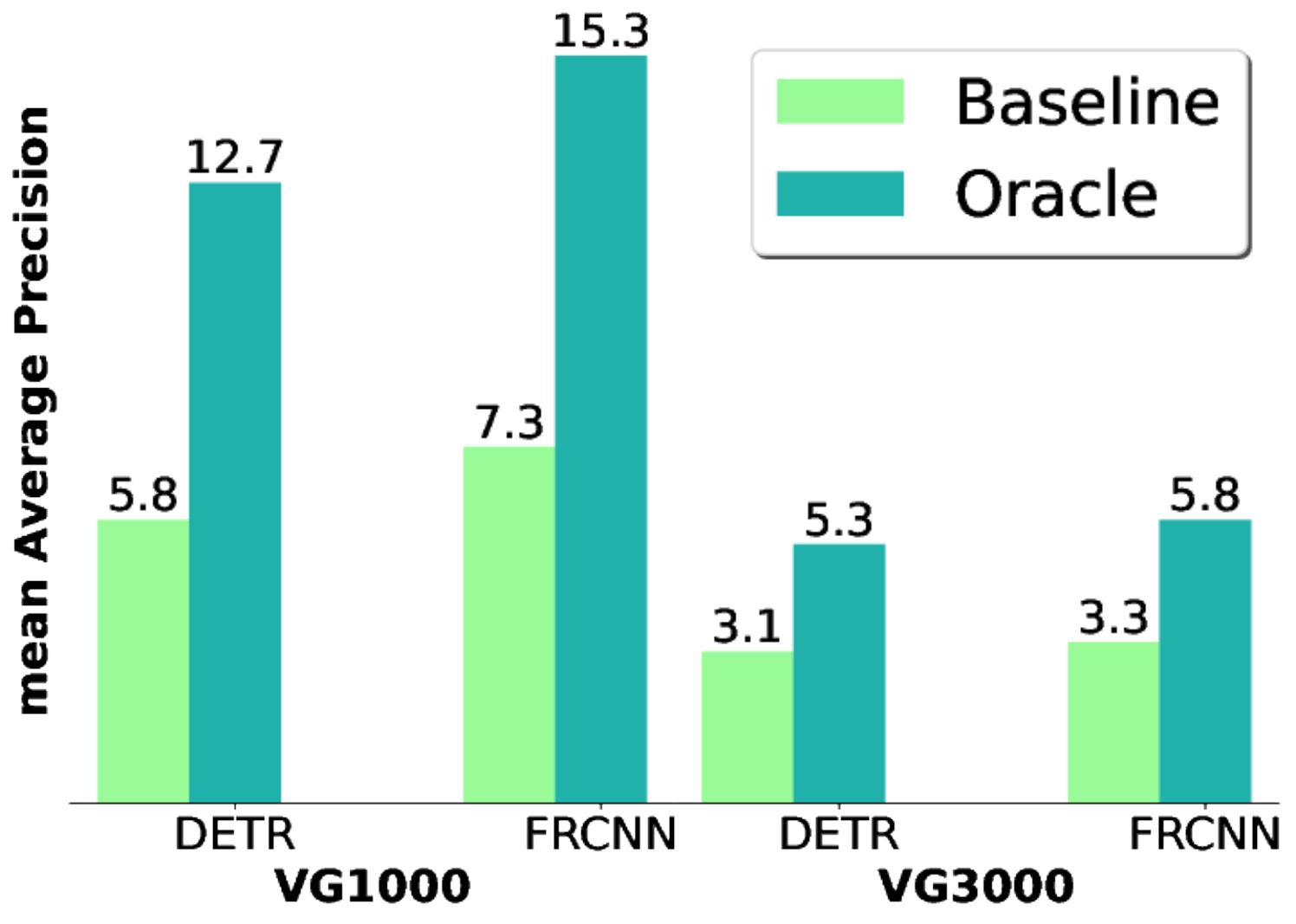}\\
  \begin{tabular}{C{0.2cm}C{2.8cm}C{2.8cm}}
     & $\text{VG}_{1000}$ & $\text{VG}_{3000}$
  \end{tabular}
  \vspace{1pt}
  \caption{\textbf{Potential of context-likelihood graph.} An oracle experiment with the ground truth context-likelihood graph, created by replacing the initial predictions by ground truth class labels in Eq.~\eqref{eq:PTP}, yields substantial improvements on the $\text{VG}_{1000}$ and $\text{VG}_{3000}$ partitions of visual genome~\cite{krishna2017visual}, for both a DETR~\cite{nicolas2020detr} and F-RCNN~\cite{ren2015faster_rcnn} object detector.}
  \label{fig:gt_improvements}
\end{figure}

\boldparagraph{Analysis.}
We use two proposal-based object detectors as our base network, the transformer-based DETR~\cite{nicolas2020detr} and convolution-based Faster-RCNN~\cite{ren2015faster_rcnn}. Both these methods first identify rectangular regions of interest in an image, extract their corresponding feature encodings and create classification predictions for each such region. From \cref{fig:gt_improvements}, we see that knowing the relations between objects in a scene improves the detection results of the base network by a large margin. 
For the $\text{VG}_{1000}$ dataset  we achieve an $119\%$ relative improvement in mean average precision results for DETR, while for Faster-RCNN, it is around $110\%$. Note that we do not embed the ground truth information in any way other than to use it for looking up the true relation prior between proposed object-pairs.

This experiment shows the potential of our context-likelihood graph, irrespective of base object detectors, in the optimal setting. However, we will not have access to ground truth during inference, which prevents us from reaching optimality in the above described fashion. The edge matrix $\edgeSet$ in Eq.~\eqref{eq:PTP} is a function of the class probability matrix, $P$ which in turn is generated from feature vector set $\nodeSet$. The optimal solution $\edgeSet_{GT}$ lies in the joint distribution space of $(\nodeSet,\edgeSet)$. In order to sample a refined context-likelihood graph, we need to learn the joint distribution, for which we propose an energy-based method in the following section.


\subsection{Energy-Based Graph Refinement}  \label{sec:graph_refinement}
%
Before detailing our graph refinement approach, we first provide a brief background on energy-based models.

\boldparagraph{Energy model background.}
%
For two variables $x$ and $y$, energy-based models \cite{lecun2006tutorial} represent their joint probability distribution as $p_{\theta}(x,y){=}\frac{\exp(-E_\theta(x,y))}{Z(\theta)}$, where $\exp(-E_\theta(x,y))$ is known as the energy function that maps each input pair to a scalar value and Z($\theta$) {=} $\int \exp(-E_\theta(x,y))$ is referred to as the normalization constant or partition function. Computing ${Z(\theta)}$ typically becomes intractable since it computes the integration over the entire joint input space of $(x, y)$. Thus, we cannot learn the parameter $\theta$ of the energy-based model by using a direct log likelihood approach. 
In order to understand the problems with the partition function, we investigate the derivatives \cite{du2019implicit, song2021train} of the log likelihood function given as: 
\begin{align} \label{eq:1}
    \frac{\partial\log p_{\theta}(x, y)}{\partial\theta} &= \\
      \expect_{p_d(x, y)} &\left[-\frac{\partial E_{\theta}(x,y)}{\partial\theta}\right] 
    + \expect_{p_\theta(x, y)}\left[\frac{\partial E_{\theta}(x,y)}{\partial\theta}\right] \nonumber \enspace.
\end{align}
The first term is an expectation over the data distribution and the second term is over the model distribution. Maximizing the log-likelihood $\log p_\theta$(x,y) is equivalent to minimizing the KL-divergence $D_\text{KL}\big(p_d(x,y)||p_\theta(x,y)\big)$ which is achieved by minimizing the contrastive divergence~\cite{hinton2002training}:
\begin{equation} \label{eq:2}
    \loss = D_\text{KL}\big(p_d(x,y)||p_\theta(x,y)\big) - 
            D_\text{KL}\big(q_\theta(x,y)||p_\theta(x,y)\big) \enspace,
\end{equation}
where $q_\theta(x,y) {=} \prod_{\theta}^{t}p_d(x,y)$ represents $t$ sequential Markov chain Monte Carlo transitions \cite{hinton2002training}, used to approximate the model distribution $p_\theta(x,y)$. An effective example being Stochastic gradient Langevin Dynamics (SGLD) \cite{welling2011bayesian}: 
\begin{align} \label{eq:3}
    x^{i+1} = x^{i} - \frac{\lambda}{2} \frac{\partial E_{\theta}(x^i)}{\partial x^{i}} + \epsilon, \quad s.t. \quad \epsilon \sim \mathbb{N}(0, \lambda)  \enspace.
\end{align}
Eq.~\eqref{eq:2} can be rewritten in expectation terms as follows:
\begin{align} \label{eq:4}
    \loss = 
    & \expect_{p_d(x, y)}[E_{\theta}(x,y)] - \expect_{\lock(q_\theta(x, y))}[E_{\theta}(x,y)] \\
    & \expect_{q_\theta(x, y)}[E_{\lock(\theta)}(x,y)] + \expect_{q_\theta(x, y)}[\log q_\theta(x, y)] \nonumber   \enspace.
\end{align}
Here, the lock symbol $\lock(\cdot)$ denotes that the variables in the argument list are not optimized, \ie, their gradient flow is stopped. With this knowledge of training an energy model to learn a joint probability distribution, we formulate our context-likelihood energy-refined graph. 

\boldparagraph{Energy model for object detection.}
%
Langevin dynamics \cite{welling2011bayesian} simulate samples of the distribution expressed by the energy-based model through gradient-based updates, with no restriction on the sample initialization if the sampling steps are sufficient. We leverage this property to sample a refined graph of the image by setting the context-likelihood graph as the initialization of the Langevin process and then updating it iteratively to obtain a more accurate context-likelihood graph.

Our problem represents the image as a graph $\graph {=} (\nodeSet, \edgeSet)$, where $\nodeSet$ is the set of nodes in the graph represented by a matrix of dimension $R^{n \times D}$, $n$ being the number of objects in the image and $D$ being the object feature dimension, and $\edgeSet \in R^{n \times n}$ is the set of edges represented by the relations between the objects in the image. 
In our case, we represent $\edgeSet$ as a connectivity matrix of dimension $R^{n \times n}$ where each entry is a value between 0 and 1, representing the strength of the relation between the object pairs. 
Although estimating the underlying object relationship of the image is important to us, our main goal is to correctly detect the objects in the image, which is a discriminative task. Therefore we want to learn the conditional distribution $p(z|\nodeSet,\edgeSet)$, where $p(z)$ is the classification probability for the $n$ object proposals, along with the data distribution $p(\nodeSet,\edgeSet)$. Thus we model the joint distribution for our tasks as follows:
\begin{equation} \label{eq:5}
    p_{\theta, \phi}(\nodeSet,\edgeSet,z) = p_{\phi}(z|\nodeSet,\edgeSet)\frac{\exp(-E_\theta(\nodeSet,\edgeSet))}{Z_{\theta, \phi}},
\end{equation}
where $p_{\phi}(z|\nodeSet,\edgeSet)$ represents the classification model and $E_\theta(\nodeSet,\edgeSet)$ is the energy function, which is modeled by a neural network parameterized by $\theta$. An image $I$ is initially passed through the base detector producing initial predictions which are used to create the context-likelihood graph $\graph_0 {=} (\nodeSet_0, \edgeSet_0 \rangle$. We consider $\graph_0$ to be the initial state of our Langevin dynamics sampling process which is updated through the energy function $E_\theta(\nodeSet,\edgeSet)$ to produce the refined graph, $\graph_t$ after $t$ sampling iterations and then passed through the message passing and the classification model $p_{\phi}(z|\nodeSet,\edgeSet)$ to get the final predictions.

\boldparagraph{Optimization.}
We optimize the model parameters jointly by minimizing the joint loss function stated below 
%
\begin{align} \label{eq:6}
     \loss_\text{total} = 
        & -\expect_{p_d(\nodeSet,z)}[\log p_{\phi}(z|\nodeSet\!,\edgeSet)]
          + \expect_{p_d(\nodeSet,z)}[E_{\theta}(\nodeSet\!,\edgeSet)] \nonumber \\
        & - \expect_{\lock(q_{\theta,\phi})}[E_{\theta}(\nodeSet\!,\edgeSet)] \\
        & + \expect_{q_{\theta,\phi}(\nodeSet\!,\edgeSet)}\Big[E_{\lock(\theta)}(\nodeSet\!,\edgeSet)
          - \log p_{\lock(\phi)}(z|\nodeSet\!,\edgeSet)\Big] \nonumber \enspace,
\end{align}
where $p_d(\nodeSet,z)$ and $q_{\theta,\phi}(\nodeSet,\edgeSet)$ denote the data distribution and the $t$ sequential MCMC sampling from the joint energy distribution. The first term, referred to as $\loss_{CE}\{P, P^{*}\}$ in Eq.~\eqref{eq:taskloss} maximizes the data log likelihood for ground truth samples by training the classification model; the second and third terms are the ground truth and sampled energy values which trains the energy model; the fourth term corresponds to the KL divergence loss which trains the sampling process, and the final term corresponds to the log likelihood of the sampled data. Training the whole network jointly forces the classification model to adapt to ground truth samples, minimizes the ground truth energy value, and forces the sampling process to sample a graph representation whose energy will be close to the ground truth energy value, as well as a correct classification for the object detection task. 

\boldparagraph{Inference.}
During testing, we pass an image through the base network, gather the object proposals and initial classification probabilities, $P$. We calculate $\edgeSet_0 {=} P \times T \times P^T$ and $\nodeSet_0$ with the feature vectors creating the initial graph state $\graph_0$. Then through the SGLD sampling, we refine this graph $t$ times to produce the final graph state $\graph_t$, which is then fed through the message passing and classification model to obtain the final classification and bounding box results.   
\section{Experiments} \label{sec:experiments}

\subsection{Datasets, Evaluation and Implementation}
%
We conduct our experiments on two established benchmark datasets: Visual Genome~\cite{krishna2017visual} and MS-COCO 2017~\cite{lin2014microsoft}. The task is to localize an object by its bounding box and classify it into one of the preset object categories.

\boldparagraph{Visual Genome.} 
For the Visual Genome dataset~\cite{krishna2017visual}, we use the latest release (v1.4). Following standard practice, we consider two sets of target classes: the 1000 most frequent categories and the 3000 most frequent categories, resulting in two settings: \textit{\textbf{$\text{VG}_{1000}$}} and \textit{\textbf{$\text{VG}_{3000}$}}. We split the 92.9K images of the dataset with objects on these class sets into 87.9K for training and 5K for testing. Instead of the raw names of the categories, we use the synsets as class labels due to inconsistent label annotations, following previous works~\cite{jiang2018hybrid, xu2019reasoning}. The dataset contains an average of 21 objects per image, with a maximum of 145.

\boldparagraph{MS-COCO.} 
We evaluate on the 80 object categories of MS-COCO 2017~\cite{lin2014microsoft} with 118k images for training and 5k for evaluation. There are a maximum of 93 object instances per image in this dataset with an average of 7 objects.

\boldparagraph{Evaluation criteria.} We follow the common evaluation conventions from the literature~\cite{jiang2018hybrid, xu2019reasoning}. Thus, we adopt the full set of metrics from the COCO detection evaluation criteria comprising the mean Average Precision (mAP) across IoU thresholds ranging from 0.5 to 0.95 with an interval of 0.05 as well as mAP at IoU thresholds of 0.5 and 0.75, multiple scales (small, medium and large), Average Recall (AR) with different number of given detections per image ({1, 10, 100}) and different scales (small, medium, big).

\boldparagraph{Two base detectors.}
We implement our method on top of two base detectors: the convolutional Faster-RCNN~\cite{ren2015faster_rcnn} and transformer DETR~\cite{nicolas2020detr}, using feature pyramid networks~\cite{lin2017feature} along with a ResNet-101~\cite{he2016resnet} pretrained on ImageNet~\cite{deng2009imagenet} as their backbone. 
For training Faster-RCNN, we use an image size of $1333 {\times} 800$ and augment with random horizontal flips. We choose $N{=}512$ region proposals, represented by features of dimension $1024$. We use stochastic gradient descent as the optimizer and train on 8 GPUs with a mini-batch size of 2; an initial learning rate of 0.02, reducing it twice by 10; a weight decay of $10^{-4}$ is used along with a momentum of 0.9. We train the model for 20 epochs with learning rate reductions at the 8th and 11th epoch. For DETR, we use 6 encoder and decoder layers, 200 output queries, a learning rate of $10^{-4}$, and train it for 100 epochs on 8 GPUs with a mini-batch size of 2. 

\boldparagraph{Context-likelihood graph propagation.}
Edges in our graph represent the relationship between two objects in a scene. We treat it as an attention weight between the two nodes. Taking inspiration from \cite{gong2019exploiting}, we use a graph attention based network to propagate the node and edge features jointly, additional details of which are provided in the supplementary material. For the network, we use two layers with an input dimension of 1024 and a final output dimension of 256. We use skip connections in between the layers to propagate feature information. $\mathrm{Sigmoid}$ activation is used in the graph message passing computations. The final output features of dimension 256 are concatenated with the original features of dimension 1024, and fed into the classification and bounding box regression layers. 

\boldparagraph{Energy model.}
The graph representation is pooled into a vector via attention pooling.
We use two linear layers after pooling which converts the features into a scalar energy value. The noise added during sampling has a variance of 0.0001; the SGLD learning rate is fixed at 10, and we use 5 steps of sampling to get the final output graph.
%

\subsection{Ablations}
%

\boldparagraph{Context-likelihood graph and graph refinement ablation}
In \cref{tab:ablation_model_parts}, we report the contributions of the components of our method. We see that only using context-likelihood graph, we improve consistently on both base detectors, with context-likelihood energy-refinement our results improve further. For reference, we also provide the results of our oracle experiment with ground truth relation set $\edgeSet_{GT}$.

\begin{table}
  \centering
  \scriptsize
  \setlength{\tabcolsep}{5.0pt}          
  \renewcommand{\arraystretch}{1.0}      
  \newcommand{\grspace}{\hskip 0.7em}   
  \newcommand{\im}[1]{}   
  \newcommand{\ib}[1]{}   
  \begin{tabular}{ll@{\grspace}lllllll}
    \toprule
    & & AP$\uparrow$ & $\text{AP}_{\text{50}}\!\!\uparrow$ & $\text{AP}_{\text{75}}\!\!\uparrow$ & $\text{AP}_{\text{S}}\!\!\uparrow$ & $\text{AP}_{\text{M}}\!\!\uparrow$ & $\text{AP}_{\text{L}}\!\!\uparrow$ \\
    \midrule
    %
    %
    & F-RCNN~\cite{lin2017feature}     &\rd 7.3          &\rd 12.7          &\rd 7.6          &\rd 4.2         
    &\rd       7.8          &\rd 11.0 \\
    & F-RCNN w CL                      &\nd 8.2\im{+0.9} &\nd 13.3\im{+0.6} &\nd 8.6\im{+1.0} &\nd 4.5\im{+0.3} &\nd 8.7\im{+0.9} &\nd 12.8\im{+1.8} \\
    & F-RCNN w GR                 &\fs 8.5\ib{+1.2} &\fs 13.7\ib{+1.0} &\fs 8.8\ib{+1.2} &\fs 4.8\ib{+0.6} &\fs 8.8\ib{+1.0} &\fs 13.4\ib{+2.4}\\
    \cdashlinelr{2-8}
    & \color{gray} F-RCNN w GT     & \color{gray} 15.3\im{+8.0} & \color{gray} 20.1\im{+8.6} & \color{gray} 16.5\im{+8.9} & \color{gray} 12.5\im{+8.3} & \color{gray} 15.9\im{+8.1} & \color{gray} 17.5\im{+6.5} \\
    \cmidrule{2-8}
    & DETR~\cite{nicolas2020detr}      &\rd 5.8          &\rd 9.9           &\rd 5.7          &\rd 2.1          & \rd 5.6         & \rd 10.6  \\
    & DETR w CL                        &\nd 6.6\im{+0.8} &\nd 10.7\im{+0.8} &\nd 6.6\im{+0.9} &\nd 2.6\im{+0.5} &\nd 6.8\im{+1.2} &\nd 11.6\im{+1.0}  \\
    & DETR w GR                   & \fs 6.8\ib{+1.0} & \fs 11.2\ib{+2.3} & \fs 6.8\ib{+1.1} & \fs 2.7\ib{+0.6} & \fs 7.0\ib{+1.4} & \fs 12.1\ib{+1.5} \\
    \cdashlinelr{2-8}
    & \color{gray} DETR w GT                     & \color{gray} 12.7\im{+6.9} & \color{gray} 21.0\im{+11.1} & \color{gray} 12.9\im{+7.2} & \color{gray} 5.2\im{+3.1} & \color{gray} 11.8\im{+6.2} & \color{gray} 19.7\im{+9.1}  \\
    \bottomrule
  \end{tabular}
  \vspace*{0.1mm}
  \caption{\textbf{Context-likelihood graph and graph refinement ablation} on $\text{VG}_{1000}$. We compare the baseline detector with context-likelihood graph (w \textbf{CL}) and the graph refinement (w \textbf{GR}). For reference we show the result obtained with ground truth relation set (w \textbf{GT}) in gray. Best results are highlighted as \colorbox{colorFst}{\bf first}, \colorbox{colorSnd}{second}, and \colorbox{colorTrd}{third}. Using context-likelihood, we achieve consistent improvement while graph refinement improves results further.}
  \label{tab:ablation_model_parts}
  \vspace{-8pt}
\end{table}


\boldparagraph{Detectors and proposals.}
In \cref{tab:num proposals}, we ablate the influence of selecting the number of object proposals for different object detectors. For DETR, there is a  performance gap between 200 and 300 object queries with AP values changing from 6.6 to 7.2. With higher number of queries, the size of attention computation increases exponentially, leading to slower training. For F-RCNN however, the number of proposals during training influences our results only slightly, due to already high number of proposals.

\begin{table}
  \centering
  \footnotesize
  \setlength{\tabcolsep}{4.1pt}        
  \renewcommand{\arraystretch}{1}      
  \newcommand{\grspace}{\hskip 1.4em}  
  \begin{tabular}{ll@{\grspace}ccccc}
    \toprule
    & & {\#Proposals} & AP$\uparrow$ & $\text{AP}_{\text{50}}\!\!\uparrow$ & $\text{AP}_{\text{75}}\!\!\uparrow$  \\
    %
    \midrule
    %
    & \textit{\textbf{This paper}} w DETR & \phantom{1}200 & 6.6 & 11.2 & 6.8 \\
    & \textit{\textbf{This paper}} w DETR & \phantom{1}300 & \rd 7.2 & \rd 11.8 & \rd 7.4 \\
    & \textit{\textbf{This paper}} w F-RCNN & \phantom{1}256 & \nd 8.0 & \nd 13.0 & \nd 8.4 \\
    & \textit{\textbf{This paper}} w F-RCNN & \phantom{1}512 & \fs 8.2 & \fs 13.4 & \fs 8.5 \\
    %
    %
    \bottomrule
  \end{tabular}
  \vspace*{0.1mm}
  \caption{\textbf{Detectors and proposals.} We test our context-likelihood energy-refined graph on $VG_{1000}$~\cite{krishna2017visual} with DETR~\cite{nicolas2020detr} and F-RCNN~\cite{lin2017feature} as backbones. Results on DETR are more affected by the number of proposals than F-RCNN.}
  \label{tab:num proposals}
  \vspace{-6pt}
\end{table}

\begin{figure}[t]
  \centering
  \footnotesize
  \includegraphics[width=1\linewidth, height=0.6\linewidth]{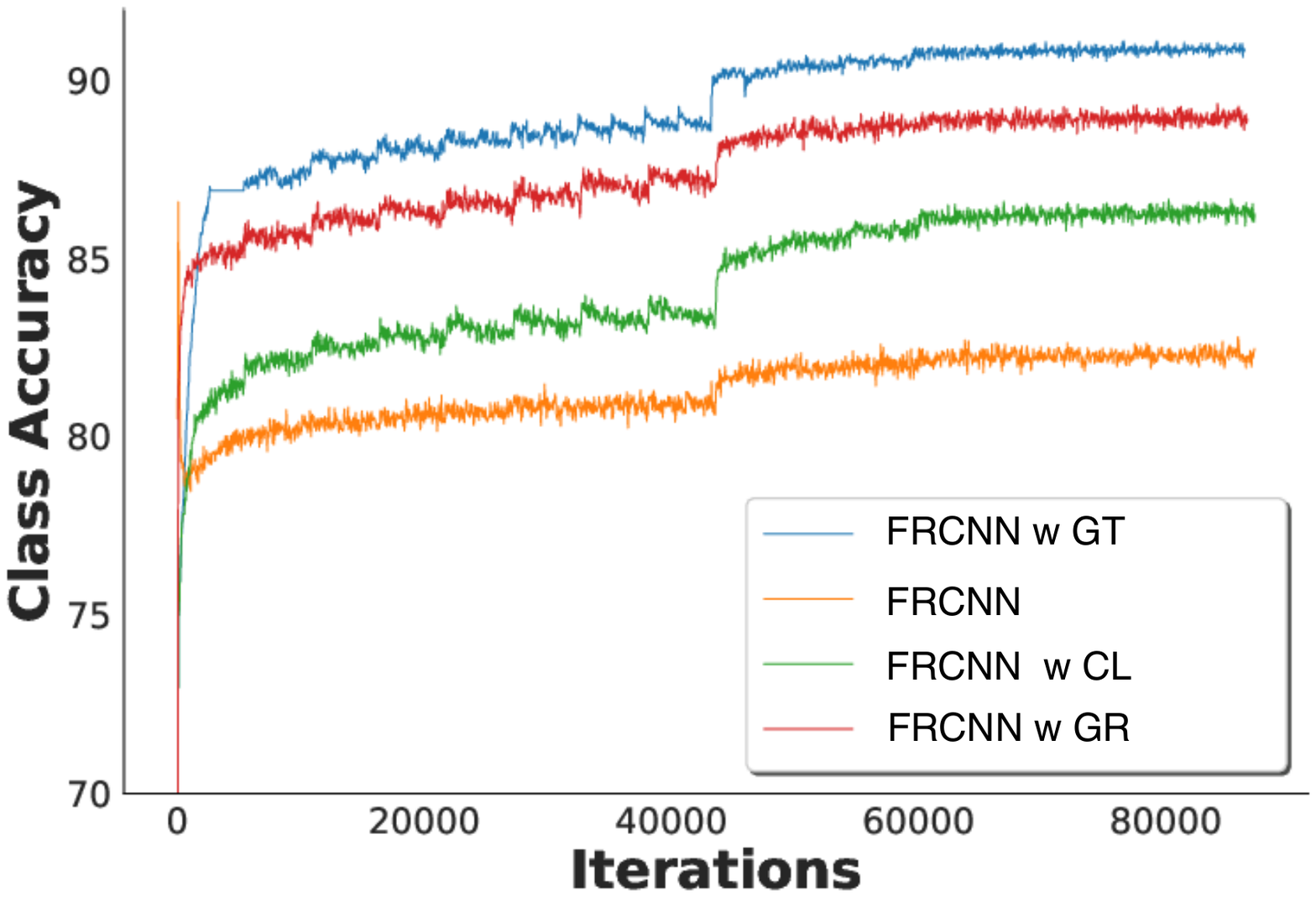}\\[-4pt]
  \caption{\textbf{Energy model training.} 
  As expected, the oracle experiment with ground truth context-likelihood graph (w GT) achieves the highest accuracy. Context-likelihood graph (w CL) reports higher training accuracy over the baseline \textbf{F-RCNN}. When context-likelihood energy-refinement (w GR) is introduced, the accuracy of our method improves further.}
  \label{fig:training_loss}
\end{figure}

\boldparagraph{Energy model training.}  
To understand the impact of energy model training in our framework, we plot the classification accuracy as a function of training time in \cref{fig:training_loss}. As expected, training accuracy is highest for the oracle experiment with ground truth context-likelihood graph. The baseline experiment has the lowest accuracy, followed by the one with context-likelihood graph. For context-likelihood energy-refined graph, the training accuracy further improves, moving closer to the oracle experiment. 

\begin{figure}[t]
  \centering
  \footnotesize
  \includegraphics[width=1\linewidth, height=0.6\linewidth]{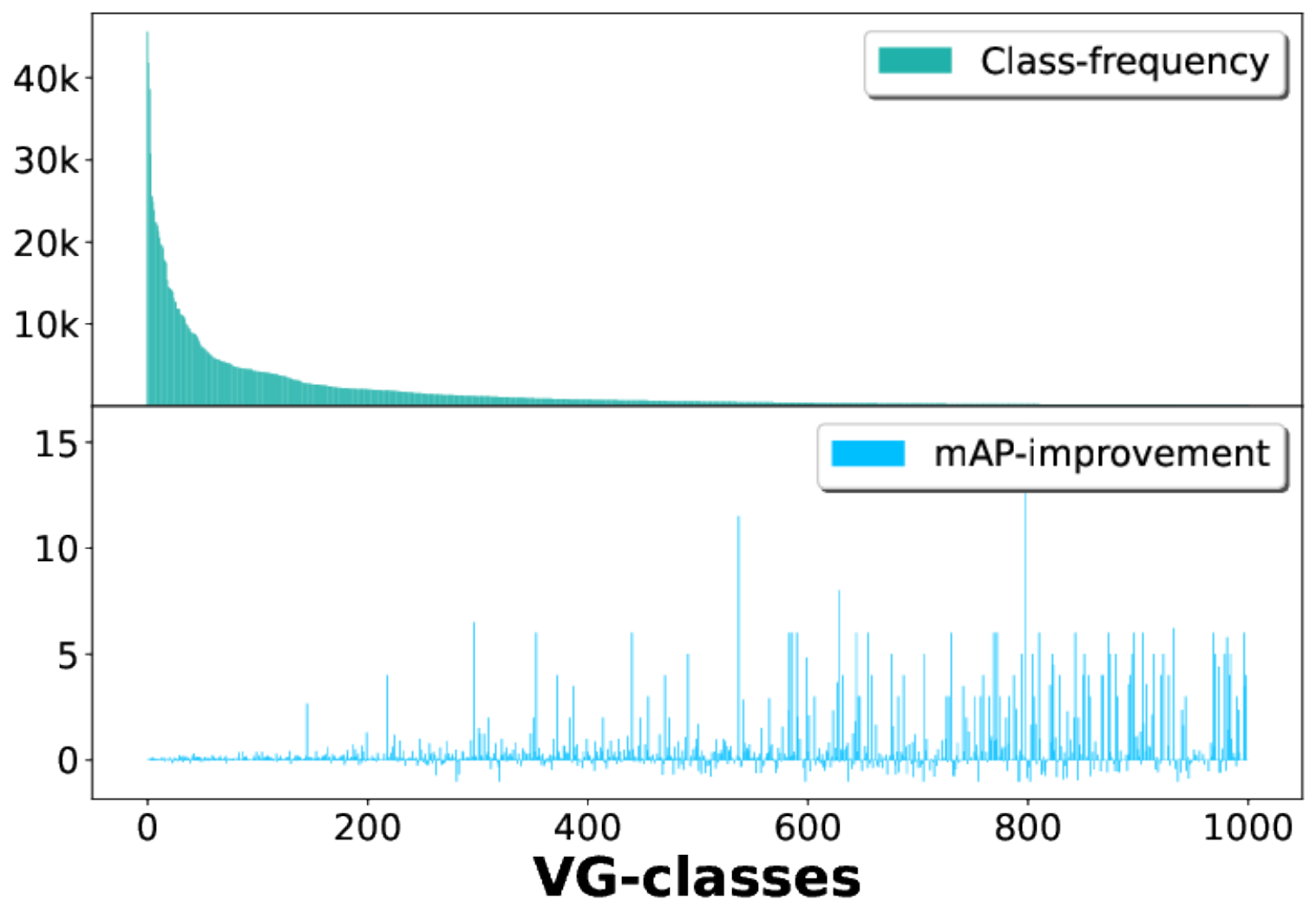}\\[-4pt]
  \caption{\textbf{Benefit for rare objects on $\text{VG}_{1000}$~\cite{krishna2017visual}}. Object classes plotted on the x-axis are arranged in decreasing order of their frequency in the dataset. As expected, our method has a higher relative improvement for rare object classes.}
  \label{fig:class_histogram}
  \vspace{-6pt}
\end{figure}

\boldparagraph{Benefit for rare objects.}
In \cref{fig:class_histogram}, we show the impact of our method on different object classes. We plot the classes on the $x$-axis in decreasing order of frequency of occurrence in the $\text{VG}_{1000}$ dataset; below it we plot the relative improvement of our method over Faster-RCNN for each class. As expected, most of the improvement happens for rare classes of the dataset. We attribute this to the rare objects not having enough examples to learn a meaningful feature representation, so they benefit from inter-object relation statistics more than classes with ample instances. 

\boldparagraph{Inference speed.}
We compare the inference speed of our method with F-RCNN~\cite{lin2017feature} on $\text{VG}_{1000}$~\cite{krishna2017visual}. 
Our method improves over the baseline from $7.3$ to $8.5$ AP, at the expense of a drop in fps from $14.6$ to $9.4$, due to the graph refinement stage. 
The supplementary material also provides a plot for inference time and AP for various refinement steps.

\subsection{Comparison with Others}
%
\boldparagraph{Comparison on $\text{VG}_{1000}$ and $\text{VG}_{3000}$~\cite{krishna2017visual}.} For the comparative evaluation of our method, we first focus on the large-scale detection benchmarks: $\text{VG}_{1000}$ (Visual Genome with 1000 categories) and $\text{VG}_{3000}$ (Visual Genome with 3000 categories). For both these settings, we report the results of our method on top of F-RCNN and DETR and compare with alternative methods modeling object interrelationships which are trained with similar backbone and dataset settings.
As can be seen from \cref{tab:VG_results}, we achieve competitive results for both datasets on almost all metrics when our method is applied on F-RCNN, be it that SGRN~\cite{xu2019spatial} outperforms us slightly on the $AP_L$, $AR_1$ and $AR_L$ metrics for the $\text{VG}_{3000}$ dataset. 
For $\text{VG}_{1000}$ we outperform Xu~\etal \cite{xu2020universal} when we compare with their intra-domain setting using the same backbone and training data (8.5 AP vs. 7.9 AP). They have an edge for their inter-domain setting (8.8 AP), when training with multiple datasets (COCO, VisualGenome and ADE), suggesting our approach may profit as well from training with more data.
We also see a good improvement in our average recall numbers with $AR_M$ improving by 6.5 and 5.1 over baseline F-RCNN for $\text{VG}_{1000}$ and $\text{VG}_{3000}$ respectively. DETR results are below the results on F-RCNN due to DETR not producing region proposals, making it dependant on the number of object instances per image. Nonetheless, we still observe an improvement over DETR on VG1000 from 5.8 AP to 6.8 AP when our method is applied. 

\boldparagraph{Comparison on MS-COCO~\cite{lin2014microsoft}.} We report results based on F-RCNN on MS-COCO and compare with previously reported results in \cref{tab:coco}. We achieve an absolute improvement of 4.2 AP over F-RCNN, and are either better or comparable to alternatives. We do not achieve as much improvements as in the Visual Genome datasets, partly because, the prior statistics have been directly transferred from the Visual Genome dataset based on the common classes between the two, and hence, fail to capture the true prior distribution of the object classes in MS-COCO. For the transformer-based detectors  DETR~\cite{nicolas2020detr} and Deformable DETR~\cite{zhu2020deformable} we also achieve a consistent improvement.

%
\begin{table*}
  \centering
  \footnotesize
  \setlength{\tabcolsep}{4.2pt}        
  \renewcommand{\arraystretch}{1}      
  \newcommand{\grspace}{\hskip 3.0em}  
  \begin{tabular}{ll@{\grspace}cccccc@{\grspace}cccccc}
    \toprule
    & Method & AP$\uparrow$ & $\text{AP}_{\text{50}}\!\!\uparrow$ & $\text{AP}_{\text{75}}\!\!\uparrow$ & $\text{AP}_{\text{S}}\!\!\uparrow$ & $\text{AP}_{\text{M}}\!\!\uparrow$ & $\text{AP}_{\text{L}}\!\!\uparrow$ &
    $\text{AR}_{\text{1}}\!\!\uparrow$ & $\text{AR}_{\text{10}}\!\!\uparrow$ & $\text{AR}_{\text{100}}\!\!\uparrow$ & $\text{AR}_{\text{S}}\!\!\uparrow$ & $\text{AR}_{\text{M}}\!\!\uparrow$ & $\text{AR}_{\text{L}}\!\!\uparrow$ \\
    \midrule
    \multirow{11}{*}{\rotatebox{90}{\hspace{-1em}$\text{VG}_{1000}$}}
    & Light-head RCNN~\cite{li2017light}    & 6.2 & 10.9 & 6.2 & 2.8 & 6.5 & 9.8 & 14.6 & 18.0 & 18.7 & 7.2 & 17.1 & 25.3 \\
    & Cascade RCNN~\cite{cai2018cascade}    & 6.5 & 12.1 & 6.1 & 2.4 & 6.9 & 11.2 & 15.3 & 19.4 & 19.5 & 6.1 & 19.2 &\rd 27.5\\
    & R-RCNN~\cite{xu2019reasoning}         &\nd 8.2 &\rd 13.3 &\nd 8.5 &\nd 4.4 &\fs 8.9 &\nd 12.9 &\rd 16.4 &\rd 22.2 & 22.5 &\rd 12.3 &\rd 22.1 & 27.1\\
    & SGRN~\cite{xu2019spatial}             &\rd 8.1 &\nd 13.6 &\rd 8.4 &\nd 4.4 &\rd 8.2 &\rd 12.8 &\nd 19.5 &\nd 26.0 & 26.2 &\nd 12.4 &\nd 23.9 &\nd 34.0\\
    & U-RCNN-intra~\cite{xu2020universal}   & 7.9 &\fs 13.7 & 8.3 & - & - & - & - & - & - & - & - & -\\
    & \color{gray} U-RCNN-inter ~\cite{xu2020universal}   & \color{gray} 8.8 & \color{gray} 14.2 & \color{gray} 9.3 & \color{gray} 5.0 & \color{gray} 9.4 & \color{gray} 13.3 & \color{gray} 17.5 & \color{gray} 23.7 & \color{gray} 24.0 & \color{gray} 13.4 & \color{gray} 23.3 & \color{gray} 28.8\\
    & F-RCNN~\cite{lin2017feature}          & 7.1 & 12.7 & 7.2 & 3.9 & 7.6 & 11.1 & 14.8 & 19.7 & 19.9 & 10.6 & 18.8 & 24.9\\
    \cdashlinelr{2-14}
    & F-RCNN*~\cite{lin2017feature}         & 7.3 & 12.7 & 7.6 &\rd 4.2 & 7.8 & 11.0 & 15.1 & 20.1 & 20.3 & 11.2 & 19.4 & 24.3\\
    & \textit{\textbf{This paper}} with F-RCNN*  &\fs 8.5 &\fs 13.7 &\fs 8.8 &\fs 4.8 &\nd 8.8 &\fs 13.4 &\fs 19.8 &\fs 27.2 &\fs 27.5 &\fs 14.1 &\fs 25.9 & \fs 34.2\\
    \cdashlinelr{2-14}
    & DETR*~\cite{nicolas2020detr}          & 5.8 & 9.9 & 5.7 & 2.1 & 5.6 & 10.6 & 12.1 & 17.4 & 18.0 & 6.6 & 15.8 & 26.1\\
    & \textit{\textbf{This paper}} with DETR*    & 6.8 & 11.2  & 6.8 & 2.7 & 7.0 & 12.1 & 13.8 & 20.0 & 20.6 & 8.3 & 18.2 & 28.9\\
    \midrule
    \midrule
    \multirow{7}{*}{\rotatebox{90}{\hspace{-0em}$\text{VG}_{3000}$}}
    & Light-head RCNN~\cite{li2017light}    & 3.0 & 5.1 & 3.2 & 1.7 & 4.0 & 5.8 & 7.3 & 9.0 & 9.0 & 4.3 & 10.3 & 15.4\\
    & Cascade RCNN~\cite{cai2018cascade}    &\rd 3.8 & 6.5 & 3.4 & 1.9 & 4.8 & 4.9 & 7.1 & 8.5 & 8.6 & 4.2 & 9.9 & 13.7\\
    & R-RCNN~\cite{xu2019reasoning}         &\nd 4.3 &\rd 6.9 &\nd 4.6 &\fs 3.2 &\nd 6.0 &\rd 7.9 &\rd 8.5 &\rd 11.1 &\rd 11.2 &\nd 8.3 &\rd 13.7 &\rd 16.2\\
    & SGRN~\cite{xu2019spatial}             &\fs 4.5 &\fs 7.4 &\rd 4.3 &\rd 2.9 &\nd 6.0 &\fs 8.6 &\fs 10.8 &\nd 13.7 &\nd 13.8 &\rd 8.1 &\nd 15.1 & \fs 21.8\\
    & F-RCNN~\cite{lin2017feature}          & 3.7 & 6.5 & 3.7 & 2.1 &\rd 4.9 & 6.8 & 7.6 & 9.8 & 9.9 & 6.8 & 11.8 & 14.6 \\
    \cdashlinelr{2-14}
    & F-RCNN*~\cite{lin2017feature}          & 3.3 & 5.8 & 3.4 & 2.4 & 4.6 & 5.9 & 6.8 & 8.9 & 9.0 & 6.5 & 11.1 & 13.0\\
    & \textit{\textbf{This paper}} with F-RCNN*    & \fs 4.5 &\fs 7.4 &\fs 4.7 &\nd 3.1 &\fs 6.1 &\nd 8.4 &\nd 10.6 &\fs 14.0 & \fs 14.1 & \fs 8.8 & \fs 16.2 &\nd 21.5\\
    \bottomrule
  \end{tabular}
  \vspace*{0.1mm}
  \caption{\textbf{Comparison with others on $\text{VG}_{1000}$ and $\text{VG}_{3000}$~\cite{krishna2017visual}.} We report the results of our context-likelihood energy-refined graph on top of F-RCNN and DETR, where we applied context-likelihood energy-refinement graph. We compare with alternative methods modeling object interrelationships which are trained with similar backbone and dataset settings. Methods denoted with an asterisk like F-RCNN* refer to results we obtained by running the original code with default parameters, while the other numbers are the reported values. Rows in gray use more datasets for training. We achieve competitive results for both datasets on almost all metrics. Best results are highlighted as \colorbox{colorFst}{\bf first}, \colorbox{colorSnd}{second}, and \colorbox{colorTrd}{third}.
  }
  \label{tab:VG_results}
\end{table*}

\begin{figure*}[t]
  \vspace{-7pt}
  \centering
  \footnotesize
  \setlength{\tabcolsep}{1.5pt}
  \newcommand{\sz}{0.184}
  \begin{tabular}{ccccc}
    \includegraphics[width=\sz\linewidth]{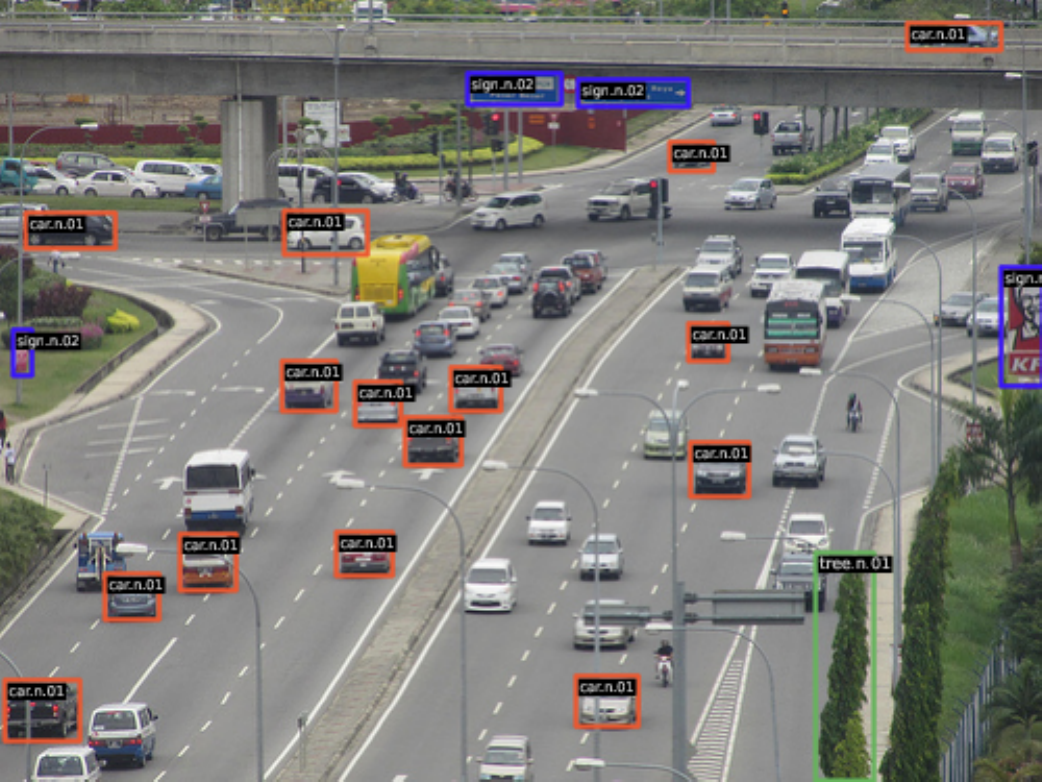} &
    \includegraphics[width=\sz\linewidth]{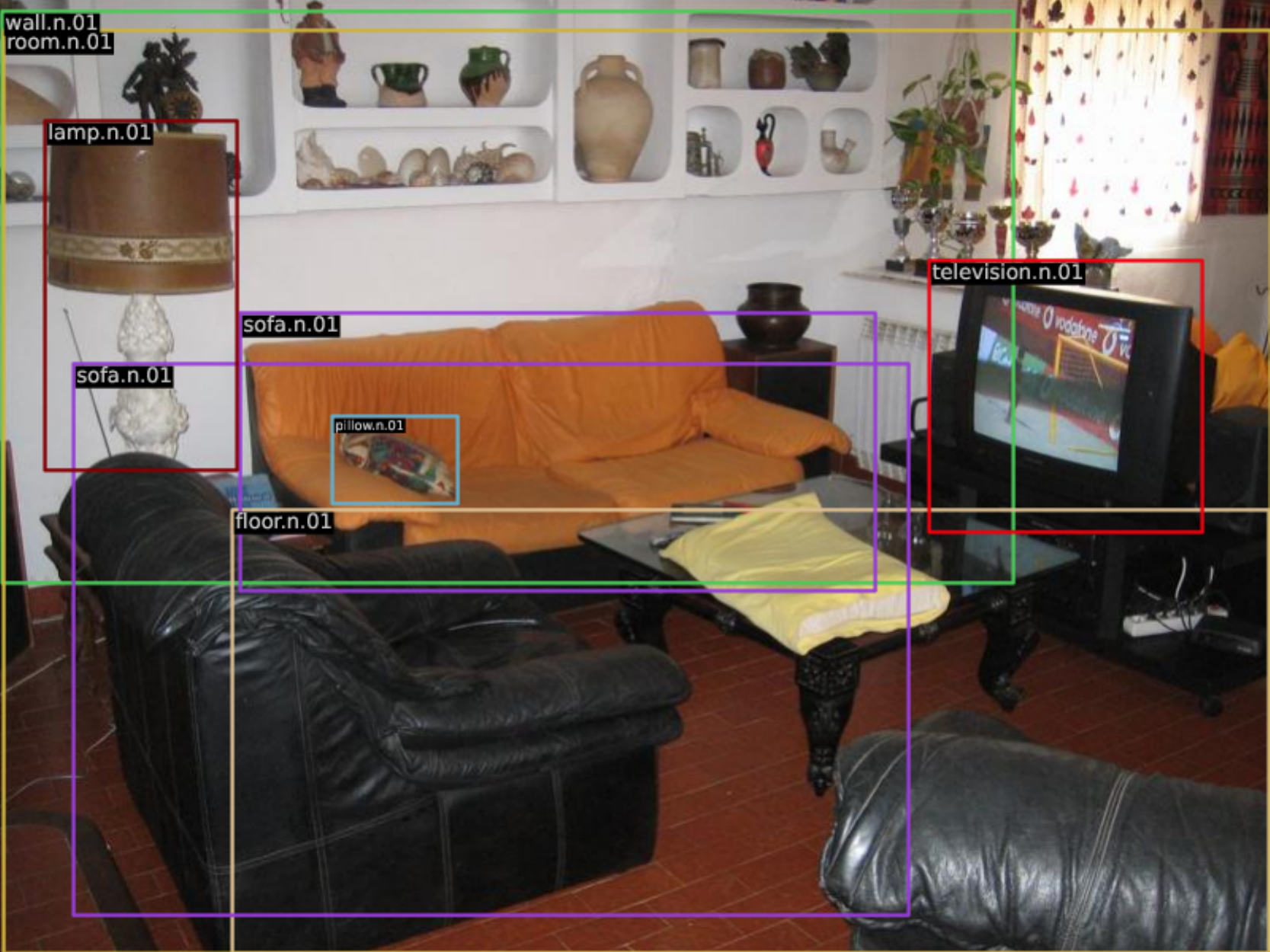} &
    \includegraphics[width=\sz\linewidth]{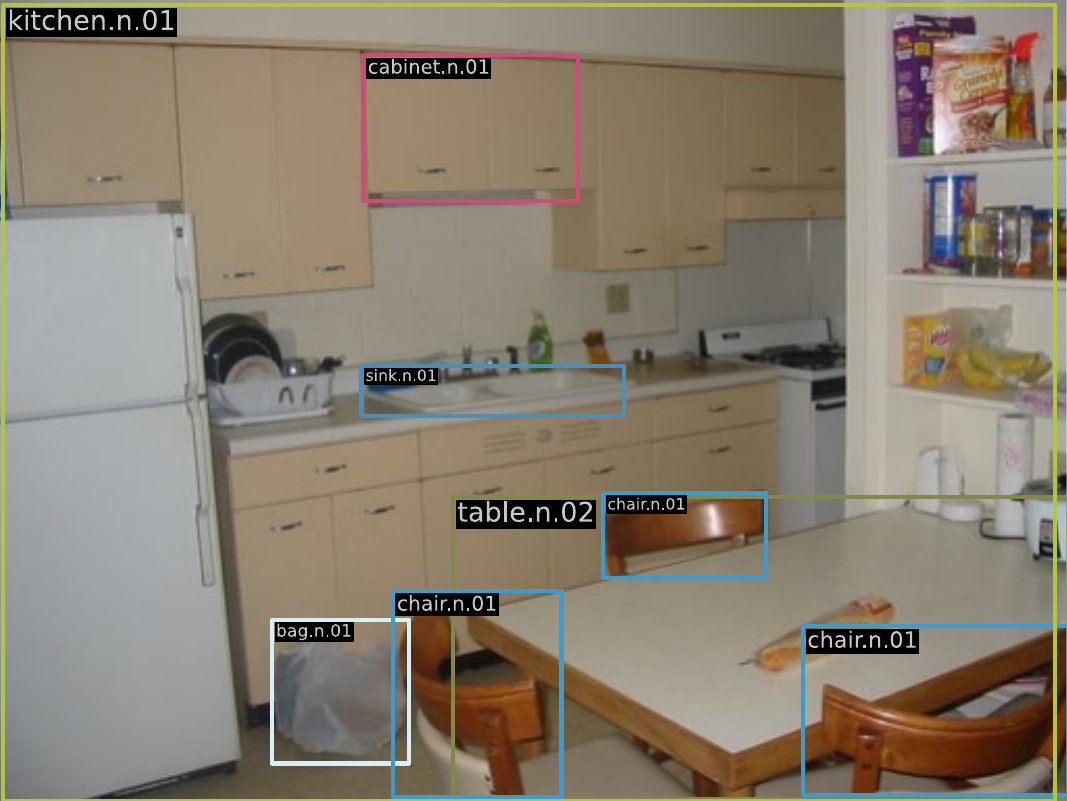} &
    \includegraphics[width=\sz\linewidth]{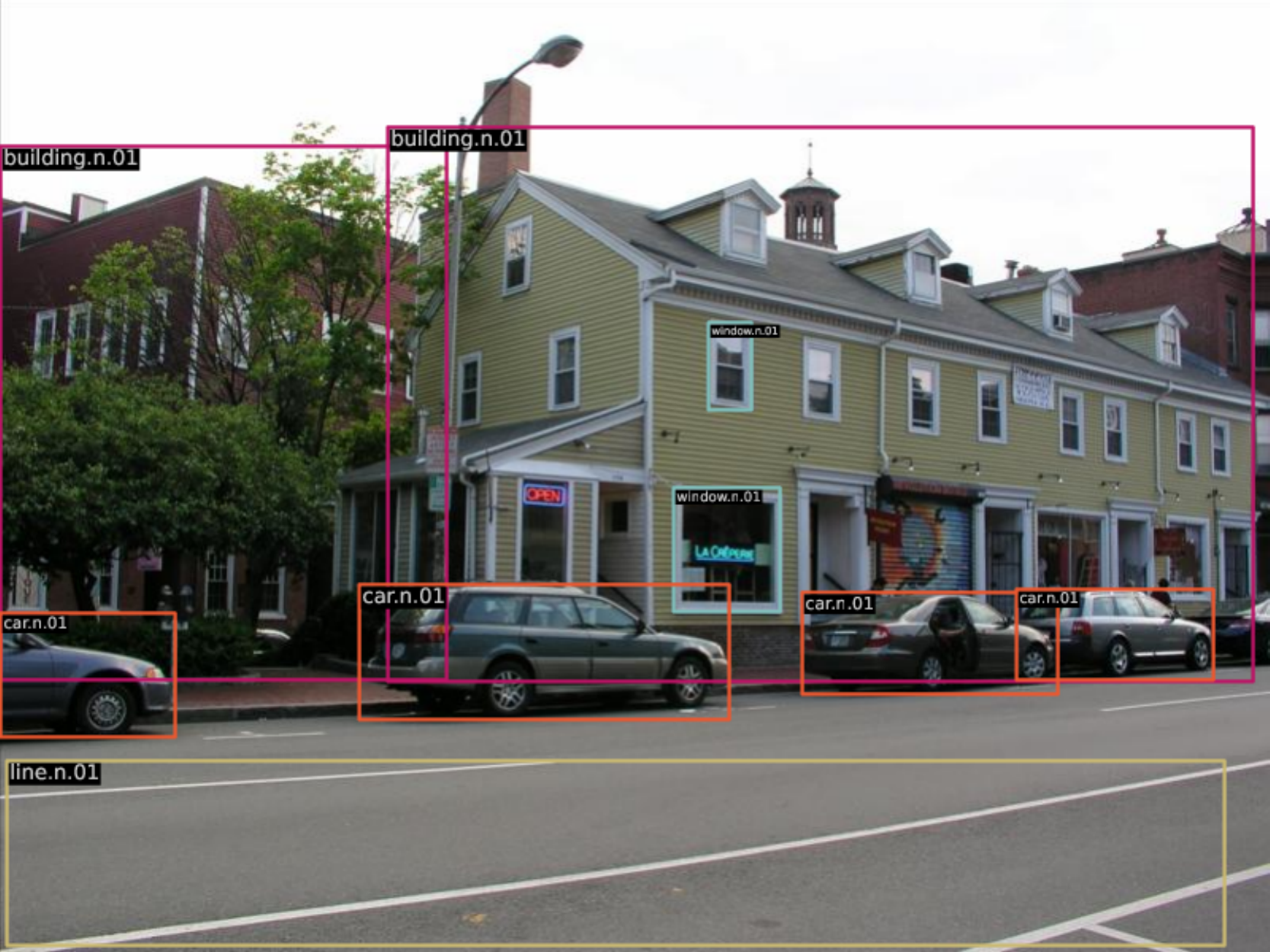} & 
    \includegraphics[width=0.22\linewidth]{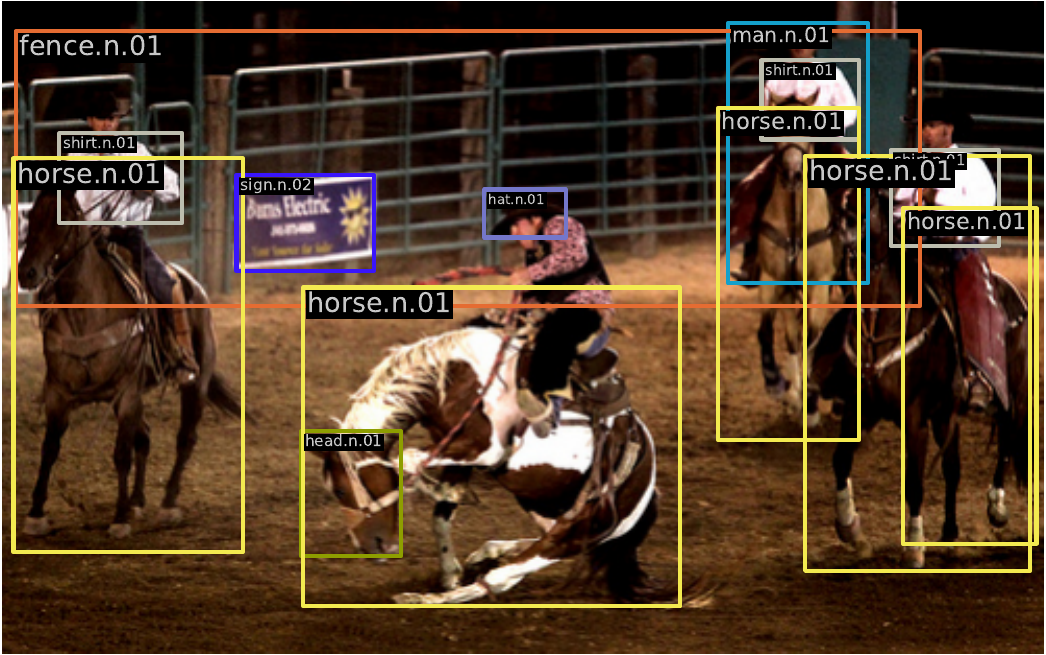} \\
    \includegraphics[width=\sz\linewidth]{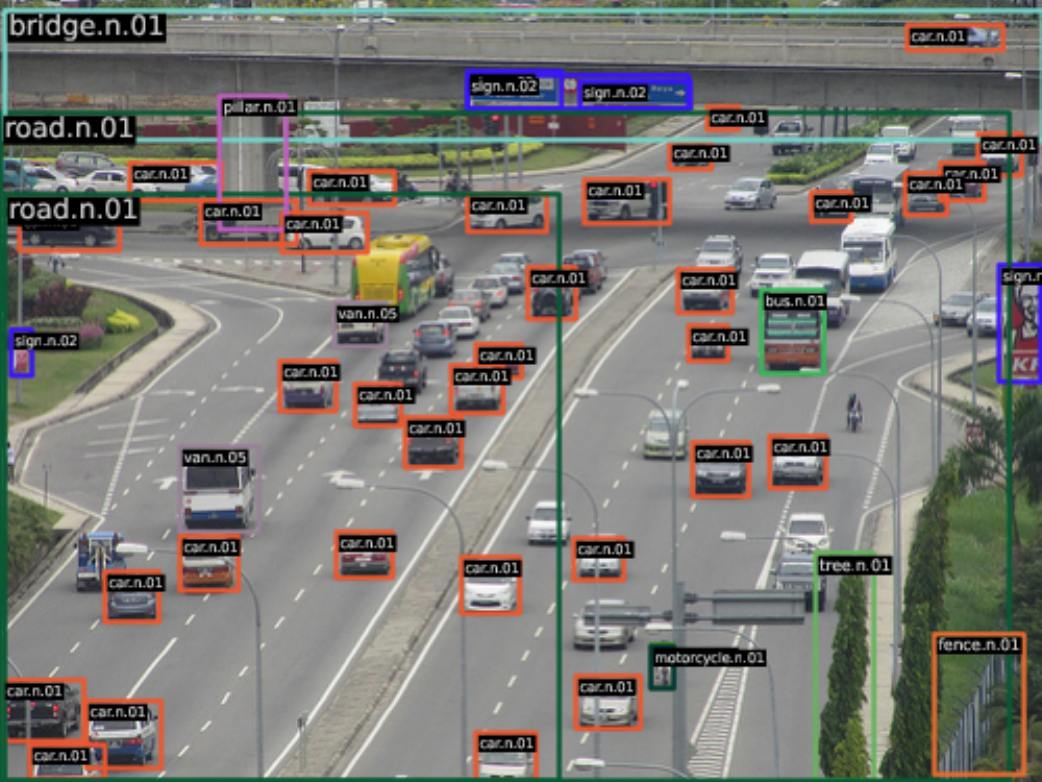} &
    \includegraphics[width=\sz\linewidth]{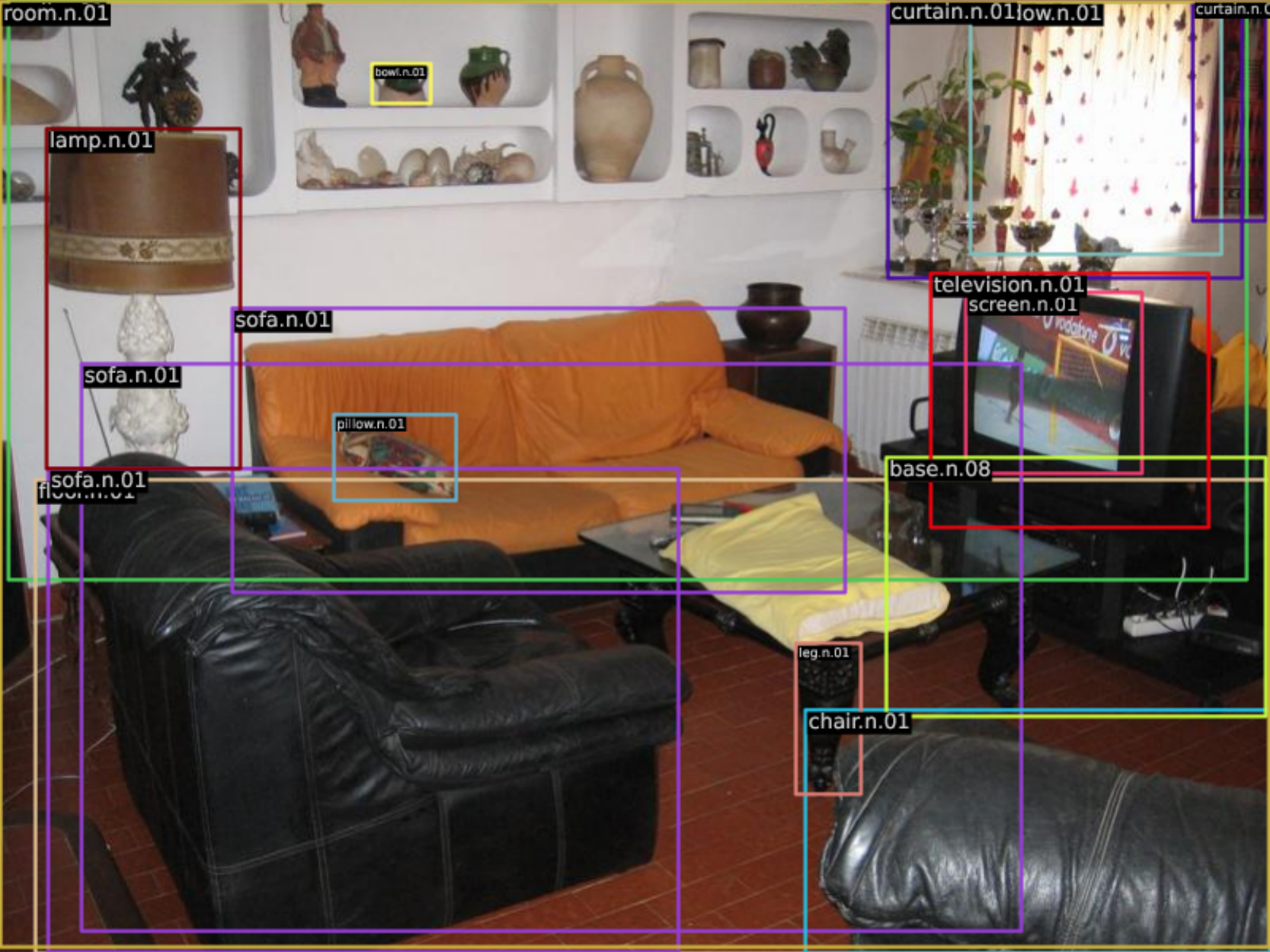} &
    \includegraphics[width=\sz\linewidth]{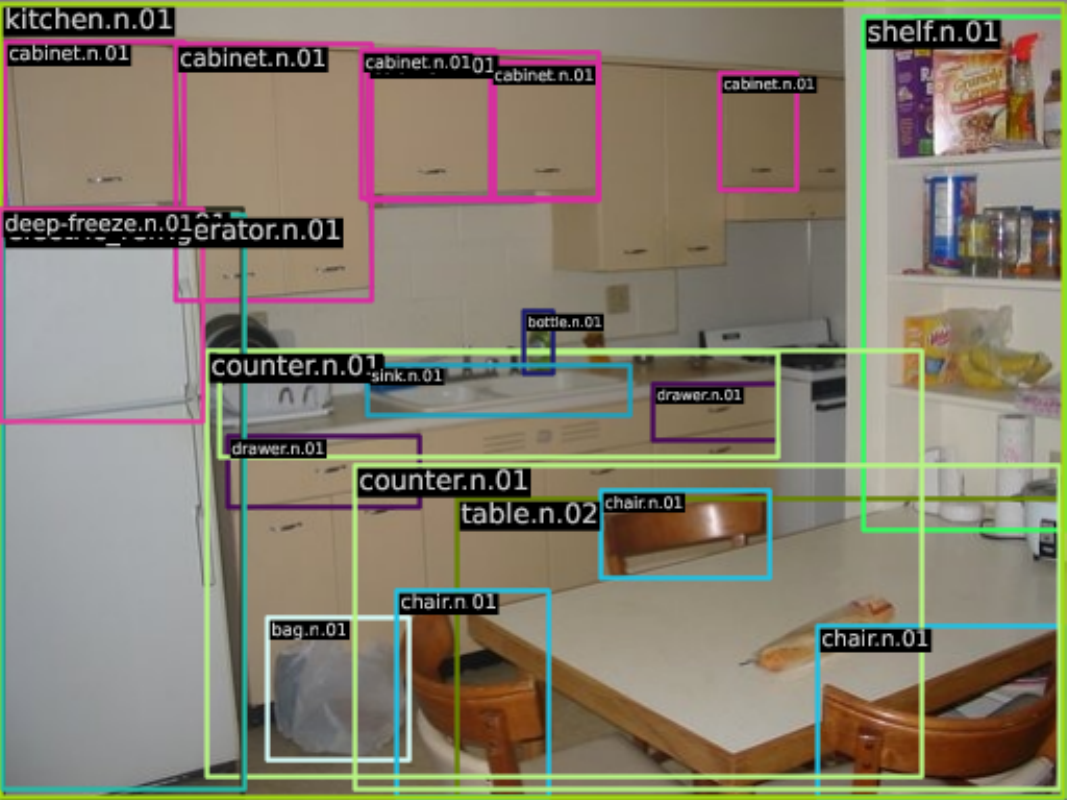} &
    \includegraphics[width=\sz\linewidth]{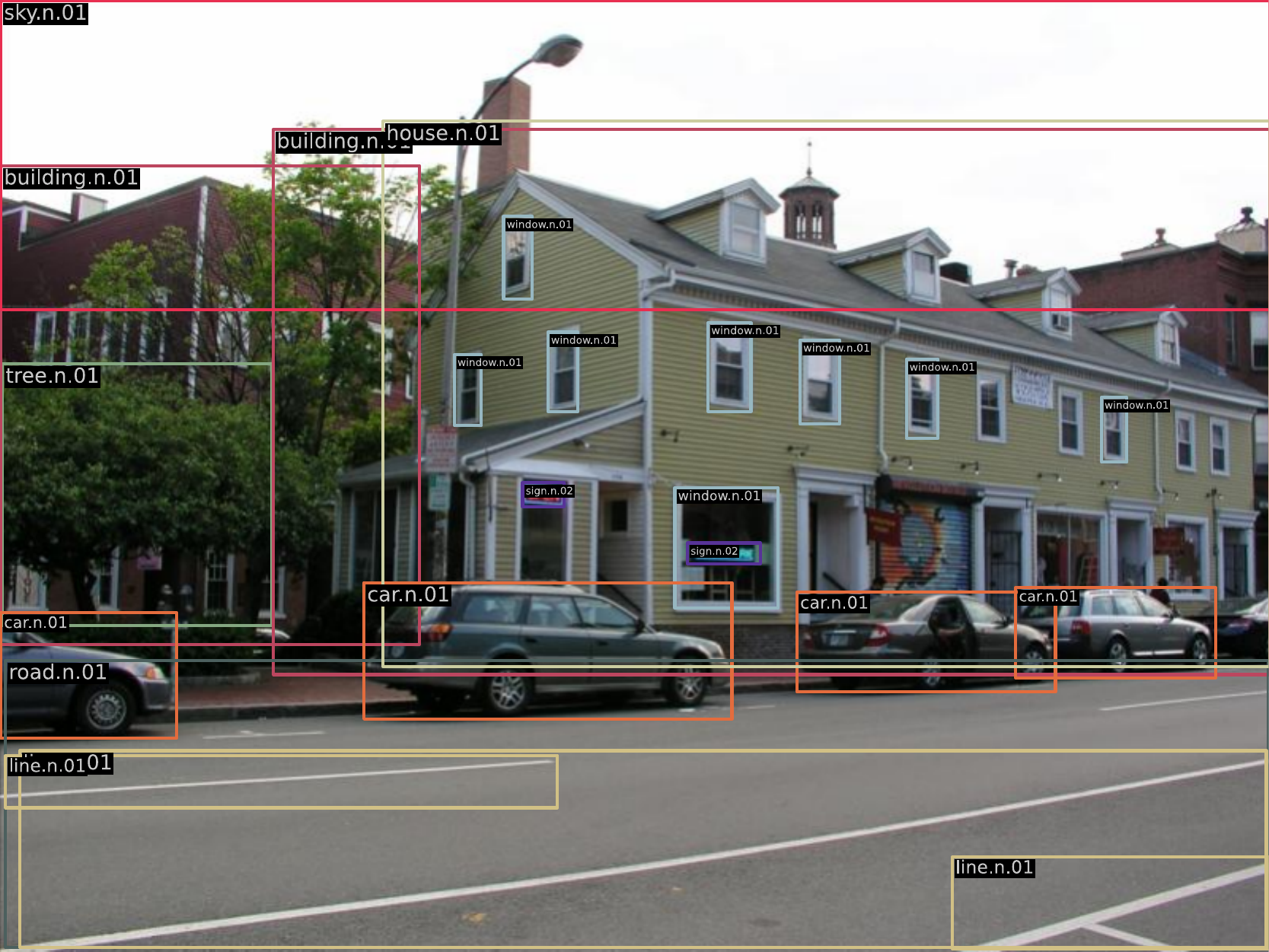} &
    \includegraphics[width=0.22\linewidth]{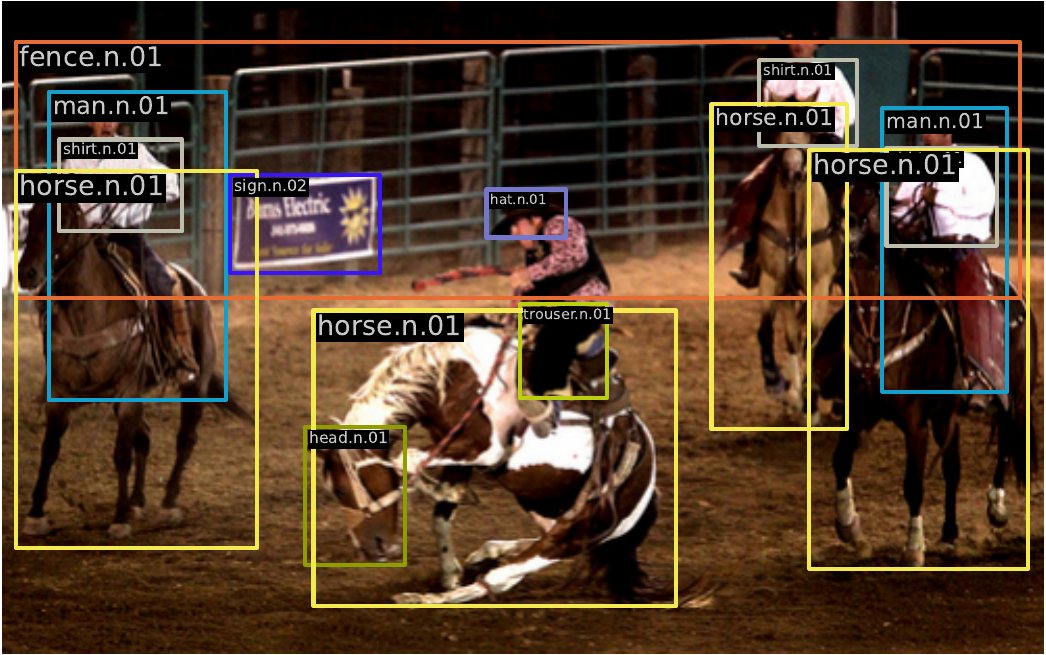} \\
  \end{tabular}
  \caption{\textbf{Qualitative results} from the $\text{VG}_{1000}$ test set~\cite{krishna2017visual} for a F-RCNN* object detector (top) and our results (bottom). We observe positive effects of exploiting contextual-likelihood graphs with energy-based refinement. Especially for repetitive occurrence of objects like the cars, cabinet doors or windows, the information that other such objects exist in the scene helps detection. Baseline performance can be poor, as is evident by  failure to detect many cars in the first top-left image, our approach is also beneficial for such challenging cases.
  }
  \label{fig:qualitative}
  \vspace{-10pt}
\end{figure*}

\begin{table}
  \centering
  \footnotesize
  \setlength{\tabcolsep}{2pt}          
  \renewcommand{\arraystretch}{1}      
  \newcommand{\grspace}{\hskip 0.7em}  
  \begin{tabular}{ll@{\grspace}cccccc}
    \toprule
    & Method & AP$\uparrow$ & $\text{AP}_{\text{50}}\!\!\uparrow$ & $\text{AP}_{\text{75}}\!\!\uparrow$ & $\text{AR}_{\text{1}}\uparrow$ & $\text{AR}_{\text{10}}\!\!\uparrow$ & $\text{AR}_{\text{100}}\!\!\uparrow$  \\
    %
    \midrule
    %
    & HKRM~\cite{nicolas2020detr}       &    37.8 &    58.0 & 41.3 & - & - & - \\
    & SGRN~\cite{xu2019reasoning}       &\rd 41.7 &\fs 62.3 &\nd 45.5 & - & - & - \\
    & R-RCNN~\cite{xu2019reasoning}     &\fs 42.9 &    -    & - & - & - & - \\
    \cdashlinelr{2-8}
    & F-RCNN~\cite{ren2015faster_rcnn}  &    38.3 &\rd 60.5 &\rd 42.0 &\nd 32.2 &\nd 50.7 &\nd 53.0 \\
    & \textit{\textbf{This paper}} w F-RCNN     &\nd 42.5 &\nd 62.1 &\fs 45.8 &\fs 34.0 &\fs 53.4 &\fs 55.9 \\
    & DETR~\cite{nicolas2020detr}  &    39.9 &\rd 60.4 &\rd 41.7 &\nd 32.2 &\nd 50.7 &\nd 53.0 \\
    & \textit{\textbf{This paper}} w DETR     &\nd 40.6 &\nd 61.2 &\fs 42.1 &\fs 34.0 &\fs 53.4 &\fs 55.9 \\
    & Def-DETR~\cite{zhu2020deformable}  &    44.3 &\rd 63.2 &\rd 48.6 &\nd 32.2 &\nd 50.7 &\nd 53.0 \\
    & \textit{\textbf{This paper}} w Def-DETR     &\nd 44.7 &\nd 64.0 &\fs 49.0 &\fs 34.0 &\fs 53.4 &\fs 55.9 \\
    %
    %
    \bottomrule
  \end{tabular}
  \vspace*{0.1mm}
  \caption{\textbf{Comparisons on MS-COCO~\cite{lin2014microsoft}.} Our context-likelihood energy-refined graph performs comparable to alternative methods, with prior statistics transferred from the Visual Genome dataset, demonstrating the generalizability.
  }
  \label{tab:coco}
  \vspace{-18pt}
\end{table}

\boldparagraph{Limitations.}
Our approach cannot create a detection if a corresponding region proposal is missing. However, we observe in practice 512 proposals are enough to capture small instances, initially classified incorrectly, and later predicted correctly via our graph refinement. Although our method is optimized for region-based algorithms, adapting it to single-stage detectors like SSD or Yolo requires only minor adjustments in the training process. For example, for Yolo v3, we use the objectness scores to select candidate nodes above a 0.5 threshold, form edges based on initial class predictions, and update features with enriched information from the context-likelihood graph for classification.



\section{Conclusion}  \label{sec:conclusion}
%
We demonstrated a new approach to exploit inter-object relations for object detection by generating a context-likelihood graph based on initial object predictions which leads to improved detector performance.
Second, we show the potential of this context-likelihood graph via an oracle experiment. Finally, we learnt the joint distribution of objects and their relations using an energy-based model, enabling us to sample a refined graph representation of an image. We further demonstrated that refinement of the graph representation at test time yields further improvements, outperforming alternative methods on the challenging $\text{VG}_{1000}$ and $\text{VG}_{3000}$ datasets.

\section*{Acknowledgements}
\noindent
This work has been financially supported by TomTom, the University of Amsterdam and the allowance of Top consortia for Knowledge and Innovation (TKIs) from the Netherlands Ministry of Economic Affairs and Climate Policy.

{\small
\bibliographystyle{ieee_fullname}
\bibliography{PaperForReview}
}

\newpage
\clearpage
\appendix

\section{Message Passing}  \label{sec:method_details}

This section provides further details about the process to obtain enhanced features and a description of the corresponding message passing.

We represent an input image $I$ with visual features $\{F_i\}_1^N$ of dimension $D$ and the edge connectivity matrix $\edgeSet \in R^{N \times N}$, where $N$ is the number of proposals.
In order to propagate this information, we use a form of graph attention network~\cite{velivckovic2017graph}, taking inspiration from \cite{gong2019exploiting}.

We create the enhanced feature vector $F_{i}^{l}$ corresponding to the $i$-th node and $l$-th layer by aggregating node (object) and edge (relationship) information in its neighborhood.
Using feature matrix notations, the message passing algorithm can be written as follows: 
\begin{equation}
     F^l = \sigma [\alpha^{l}(F^{l-1},\edgeSet^{l-1})g^l(F^{l-1})]  \enspace.
\end{equation}
Here $\sigma$ is a non-linear activation; $g^l$ is a transformation which maps the node features from the input space to the output space, and is given by:
\begin{equation}
    g^l(F^{l-1}) = W^{l}F^{l-1} \enspace,
\end{equation}
where $W^{l}$ is the parameter matrix; $\alpha^{l}$ is the attention coefficients matrix given by $\alpha^{l} {=} \mathrm{DS}(\hat{\alpha}^{l})$, where $\mathrm{DS}$ represents doubly stochastic optimization and the specific entries $\hat{\alpha}^{l}_{ij}$ are given by 
\begin{equation}
    \hat{\alpha}_{ij}^{l} = f^{l}(X_{i}^{l-1},X_{j}^{l-1})E_{ij}^{l-1}  \enspace,
\end{equation}
where $E_{ij}^{l-1}$ refers to the i-th row and j-th column entry of the edge matrix in the $l-1$ layer and $f^l$ is the attention function given by 
\begin{equation}
    f^{l}(X_{i}^{l-1},X_{j}^{l-1}) = \exp{L(a^{T}[WX_{i}^{l-1}||WX_{j}^{l-1}])}  \enspace,
\end{equation}

where $L(\cdot)$ is the LeakyReLU activation function, $a$ is the attention coefficients vector, and $||$ denotes the concatenation operation.

The edge connectivity matrix $\edgeSet$ is symmetrical which gets updated after each layer of the attention update as $\edgeSet^l {=} \alpha^l$. 
We use two layers to get our final enhanced feature vectors $F^l$, and concatenate them to the original feature vectors which are used for the final classification and bounding box regression.

\section{Qualitative Results}
\begin{figure*}[t]
  \centering
  \footnotesize
  \setlength{\tabcolsep}{1.5pt}
  \newcommand{\sz}{0.132}
  \begin{tabular}{l ccccc}
    \rotatebox{90}{\hspace{1.7em}F-RCNN*~\cite{lin2017feature}} &
    \includegraphics[height=\sz\linewidth]{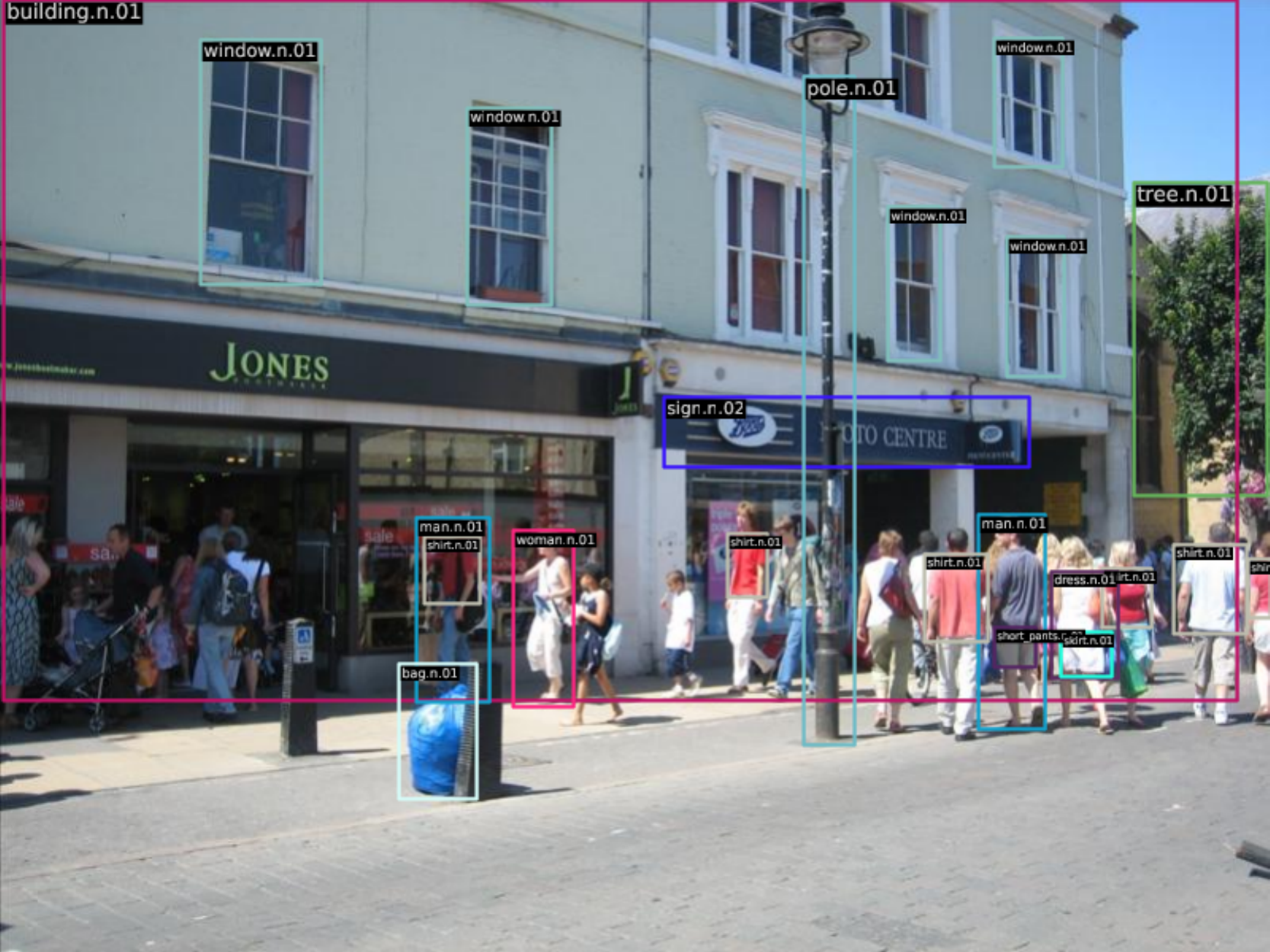} &
    \includegraphics[height=\sz\linewidth]{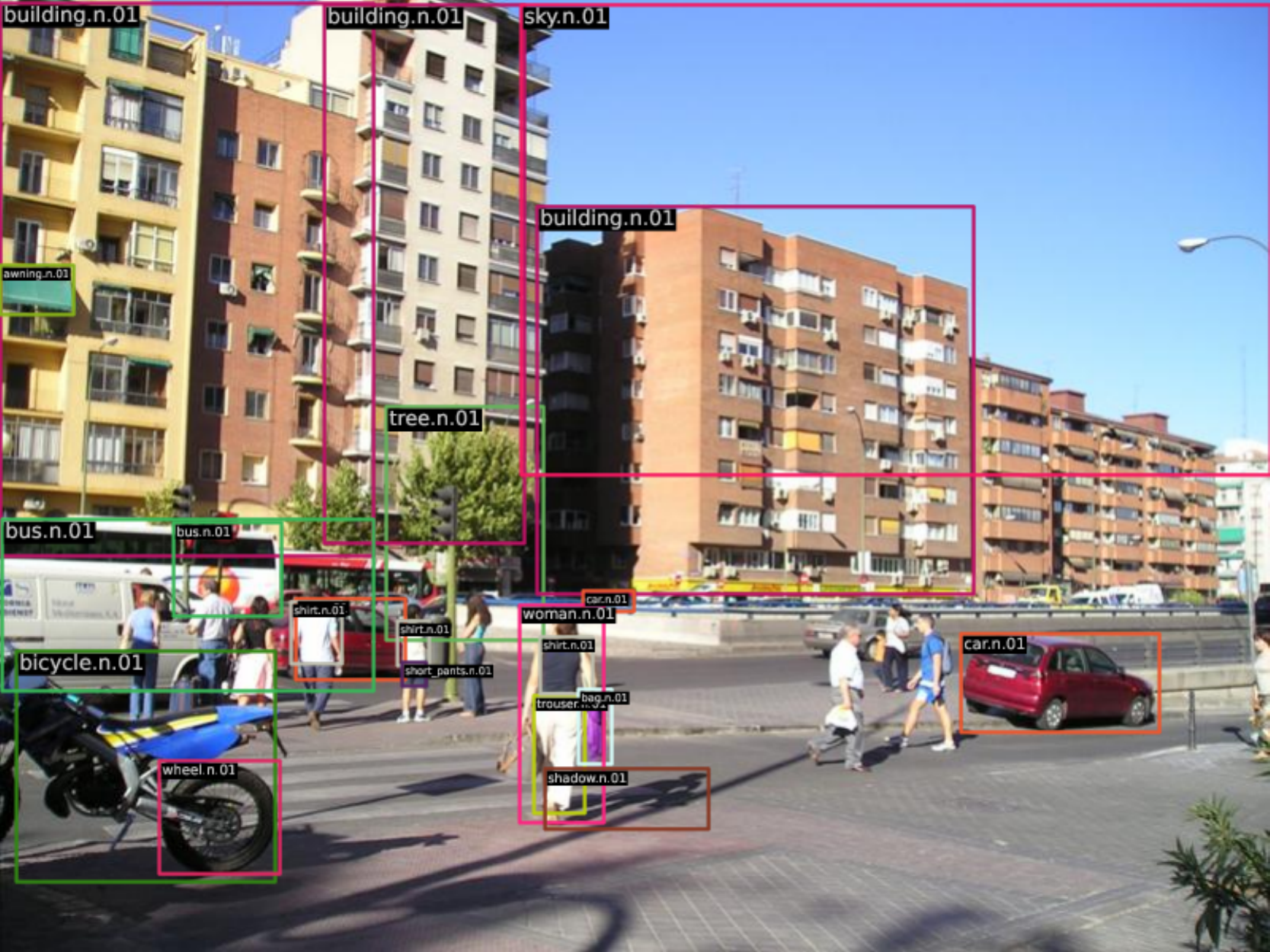} &
    \includegraphics[height=\sz\linewidth,width=0.195\linewidth]{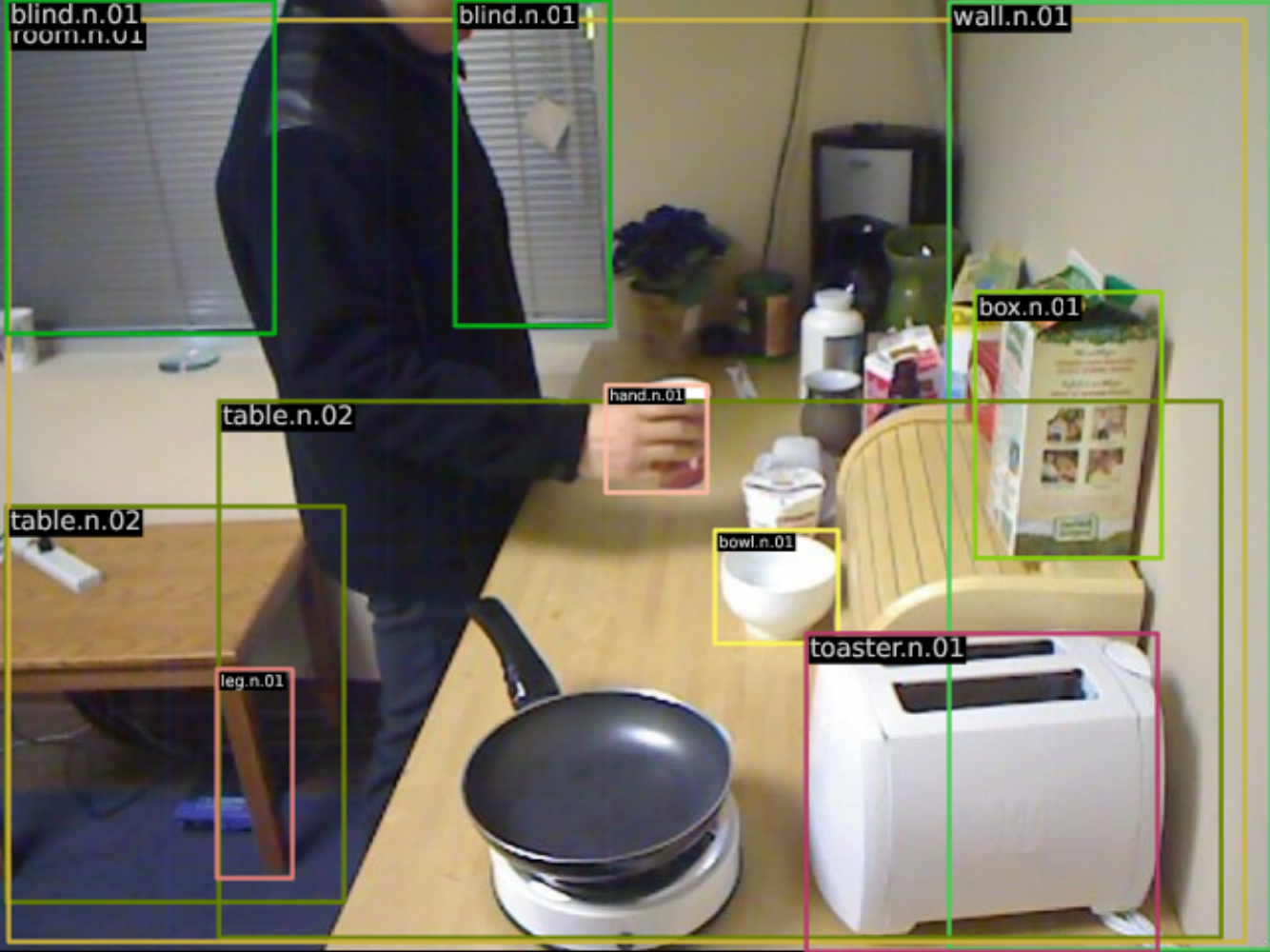} &
    \includegraphics[height=\sz\linewidth]{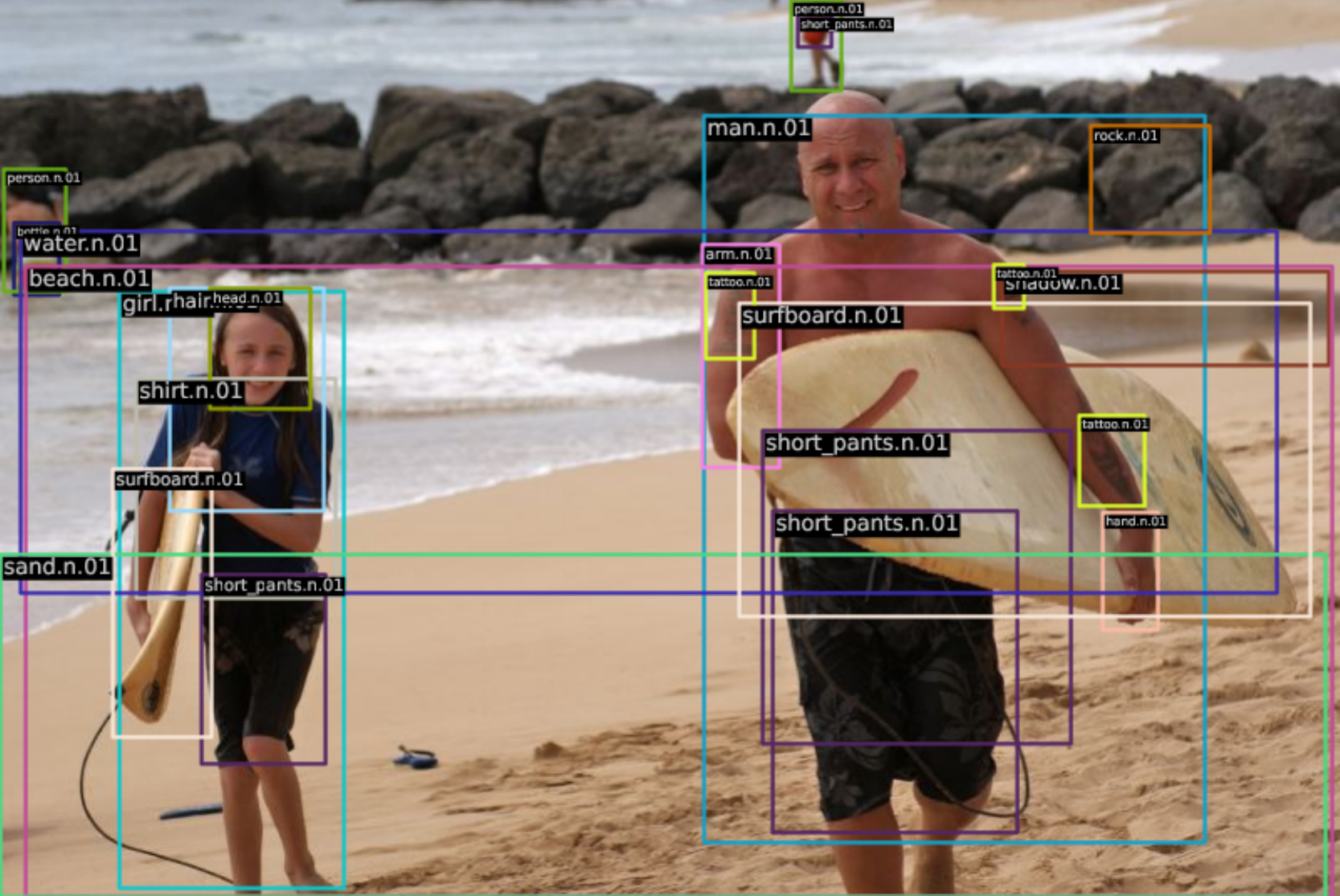} &
    \includegraphics[height=\sz\linewidth]{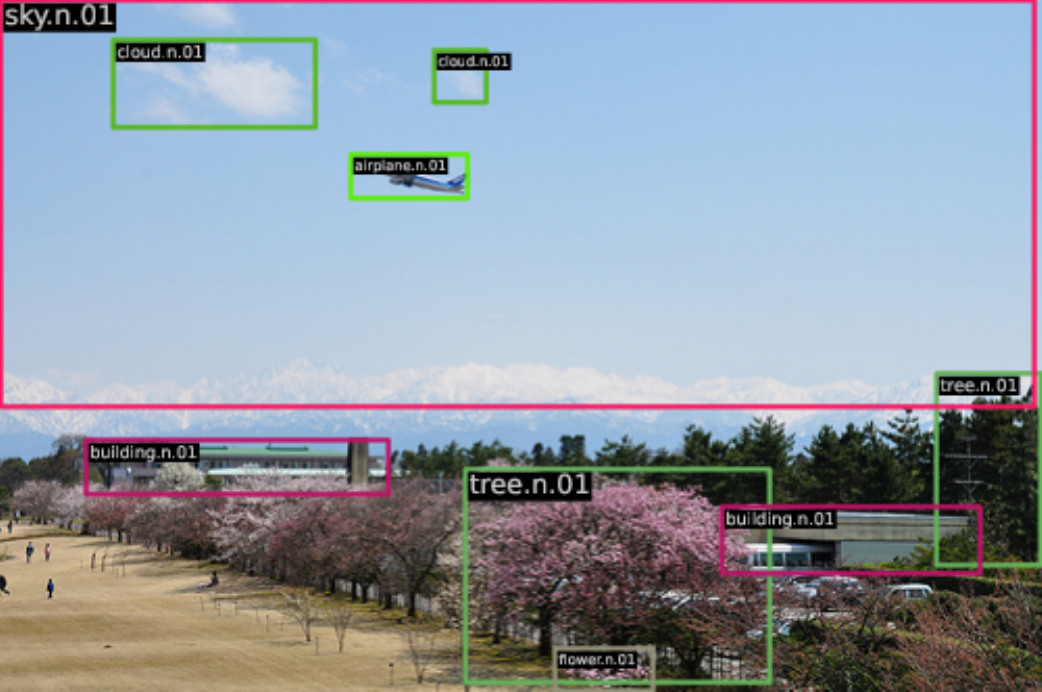} \\
    \rotatebox{90}{\hspace{0.5em}\textit{\textbf{This paper}}} &
    \includegraphics[height=\sz\linewidth]{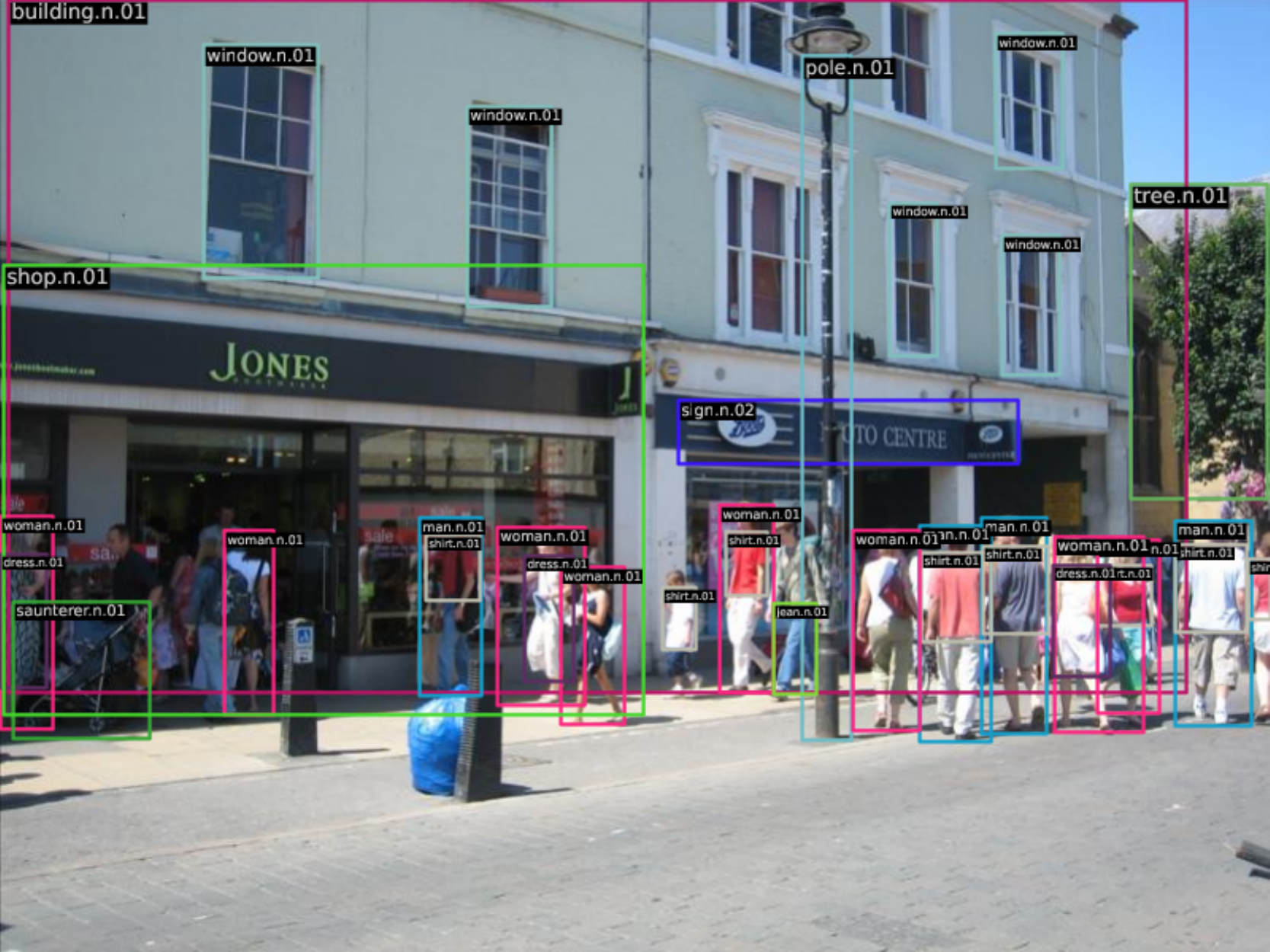} &
    \includegraphics[height=\sz\linewidth]{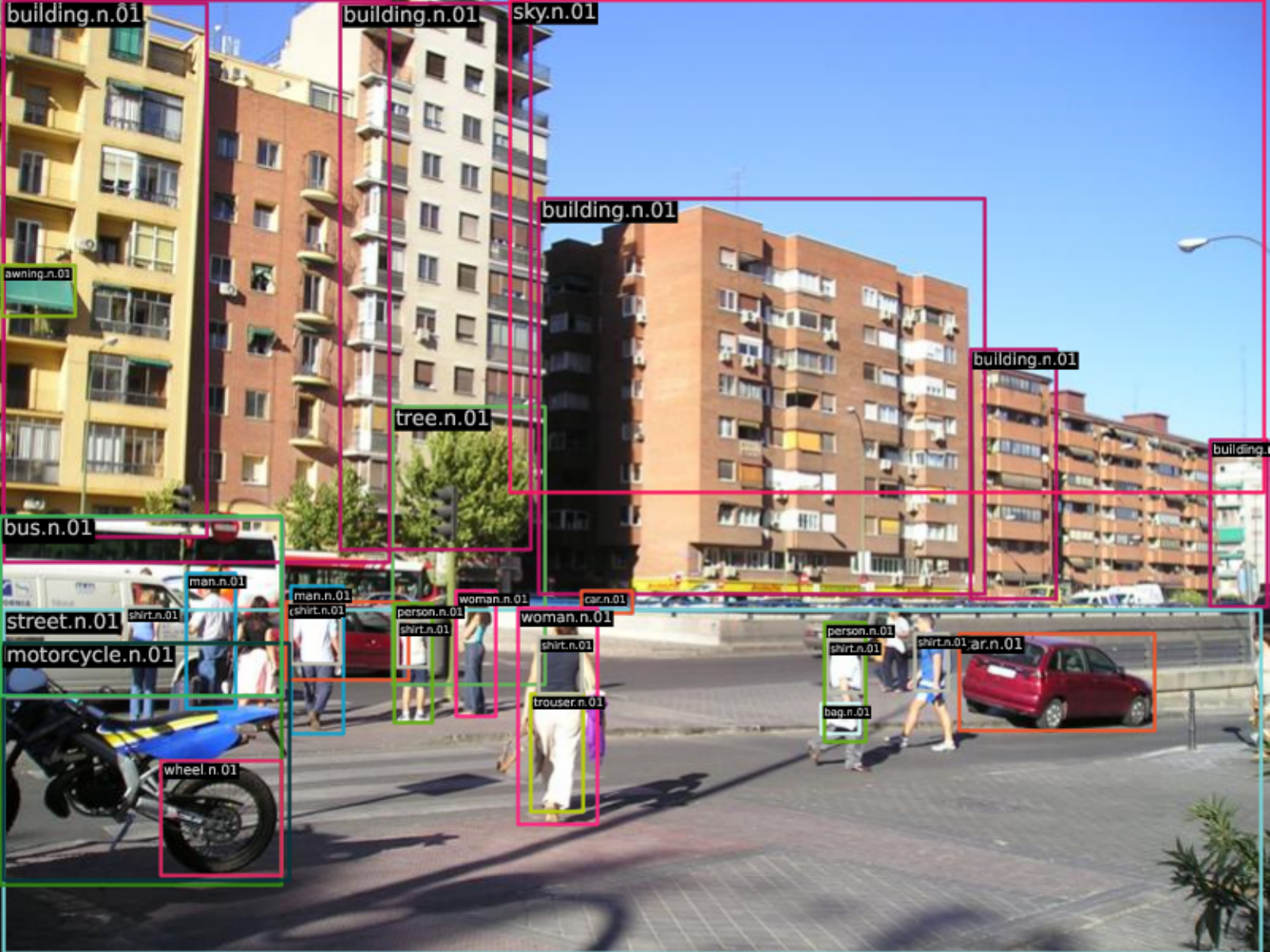} &
    \includegraphics[height=\sz\linewidth,width=0.195\linewidth]{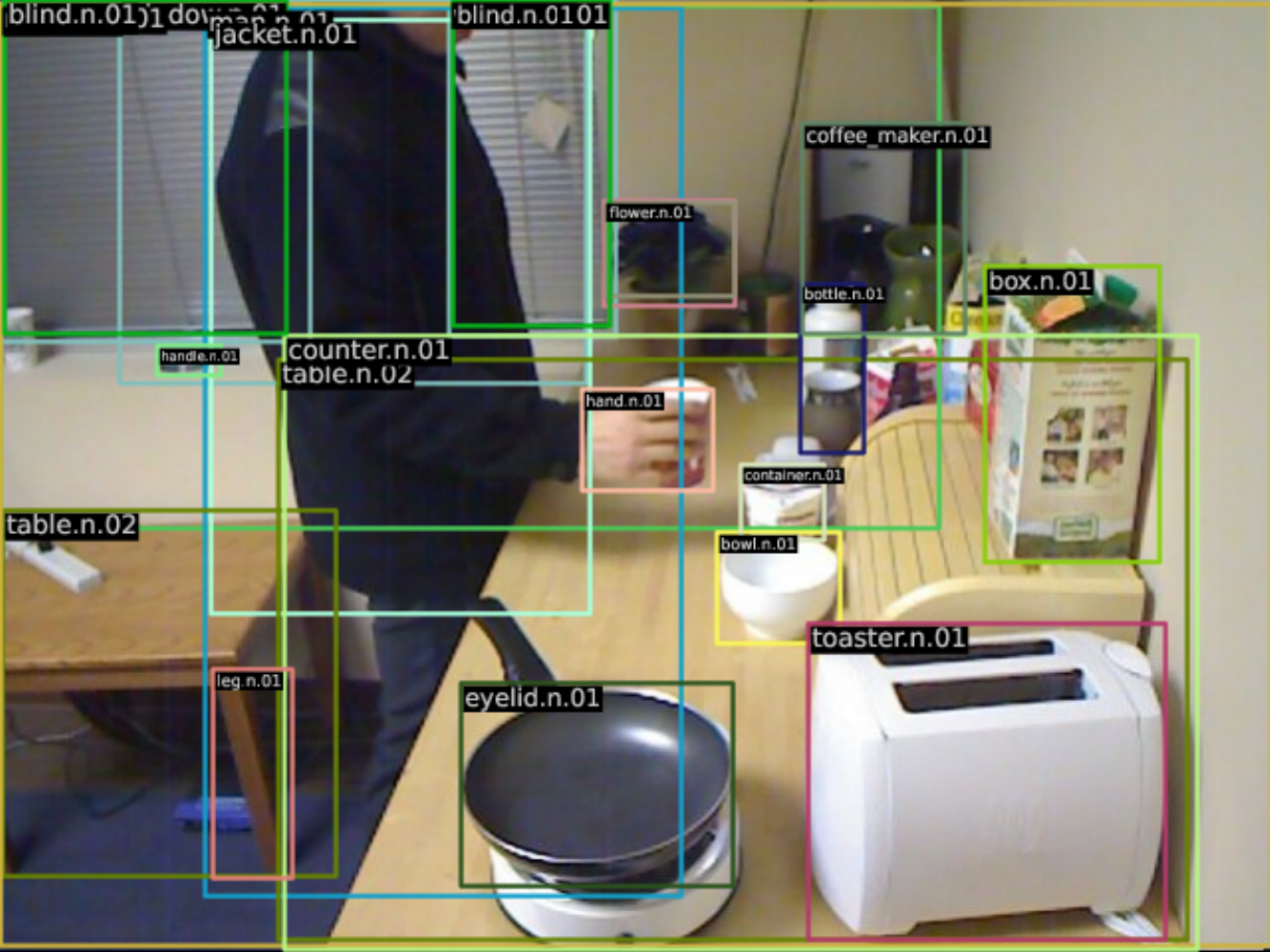} &
    \includegraphics[height=\sz\linewidth]{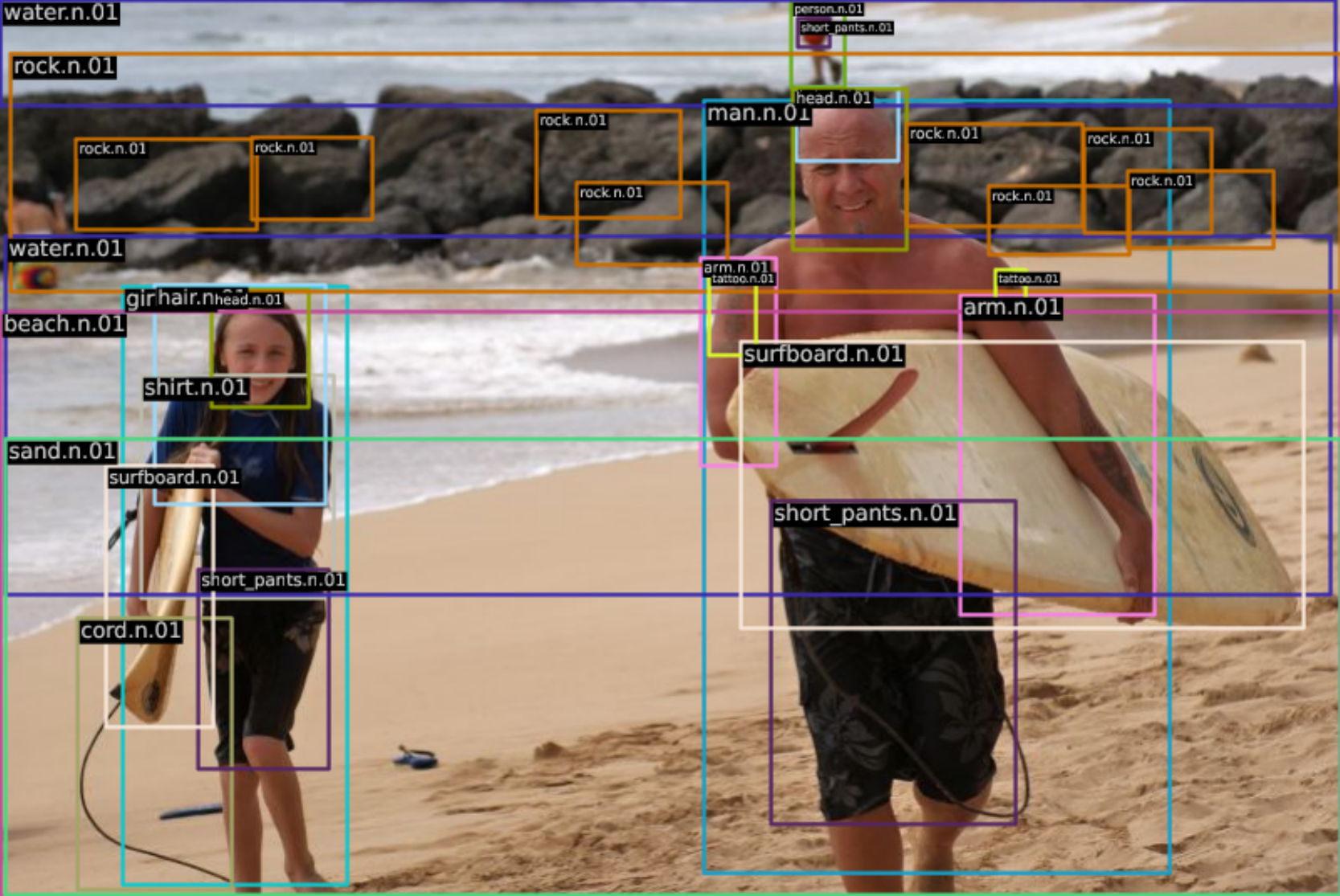} &
    \includegraphics[height=\sz\linewidth]{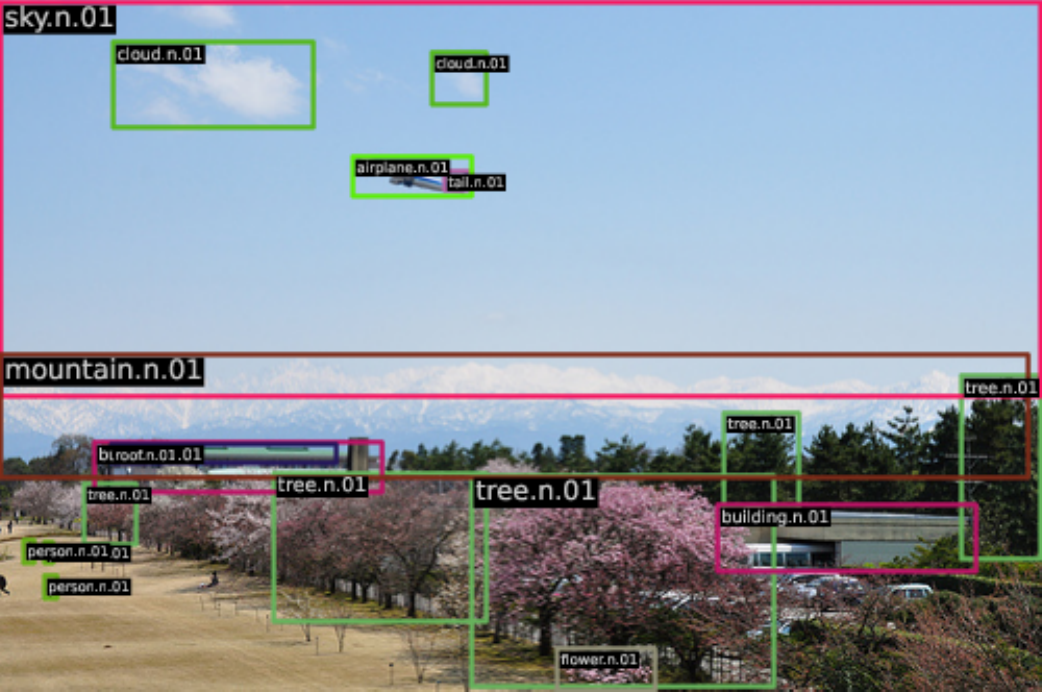} \\[8pt]
    \rotatebox{90}{\hspace{1.7em}F-RCNN*~\cite{lin2017feature}} &
    \includegraphics[height=\sz\linewidth]{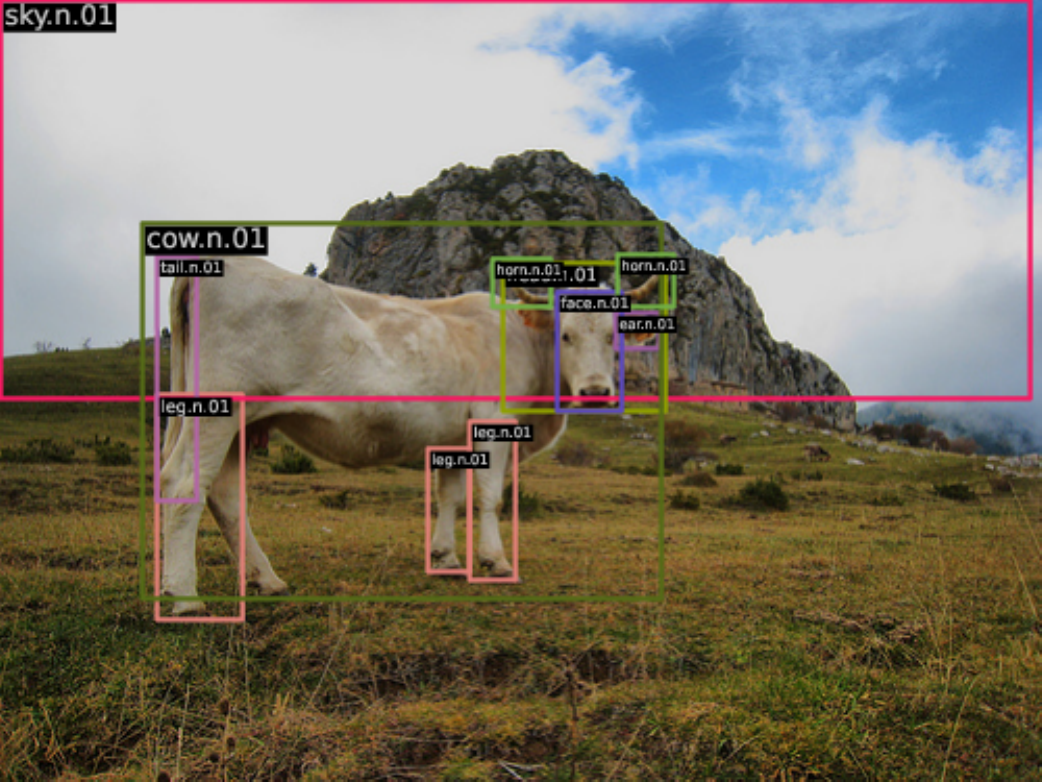} &
    \includegraphics[height=\sz\linewidth]{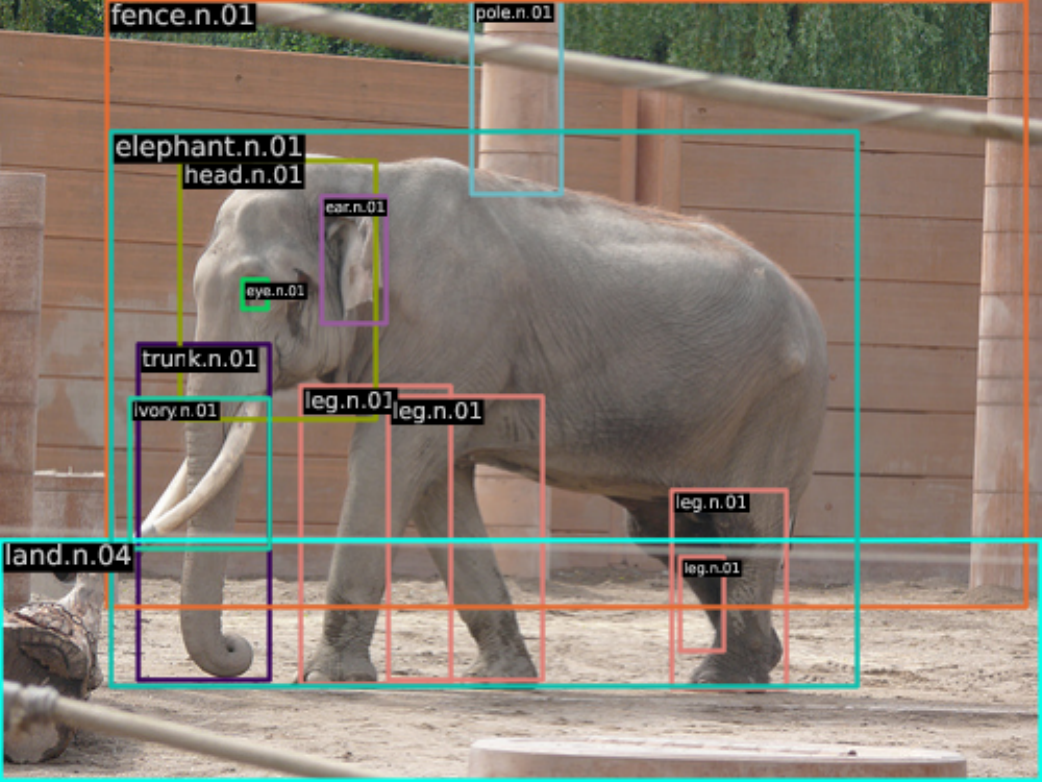} &
    \includegraphics[height=\sz\linewidth,width=0.195\linewidth]{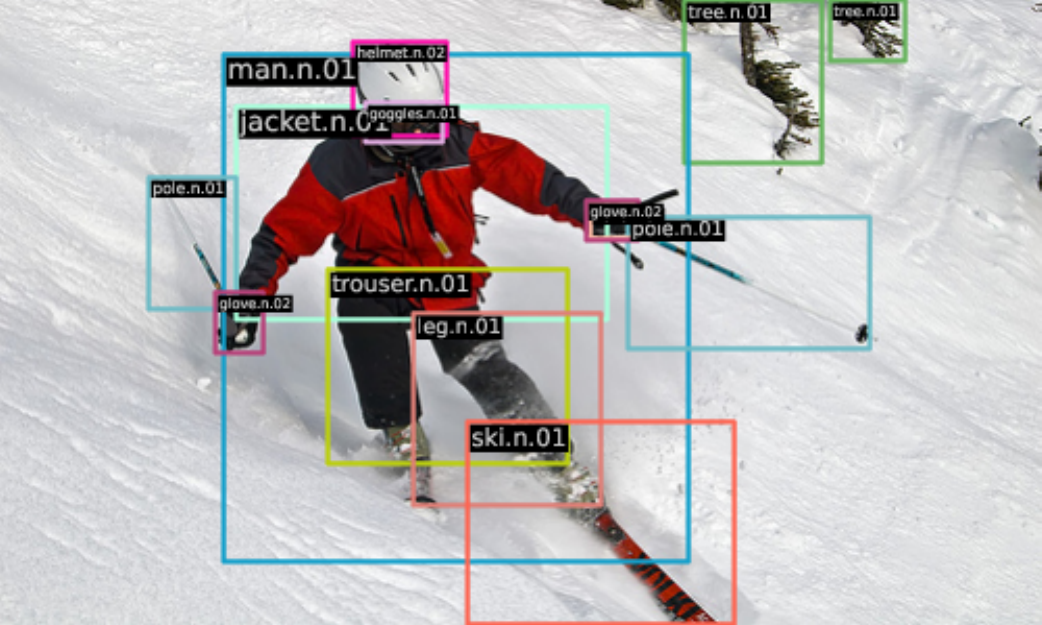} &
    \includegraphics[height=\sz\linewidth]{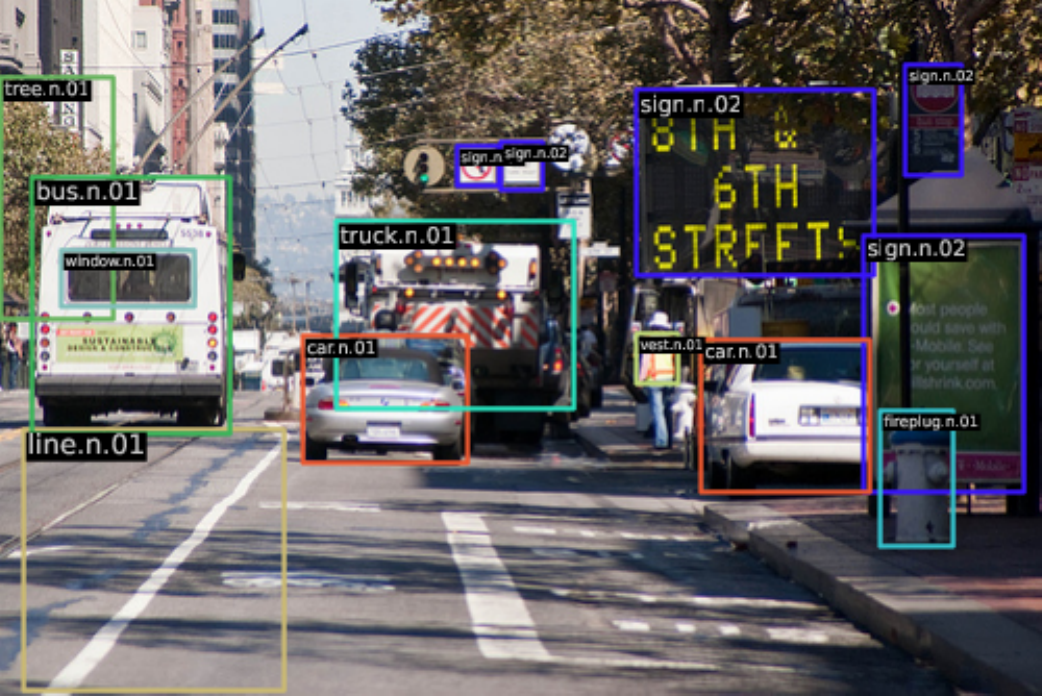} &
    \includegraphics[height=\sz\linewidth]{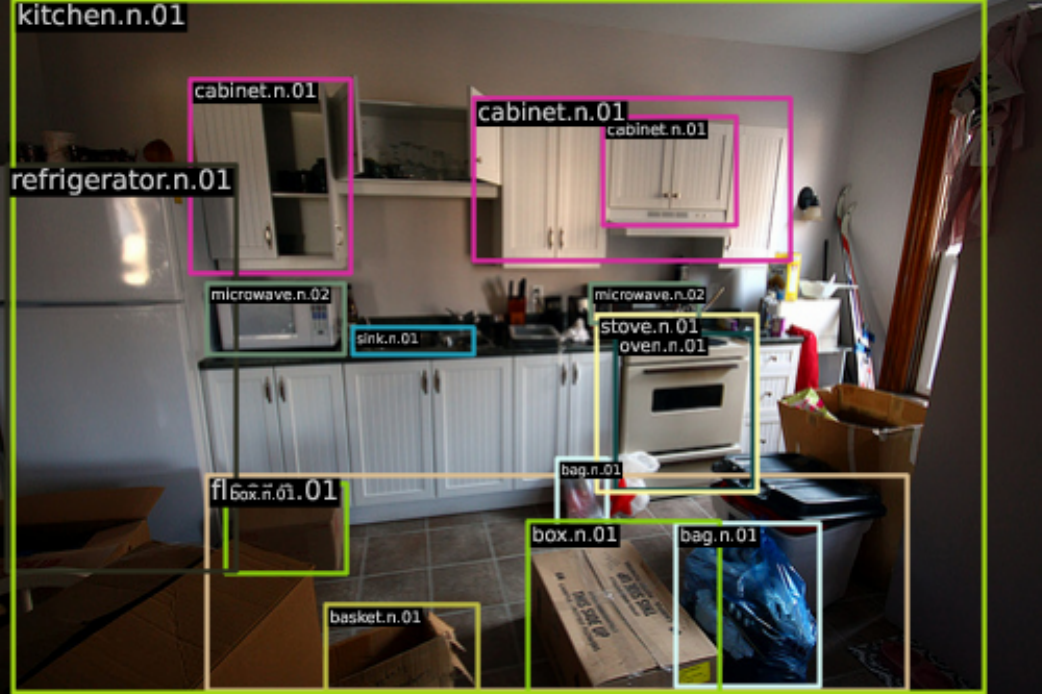} \\
    \rotatebox{90}{\hspace{0.5em}\textit{\textbf{This paper}}} &
    \includegraphics[height=\sz\linewidth]{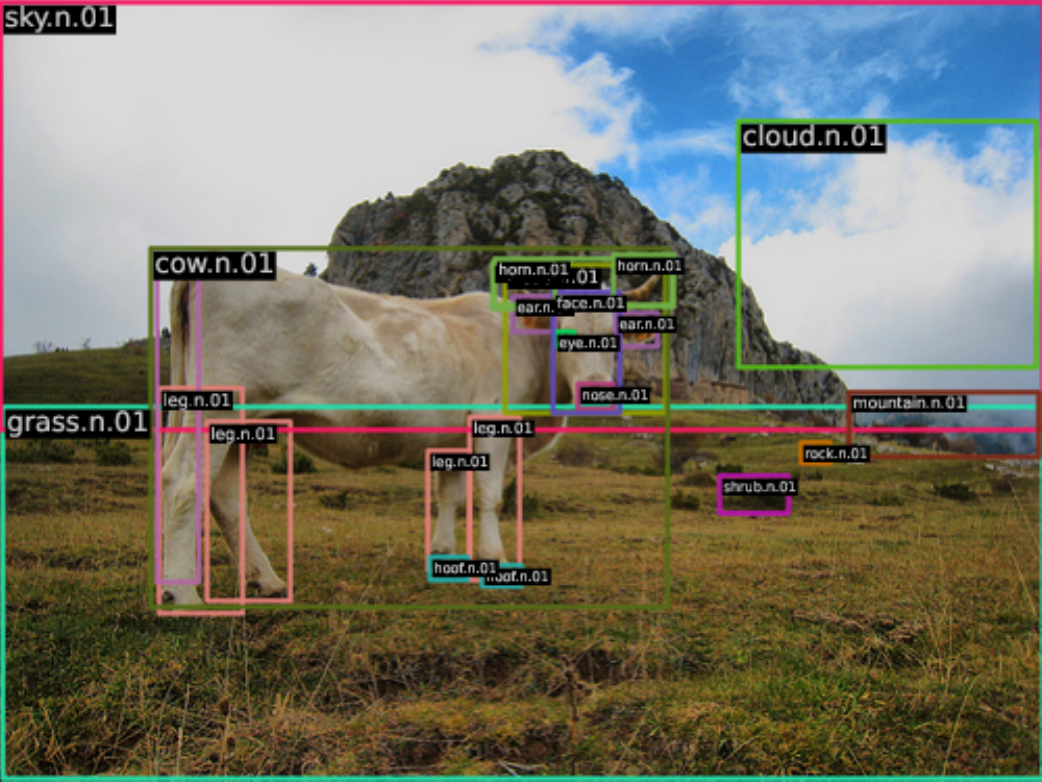} &
    \includegraphics[height=\sz\linewidth]{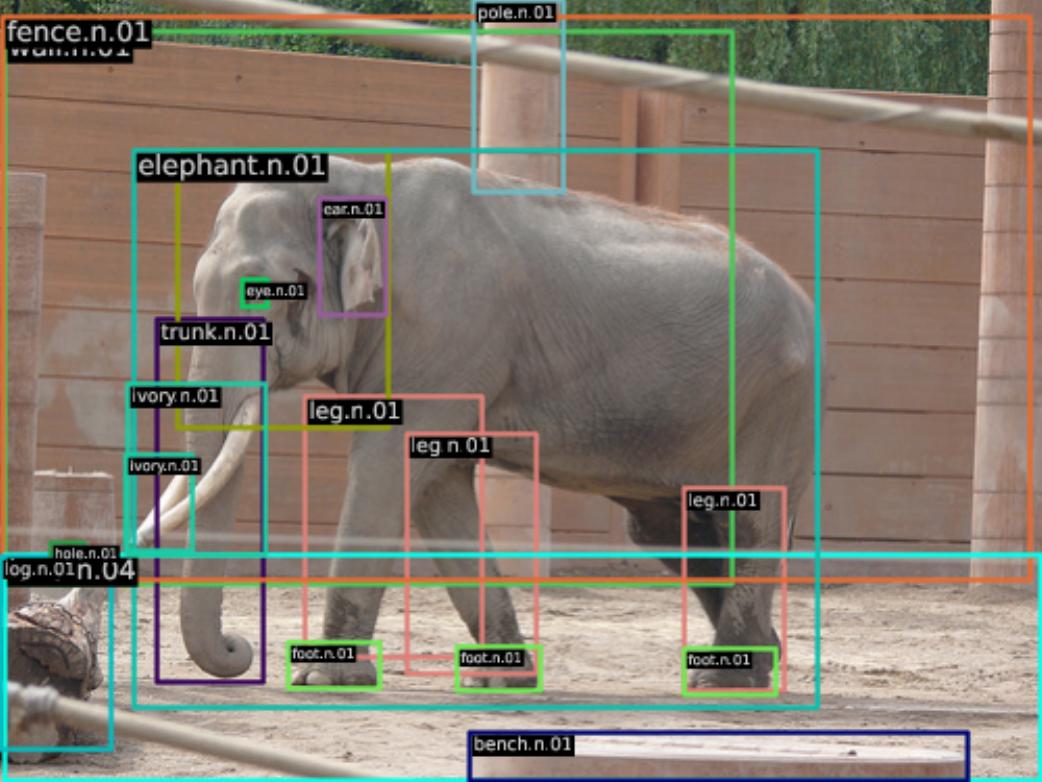} &
    \includegraphics[height=\sz\linewidth,width=0.195\linewidth]{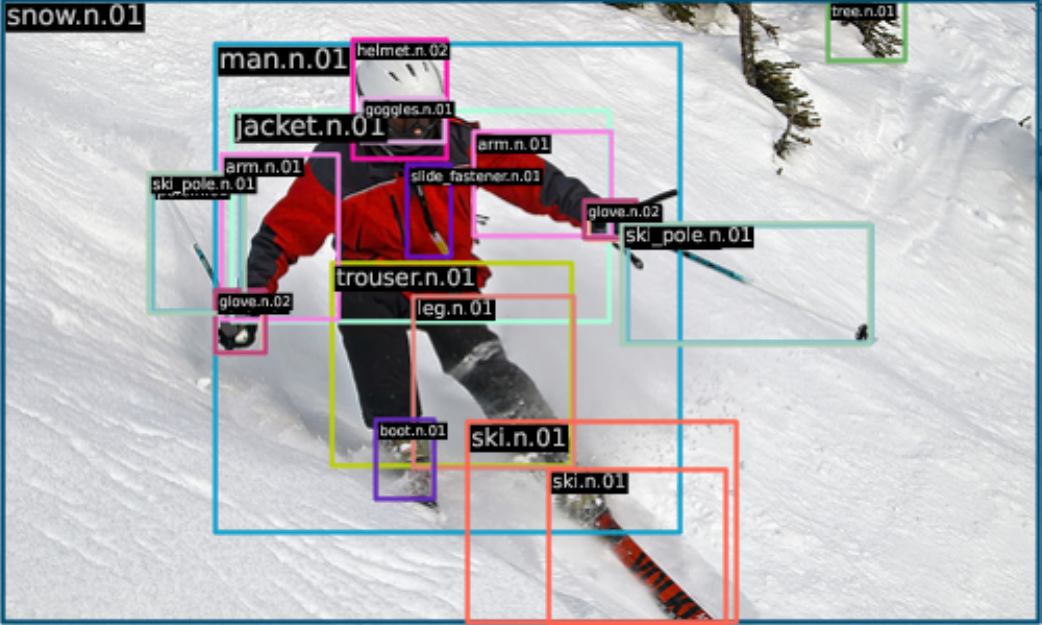} &
    \includegraphics[height=\sz\linewidth]{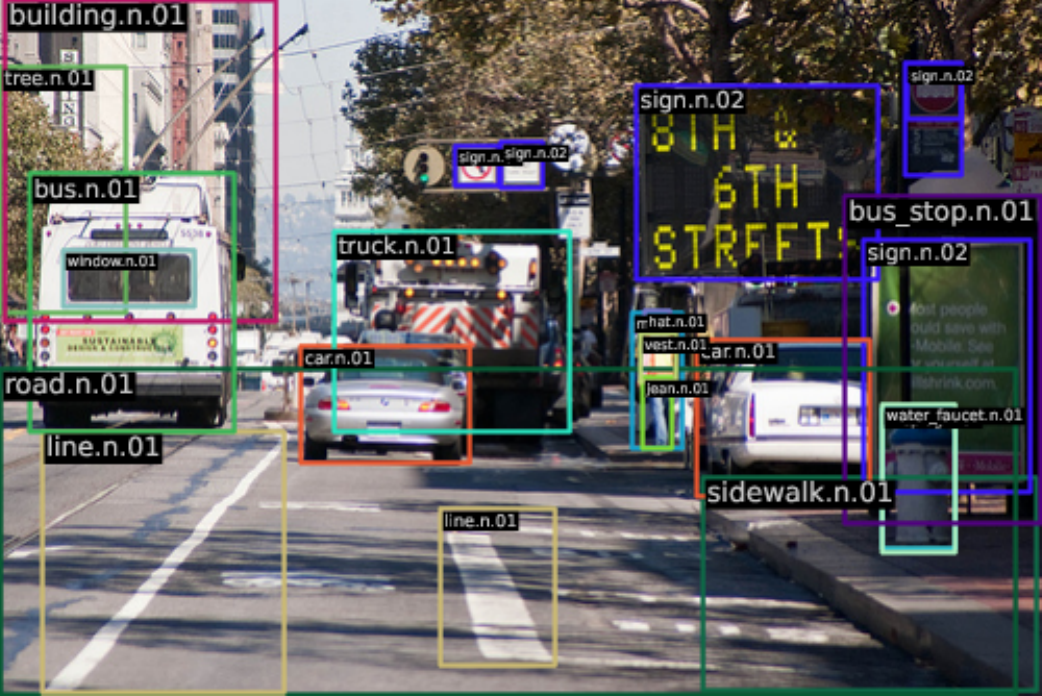} &
    \includegraphics[height=\sz\linewidth]{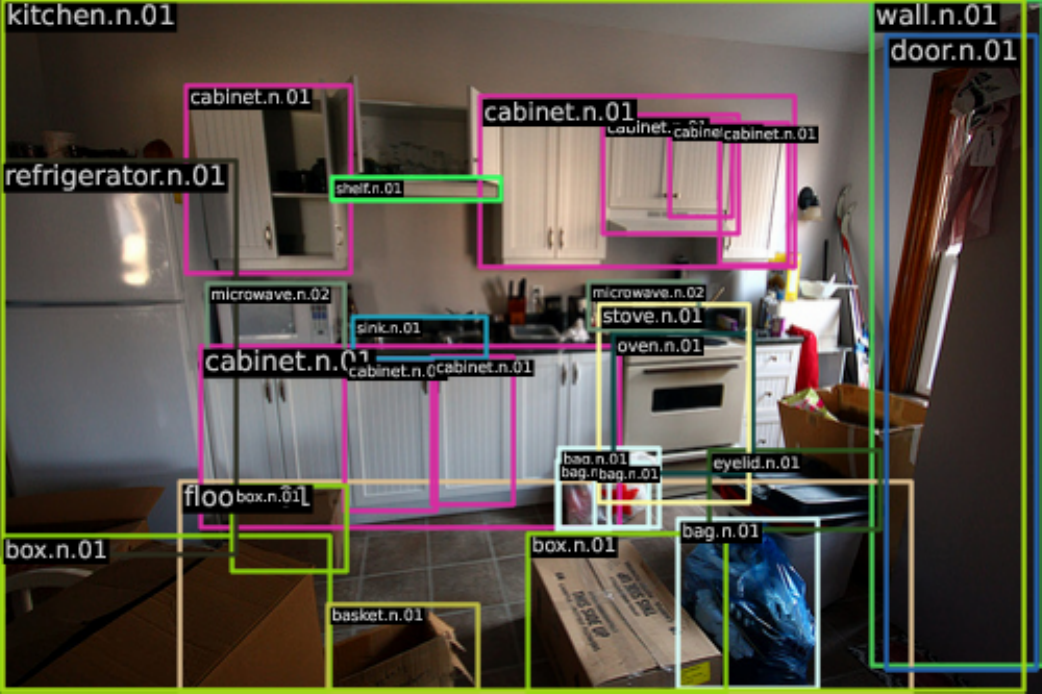} \\[8pt]
    \rotatebox{90}{\hspace{1.7em}F-RCNN*~\cite{lin2017feature}} &
    \includegraphics[height=\sz\linewidth]{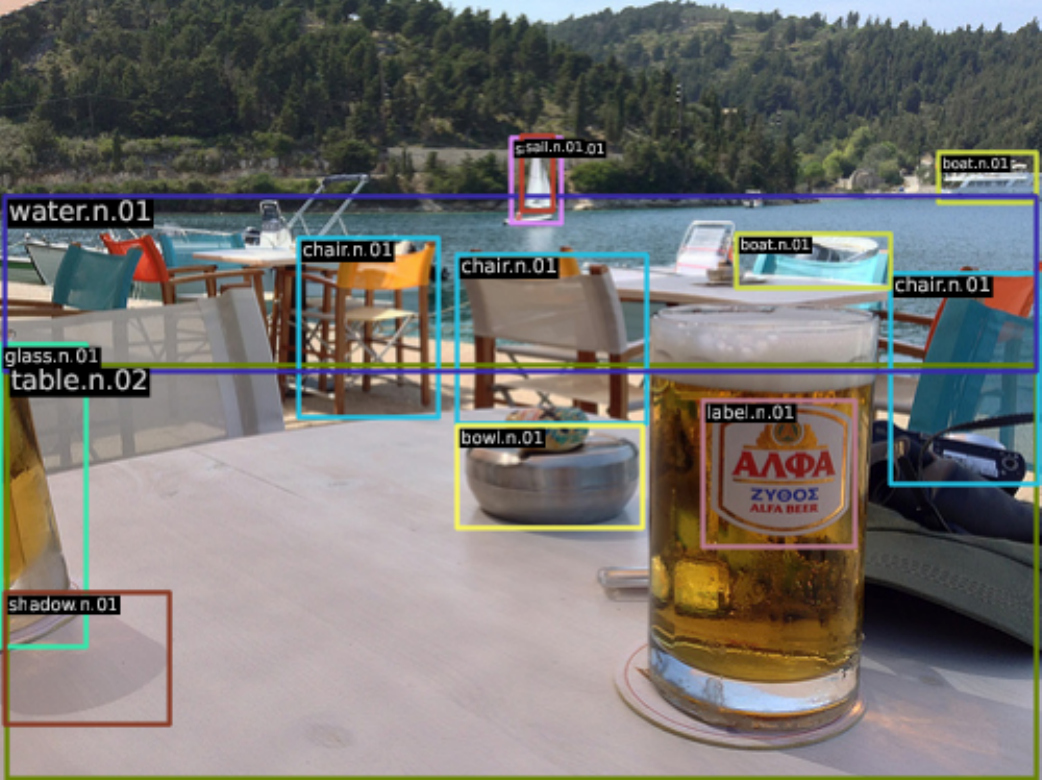} &
    \includegraphics[height=\sz\linewidth,width=0.175\linewidth]{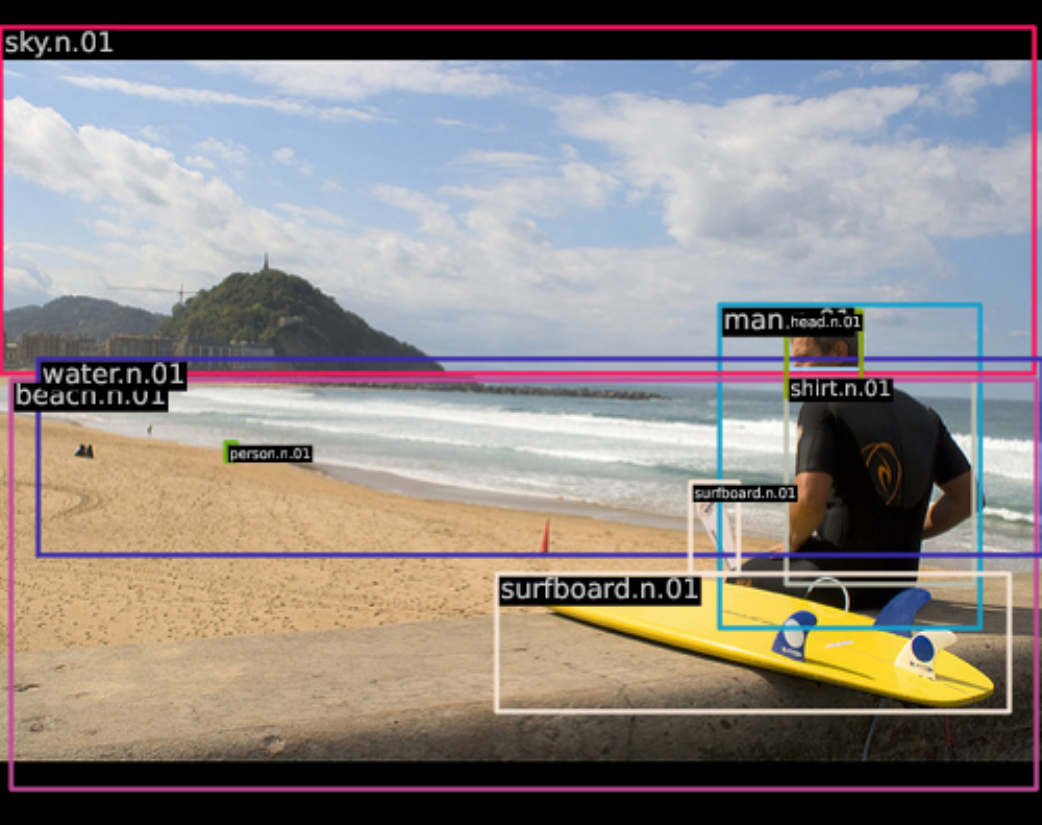} &
    \includegraphics[height=\sz\linewidth]{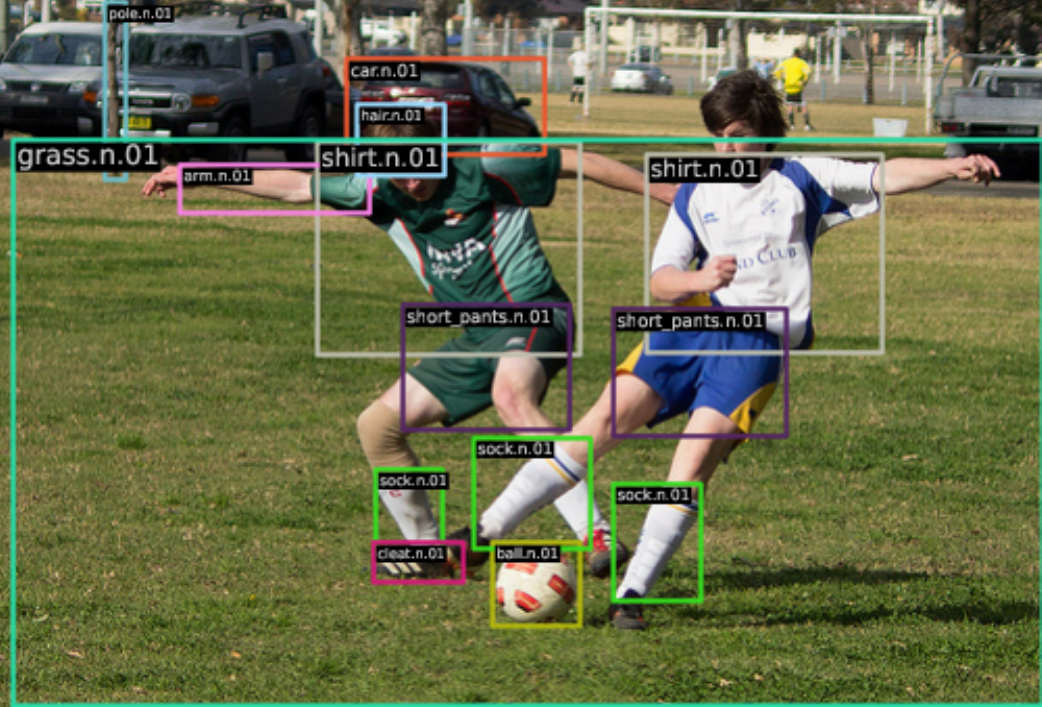} &
    \includegraphics[height=\sz\linewidth]{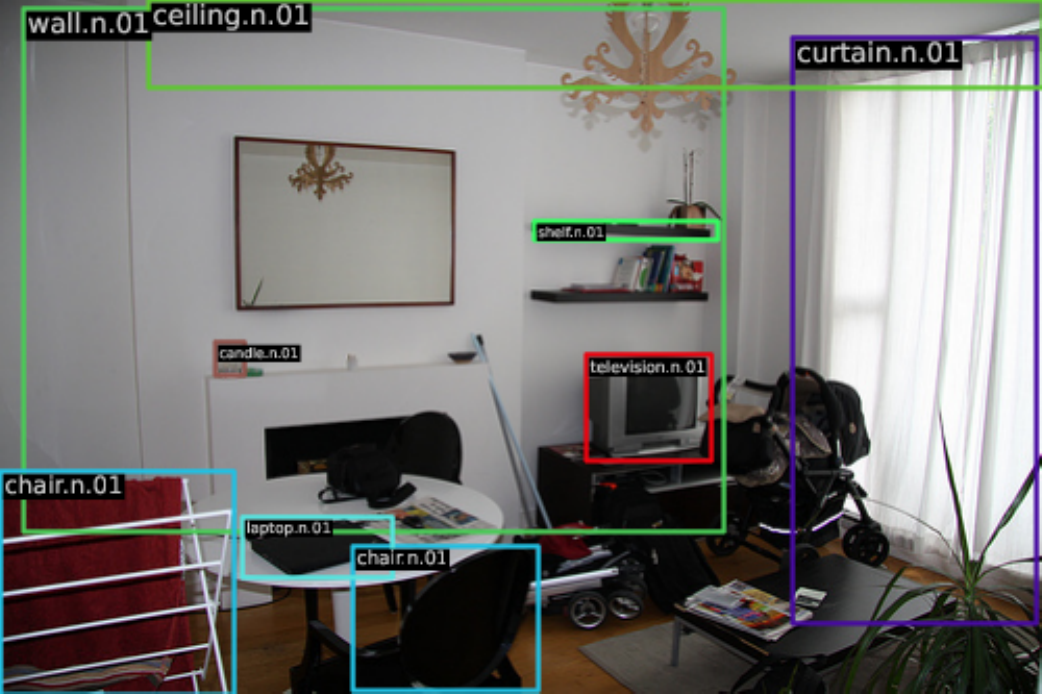} &
    \includegraphics[height=\sz\linewidth]{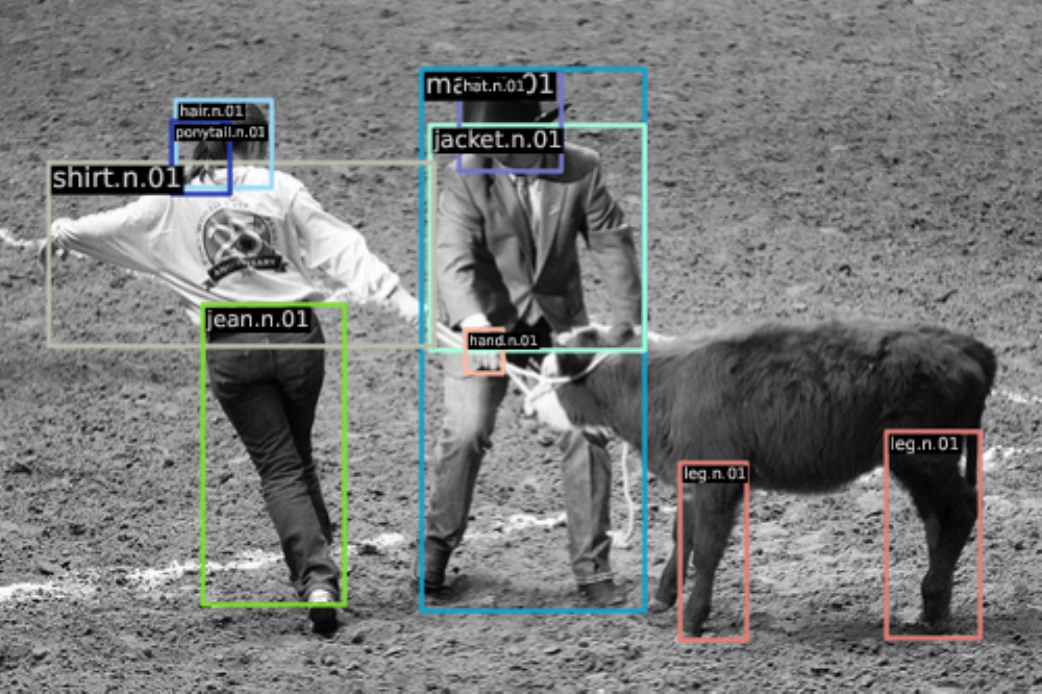} \\
    \rotatebox{90}{\hspace{0.5em}\textit{\textbf{This paper}}} &
    \includegraphics[height=\sz\linewidth]{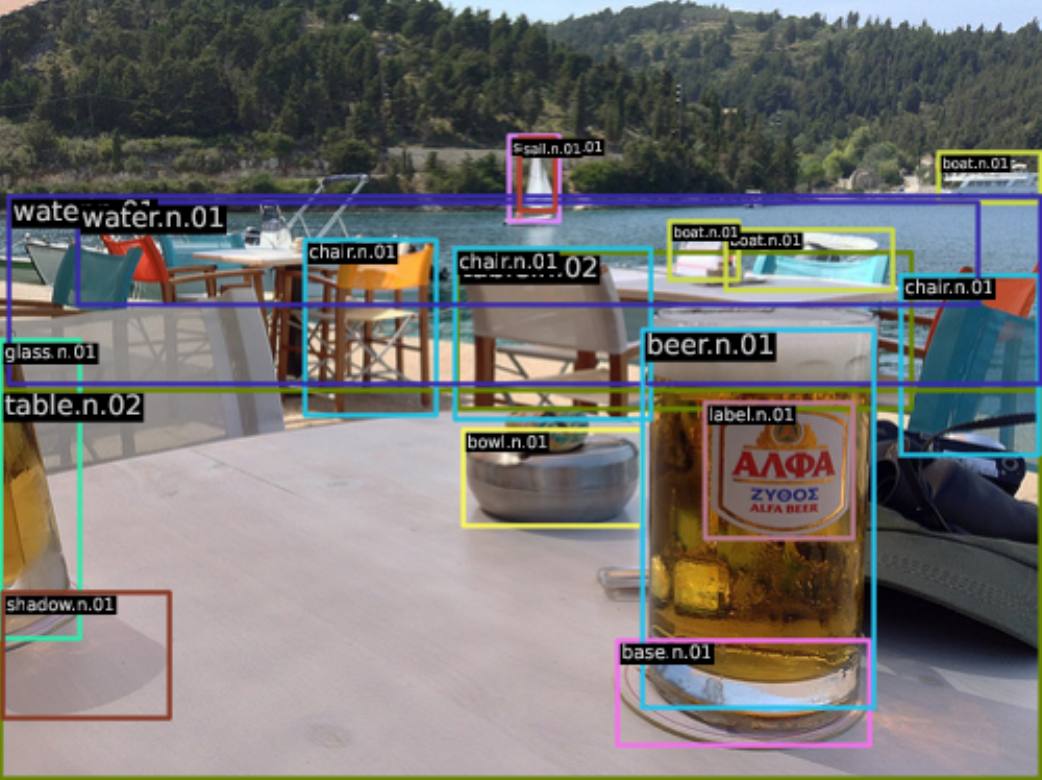} &
    \includegraphics[height=\sz\linewidth,width=0.175\linewidth]{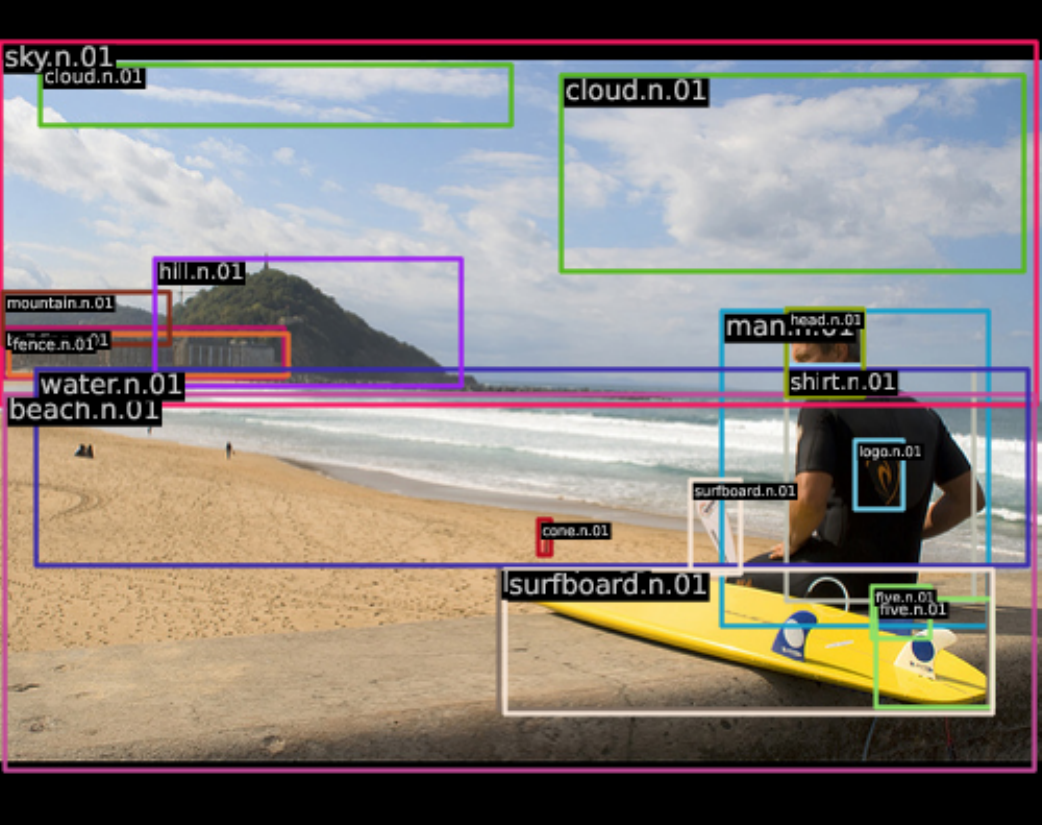} &
    \includegraphics[height=\sz\linewidth]{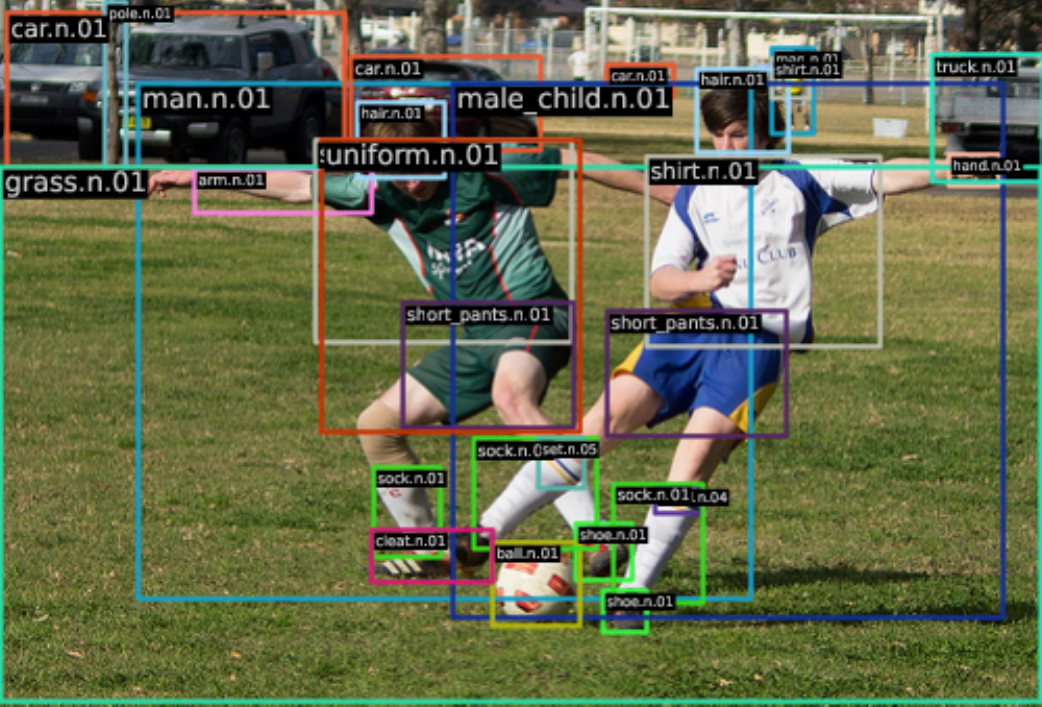} &
    \includegraphics[height=\sz\linewidth]{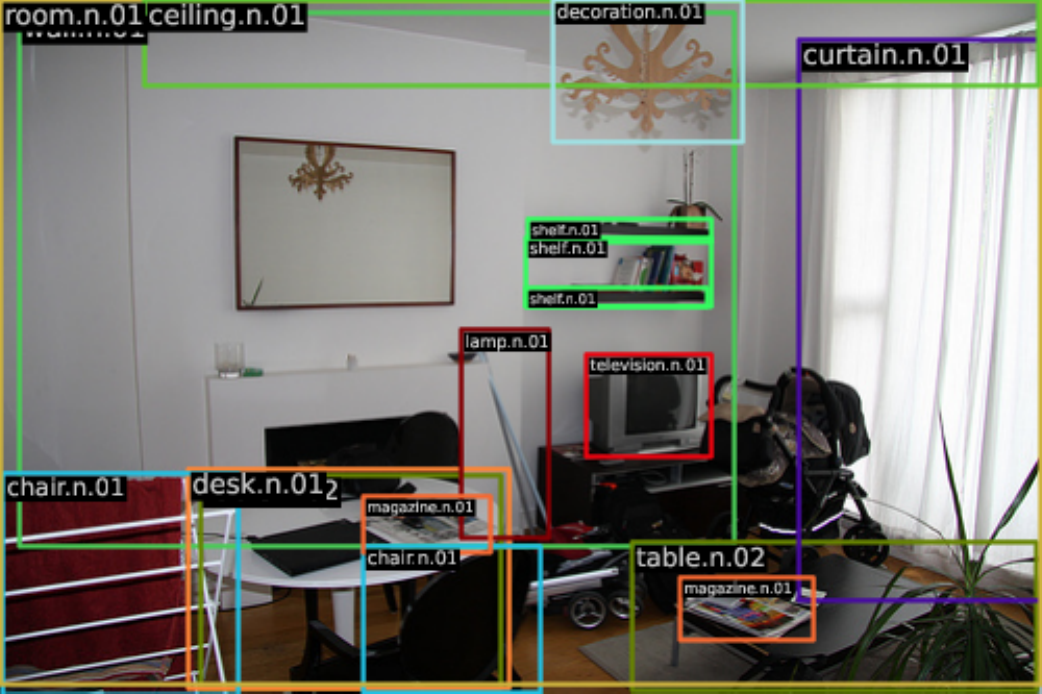} &
    \includegraphics[height=\sz\linewidth]{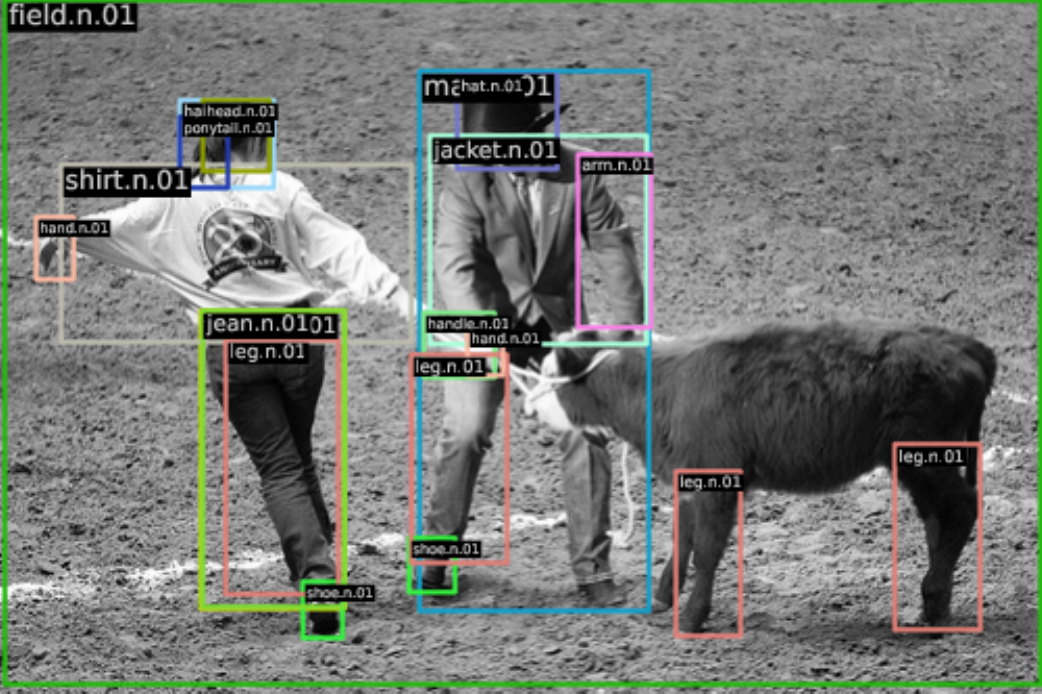} \\[0pt]
  \end{tabular} 
  \caption{\textbf{Qualitative results from the Visual Genome $\text{VG}_{1000}$ test set~\cite{krishna2017visual}.}
  The results show the positive effects of using our context-likelihood energy-refined graph. Especially for objects which are closely related to other objects such as legs and feet of animals and humans, co-occurring patterns, etc. Obtaining a better graphical representation of the image via our method helps our network to identify such objects with higher confidence.
  }
  \label{fig:qualitative_appendix}
\end{figure*}

We present additional qualitative results of our context-likelihood energy-refined graph in \cref{fig:qualitative_appendix}. 
We compare the results of Faster-RCNN~\cite{ren2015faster_rcnn, lin2017feature} to our method. For better readability, we only show results of both methods for proposals with a confidence of 0.5 or higher.
The main idea of our method revolves around the fact that it makes the algorithm aware of its surroundings. 
The prior object-relation information provided to our method has been calculated to reflect the fact that certain objects are more likely to appear together in an image. 
For instance, it is more likely to have more buildings next to another, which is seen in the second image of the first row. 
In the fourth image of the same row, Faster-RCNN predicts one rock at the background. 
But knowing that most of the time, in a beach setting, multiple rocks co-occur together helps our method to predict most of them correctly in the background.
Finally, we see the extent of our method for detecting tiny objects as well. 
In the second image of the second row, we see that Faster-RCNN predicts correctly the leg of the elephant. However it misses out on detecting the feet because it is small and does not have enough visible features to define what it is. Our method however combines the prior information that a foot most likely accompanies the leg, and thus correctly predicts the feet of the elephant.

\section{Failure Cases}
\begin{figure*}[h]
  \centering
  \footnotesize
  \setlength{\tabcolsep}{1.5pt}
  \newcommand{\sz}{0.173}
  \begin{tabular}{l ccccc}
    \rotatebox{90}{\hspace{2.5em}F-RCNN*~\cite{lin2017feature}} &
    \includegraphics[height=\sz\linewidth]{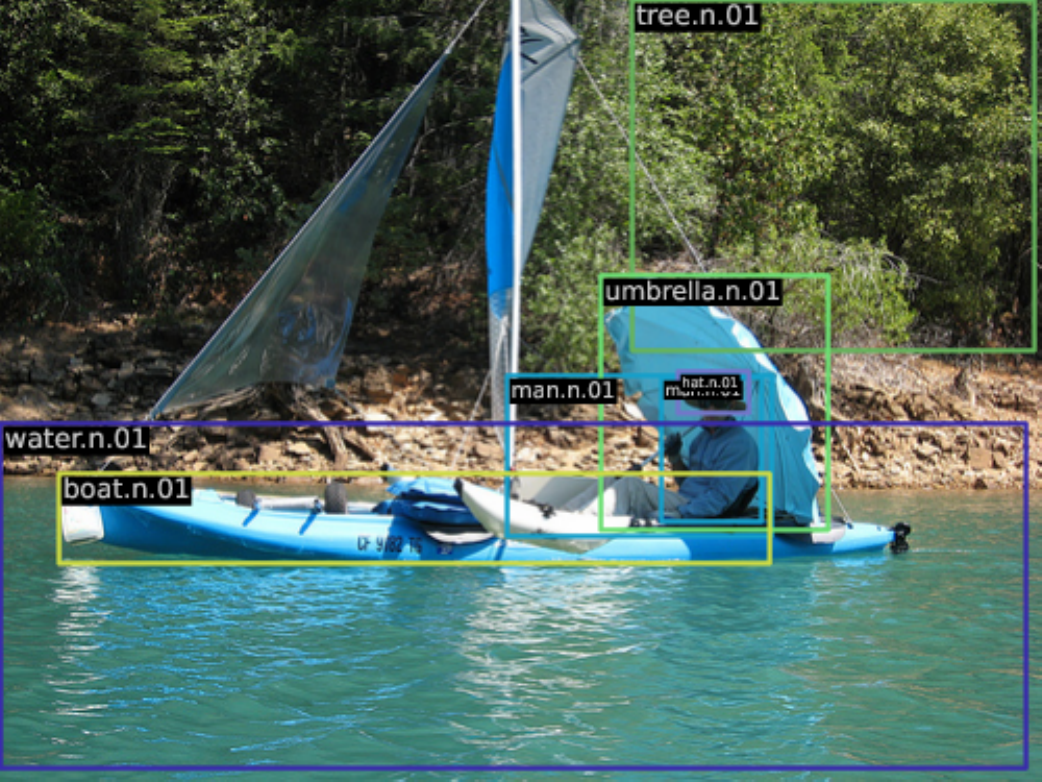} &
    \includegraphics[height=\sz\linewidth]{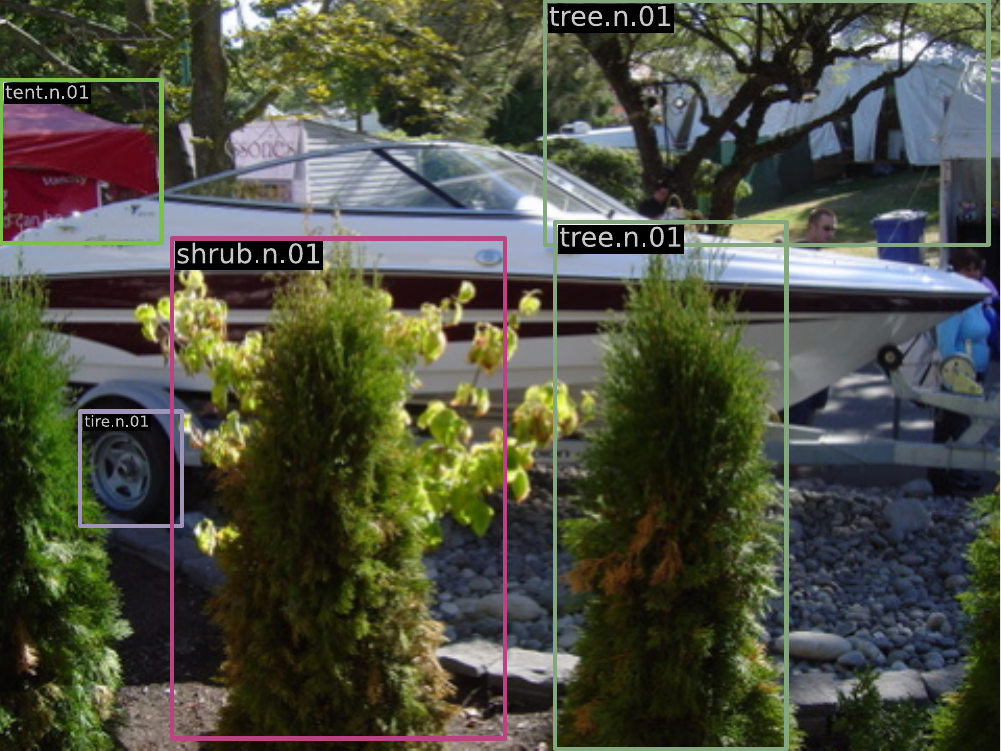} &
    \includegraphics[height=\sz\linewidth]{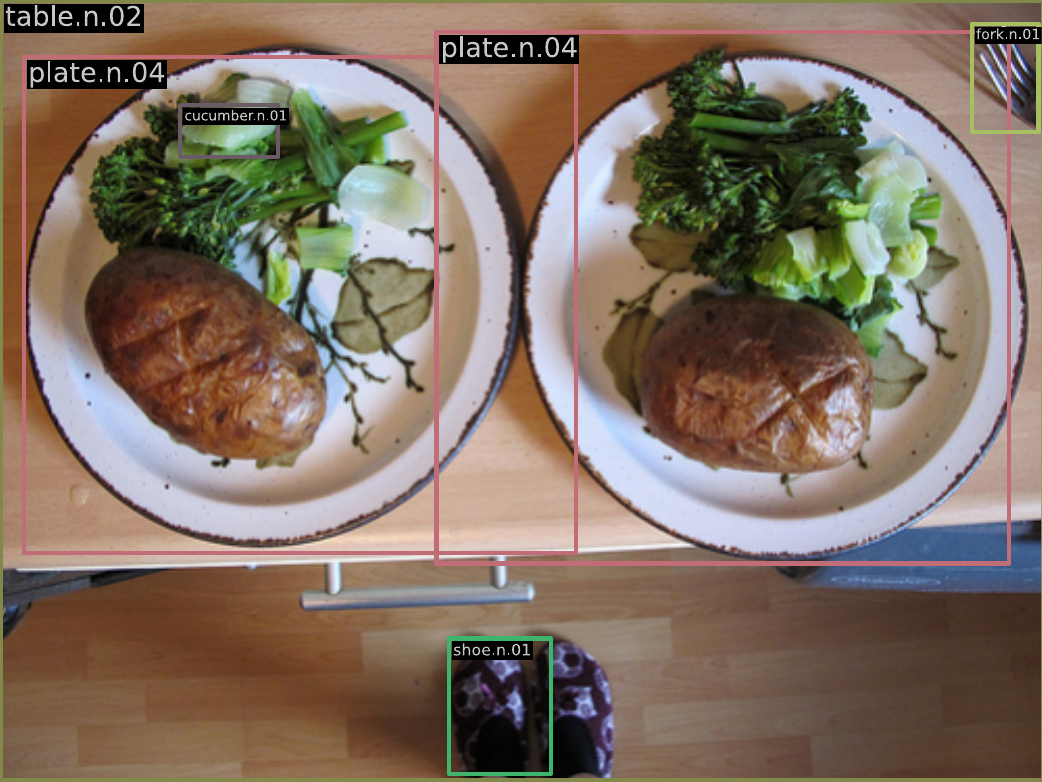} &
    \includegraphics[height=\sz\linewidth]{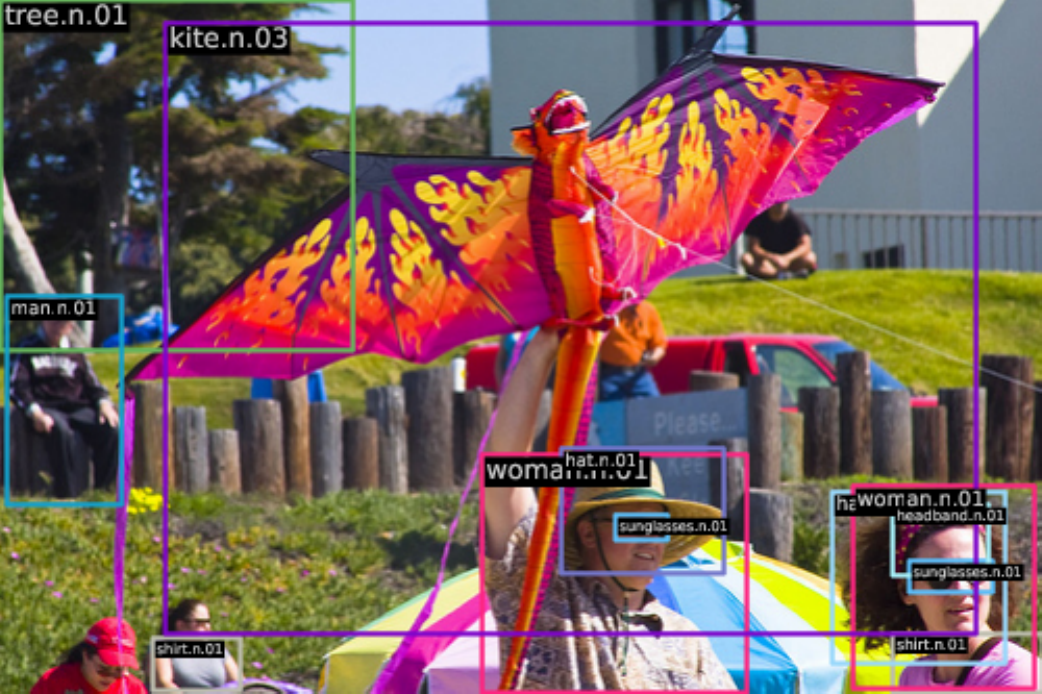} \\
    \rotatebox{90}{\hspace{1.8em}F-RCNN* w \Ours{}} &
    \includegraphics[height=\sz\linewidth]{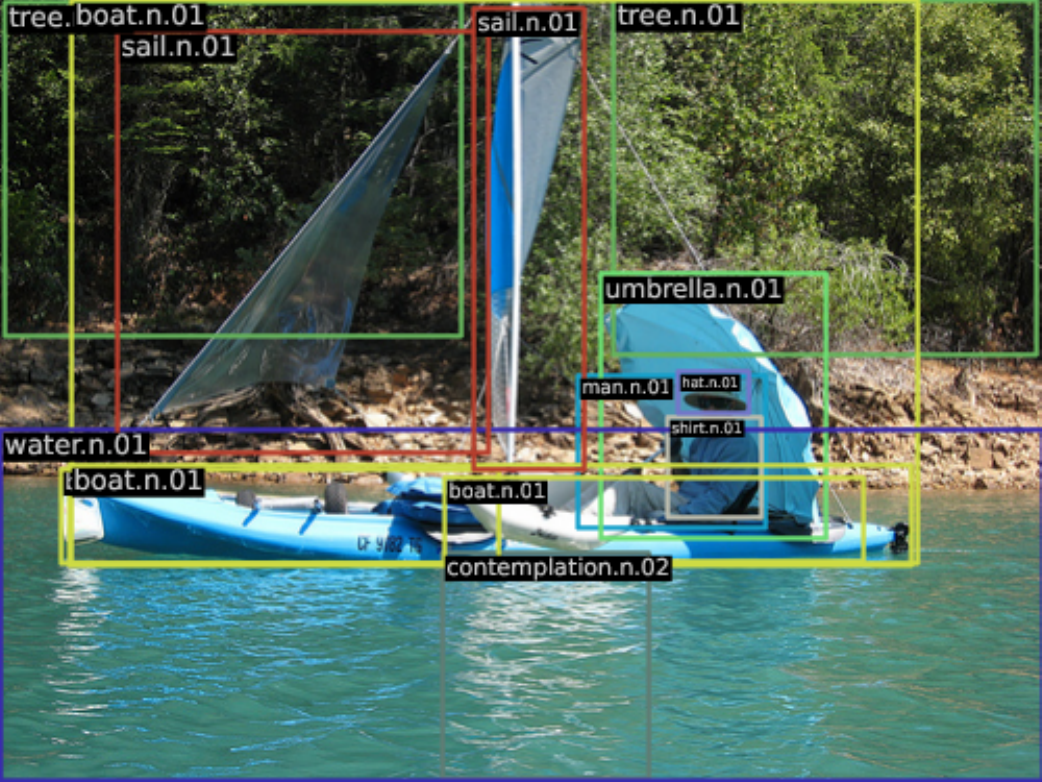} &
    \includegraphics[height=\sz\linewidth]{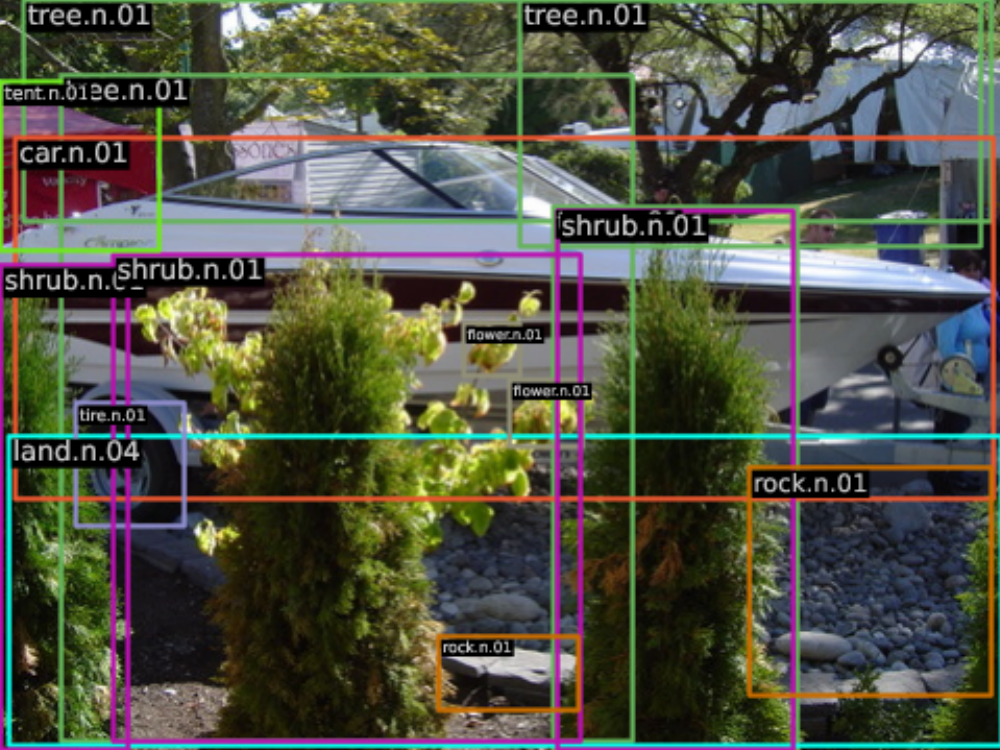} &
    \includegraphics[height=\sz\linewidth]{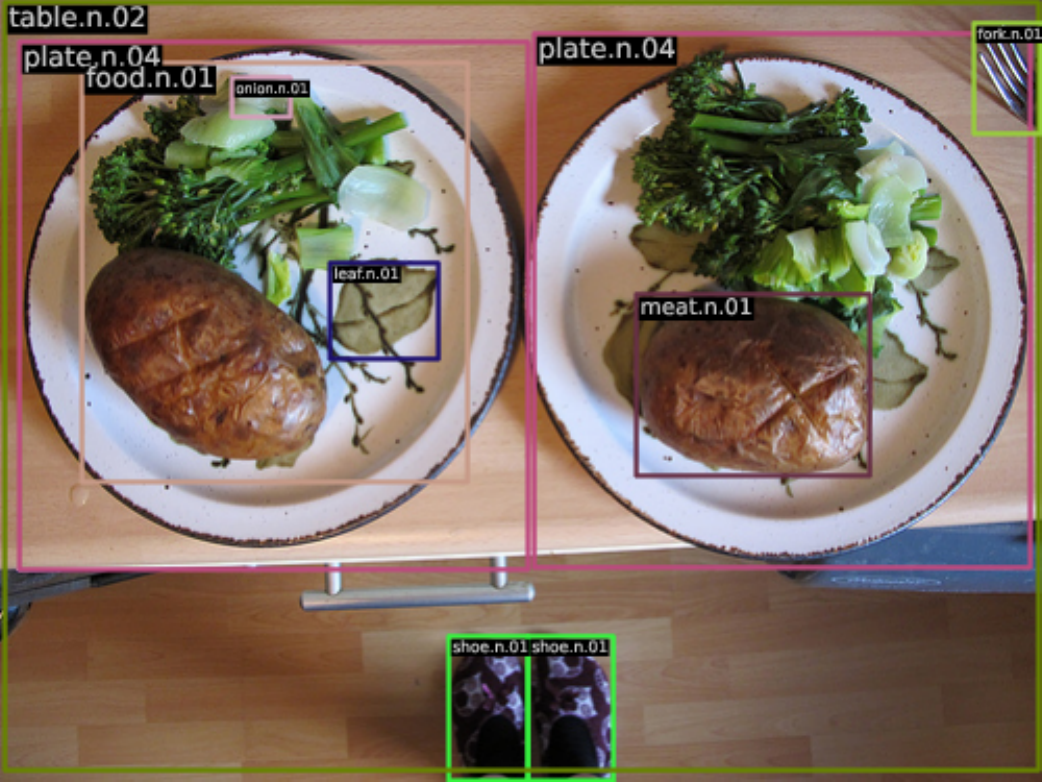} &
    \includegraphics[height=\sz\linewidth]{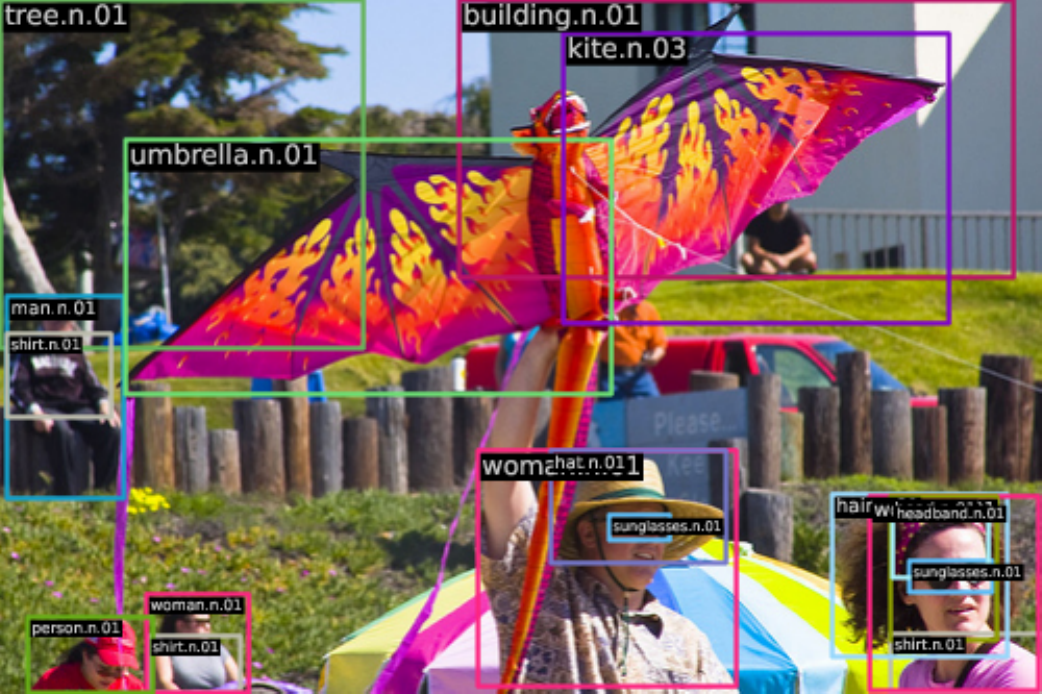} \\[0pt]
  \end{tabular} 
  \caption{\textbf{Failure cases.} In the images above, although we do detect most of the objects correctly, there are certain instances where our method fails. For instance, in the fourth image, our method mistakes a part of the kite as an umbrella. This could be due to the fact that umbrellas are often appear together with a human head when in an open state, which might have led to the wrong prediction.
  }
  \label{fig:failure_cases}
\end{figure*}

In \cref{fig:failure_cases}, we report some of the failure cases of our method. 
As can be seen, in some instances, where multiple objects can co-occur in a scene, our model relies on the object-relation information provided in the dataset and chooses the most likely one based on this prior knowledge. 
This can sometimes lead to overconfident wrong results. 
For instance, in the third image, our network mistakes a potato for meat, with high confidence, mostly due to the fact that a meat steak is also highly likely to be present on a plate, given that the object itself is hard to distinguish based on the visual features in the first place.

\end{document}